\documentclass{article}
\usepackage{iclr2027_conference,times}
\iclrfinalcopy

\usepackage{amsmath,amsfonts,bm}

\def\eqref#1{equation~\ref{#1}}

\def\1{\bm{1}}

\DeclareMathAlphabet{\mathsfit}{\encodingdefault}{\sfdefault}{m}{sl}
\SetMathAlphabet{\mathsfit}{bold}{\encodingdefault}{\sfdefault}{bx}{n}

\usepackage{hyperref}
\usepackage{url}
\usepackage{booktabs}
\usepackage{array}
\usepackage{graphicx}
\usepackage{flafter}
\usepackage{amsmath,amssymb}
\usepackage{microtype}
\usepackage{placeins}
\usepackage{tabularx,capt-of}
\usepackage{siunitx}
\usepackage{multirow}
\usepackage{mathtools}

\hypersetup{hidelinks,pdftitle={ReplayLens: Auditing Agents' Use of Outcomes}}

\title{ReplayLens: Auditing Agents' Use of Outcomes}
\author{
\textbf{Dong Xu$^{1,2}$, Zhangfan Yang$^{3}$, Jiantao Wu$^{1}$, Shipeng Zhang$^{1}$, Zexuan Zhu$^{1}$,}\\
\textbf{Jianqiang Li$^{1}$, Jun Zhang$^{1}$, Junkai Ji$^{1,2,\dagger}$}\\
\normalfont$^{1}$School of Artificial Intelligence, Shenzhen University\\
$^{2}$EasternDawn\\
$^{3}$School of Computer Science, University of Nottingham Ningbo\\
$^\dagger$Corresponding author.}

\newcommand{\method}{\textsc{ReplayLens}}
\newcommand{\noupdate}{\textsc{No update}}
\newcommand{\memory}{\textsc{Generic memory}}
\newcommand{\skill}{\textsc{Generic skill}}
\newcommand{\shuffle}{\textsc{Shuffled outcome}}
\newcommand{\replay}{\textsc{Permutation replay}}
\newcommand{\staticpred}{\textsc{Static predictor}}
\newcommand{\staticselected}{\textsc{Selected static}}
\newcommand{\contextucb}{\textsc{Contextual upper-confidence bound}}
\newcommand{\extrasampling}{\textsc{Extra sampling}}

\begin{document}
\maketitle
\fancyhead[L]{}

\begin{abstract}
When an agent reuses logged experience, a changed decision may reflect the recorded score, the action's name, or the record's position in storage. Standard memory evaluations do not reveal which relationship drives that change. We introduce \method{}, a black-box audit that changes one relationship in the stored history at a time, holds the remaining interface fixed, and measures the resulting decision. Four interventions target four relationships. \emph{Outcome reassignment} swaps which scores belong to which actions. \emph{Pair transport} moves intact action--score pairs to new record slots. \emph{Consistent renaming} relabels actions in both history and menu. \emph{Key-slot reassignment} changes both score attachment and position. A constructive separation shows why the audit is needed: two memory writers with identical endpoint accuracy respond differently to the same replay, so conventional evaluation cannot resolve the underlying dependence. On black-box LLM interfaces, swapping scores changes decisions while moving intact pairs does not, separating score attachment from record order. A bounded-memory study exposes ingestion-order sensitivity that endpoint comparison misses. In sequential experiment planning, altered historical scores redirect exploration and reduce final utility despite fresh measurements. A code-debugging agent with sealed hidden tests shows the same pattern outside model selection. \method{} provides a relationship-level audit for deciding whether logged experience can be merged, reordered, or reindexed safely.
\end{abstract}

\section{Introduction}

Persistent memory lets an agent carry experimental outcomes, tool feedback, and successful strategies into later trials or tasks~\citep{shinn2023reflexion,wang2023voyager,tu2026chain}. A growing body of work shows that memory substantially improves performance across benchmarks~\citep{xue2026pastbench,xia2025minerva}. We call the procedure that converts these records into retained state a \emph{memory writer}. It may build a textual summary, populate a key-value store, or update a learned representation.

As a motivating example, suppose a team merges two experiment logs after a schema migration. Strategy~\(A\) scored $0.9$ and strategy~\(B\) scored $0.1$. After the merge, the high score may remain in the log but attach to \(B\)'s identifier, or the intact \((A,0.9)\) pair may move to another record slot. Both strategies may also be renamed during the merge, preserving their score associations while changing their labels. The future agent still sees the same numerical values, yet its choice may change because the key--score--position relationships changed. An endpoint check can approve a migration with unchanged accuracy even when later decisions have changed.
\begin{table}[ht]
\centering
\begin{minipage}[c]{0.30\textwidth}
\small
\setlength{\tabcolsep}{2pt}
\begin{tabular*}{\linewidth}{@{\extracolsep{\fill}}rrr@{}}
\toprule
\textbf{Fit} & \textbf{Additive} & \textbf{Recurrent} \\
\midrule
1 & 0.000 & 0.738 \\
2 & 0.000 & 0.538 \\
3 & 0.000 & 0.654 \\
4 & 0.000 & 0.683 \\
5 & 0.000 & 0.796 \\
\bottomrule
\end{tabular*}
\end{minipage}\hfill
\begin{minipage}[c]{0.67\textwidth}
\setlength{\abovecaptionskip}{0pt}
\caption{\textbf{Matched endpoints, different replay responses.} Each independently fitted pair has identical aligned and empty-state actions. Entries are holdout action-disagreement proportions after moving intact key--outcome pairs, preserving the per-key sums defining the training target. Writers are frozen before replay. Each estimate averages five permutations over 48 holdout histories, and each action is scored on 48 fixed evaluator rows. Per-fit intervals are in Table~\ref{tab:content-fit-pairs}.}
\label{tab:endpoint-separation}
\end{minipage}
\vspace{-2mm}
\end{table}

Memory-on/off evaluation asks whether stored history changes behavior, while downstream performance asks whether the resulting choices are useful~\citep{xue2026pastbench}. Neither readout identifies \emph{which relationship} carries the effect or predicts which history transformations are safe. Consider a two-action history in which \(A\) receives a high score and \(B\) a low score. An agent may choose \(A\) by aggregating scores, preferring the name ``\(A\)'', or selecting the action in the first record slot. These rules agree on the original history and on the memory-free choice, yet they make different predictions when the history is replayed. Table~\ref{tab:endpoint-separation} makes this endpoint ambiguity concrete.

\begin{figure}[t]
\centering
\includegraphics[width=0.90\linewidth]{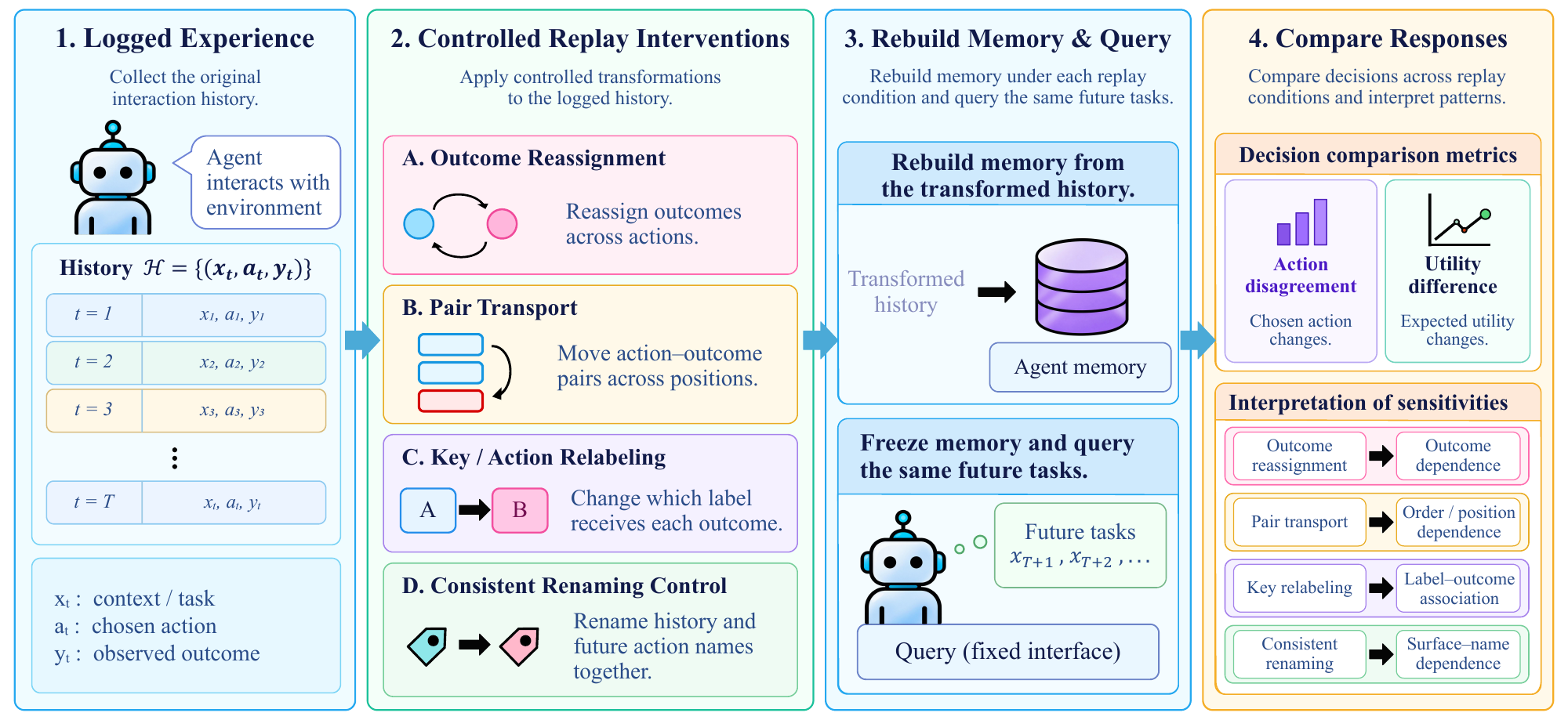}
\vspace{-2mm}
\caption{\textbf{Overview of the ReplayLens audit.} The audit starts from a logged action--outcome history and applies interventions with different preservation conditions. It then rebuilds memory under a fixed writer and initial state and queries the same future interface. Action disagreement and utility differences provide complementary readouts of the resulting dependence.}
\label{fig:overview}
\vspace{-6mm}
\end{figure}

The audit changes one relationship at a time while keeping the writer, initial state, future interface, and evaluator fixed. That relationship may concern which score belongs to which action, where a record sits, or what label it carries. Under this controlled comparison, a changed decision identifies sensitivity to the tested relationship. We measure both the fraction of changed decisions (\emph{action disagreement}) and the resulting change in task utility. Each intervention preserves specified relationships while changing others, so the joint response profile characterizes the dependence structure without requiring access to the agent's internal state. Figure~\ref{fig:overview} shows this replay boundary.

Our contributions are as follows.
\begingroup
\setlength{\leftmargini}{0.5\leftmargini}
\begin{itemize}
\item We define paired record interventions with explicit preservation conditions and use their joint responses to distinguish dependence on producer keys, outcome values, and pair placement.
\item We show that endpoint equivalence does not determine replay behavior. Five independently fitted writer pairs match aligned and memory-off decisions yet separate under intact-pair transport.
\item We demonstrate migration and deployment checks in order-sensitive storage, sequential experiment planning, and code debugging with an external evaluator.
\end{itemize}
\endgroup

\section{Related Work}

\noindent\textbf{Experience and memory.} Reflexion carries verbal feedback across trials~\citep{shinn2023reflexion}, Voyager stores reusable skills~\citep{wang2023voyager}, and Chain-of-Experience studies the reuse of iterative attempts and feedback~\citep{tu2026chain}. These methods determine \emph{what} experience reaches future decisions, but they do not test which structural relationship in that experience (outcome values, action names, or record order) drives the subsequent choice. \method{} addresses this second question.

\noindent\textbf{Diagnosing memory use.} Minerva designs programmable atomic and composite tasks to evaluate memory capabilities~\citep{xia2025minerva}. MemFail isolates failures in summarization, storage, and retrieval~\citep{garg2026memfail}, and MEMPROBE measures user-state recovery from a memory artifact~\citep{ma2026memprobe}. PAST-Bench evaluates retained-experience gains and evidence of the save--retrieve--update pathway~\citep{xue2026pastbench}. These evaluations test whether an agent recalls the correct fact and whether retrieval degrades under load. ReplayLens asks which relationship in a stored history—the score, the label, or the position—drives the agent's decision.

\noindent\textbf{Attribution and replay.} AttriMem estimates token contributions to a final answer~\citep{li2026attrimem}. CAMEO compares a memory entry in singleton and bank contexts~\citep{rocchi2026memorycredit}, and Causal Agent Replay replaces trajectory steps and measures downstream effects~\citep{shah2026causalreplay}. Those methods measure the contribution of an entry or trajectory step. ReplayLens retains the recorded outcomes and intervenes on their \emph{assignment} to actions or \emph{placement} in the history while holding the future interface fixed. REFLECT applies diagnosis-specific replay patches to test error attributions~\citep{lin2026reflect}. Related credit methods score memory updates~\citep{yan2026himpo}, delete entries before re-answering~\citep{jia2026hindsightmemoryprm}, or re-execute alternative agent actions~\citep{zhang2026creditnogt}.

\noindent\textbf{Offline and counterfactual evaluation.} Offline and off-policy evaluation estimates a policy's value from logged trajectories, often with importance weighting or doubly robust estimators~\citep{jiang2016doubly}. ReplayLens also transforms logged experience, but its estimand is different: it changes a declared history-to-memory relationship and measures future decisions and externally evaluated utility under a fixed interface. It does not estimate policy value from behavior-policy propensities, so this connection clarifies the counterfactual replay perspective without conflating value estimation with relation-level auditing~\citep{mesnard2021counterfactual}.

\noindent\textbf{Behavioral and invariance testing.} Underspecification studies show that similar test performance can conceal different deployment behavior~\citep{damour2022underspecification}. Our endpoint-separation result instantiates this issue for memory writers. CheckList tests minimum functionality, invariance, and directional expectations~\citep{ribeiro2020checklist}, while the pair-transport zero law predicts which writers should remain invariant to pair reordering. Causal tracing localizes effects through internal activation interventions~\citep{meng2022rome}. ReplayLens instead evaluates externally accessible records and decisions. Its contrasts are structurally related to causal mediation analyses that perturb an intermediate state~\citep{vig2020causalmediation}, but its estimand is behavioral and conditioned on the writer, initial state, future tasks, and evaluator.

\section{Auditing Outcome Assignment}

The audit changes one relationship in a history at a time and holds the remaining interface fixed. A change in the future decision therefore identifies the relationship tested by that intervention. Four interventions target four relationships.
\emph{Outcome reassignment} tests whether the agent follows scores by swapping which scores belong to which actions.
\emph{Pair transport} tests whether record position matters by moving intact action--score pairs to new record slots.
\emph{Consistent renaming} tests whether the agent tracks labels by relabeling actions in both history and future menu, then decoding back.
\emph{Key-slot reassignment} provides a combined test by moving action names over unchanged scores, changing both attachment and position.

Every replay starts from the same initial state, uses the same writer, asks the same future query, and applies the same external evaluator. We call the association between a selected action and its observed result \emph{outcome assignment}. The writer, policy, and evaluator remain black boxes, so the audit can be applied to learned representations, retrieval-augmented pipelines, and prompted LLMs with in-context memory.

\begin{figure}[ht]
\centering
\includegraphics[width=\linewidth]{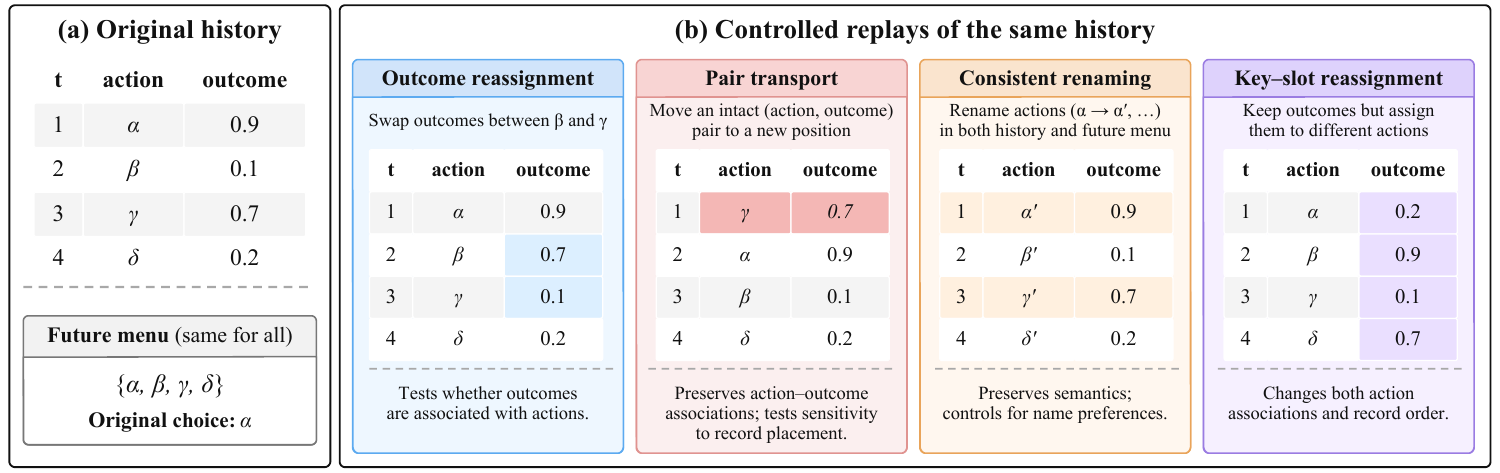}
\vspace{-6mm}
\caption{\textbf{A worked outcome-assignment audit.} The left panel shows the original action--outcome log and shared future menu. Each replay panel changes one relationship. Outcome reassignment changes which value belongs to a key. Pair transport moves an intact pair. Consistent renaming changes history and menu labels together. Key-slot reassignment changes both attachment and placement. Every condition uses the same writer, initial state, future query, and evaluator. Figure~\ref{fig:overview} gives the complete replay boundary.}
\label{fig:outcome-assignment}
\vspace{-2mm}
\end{figure}

\noindent\textbf{A four-action example.} Figure~\ref{fig:outcome-assignment} starts with a four-action history whose outcomes are $0.9,0.1,0.7,0.2$. Consider the two-action slice where \(A\) has score $0.9$ and \(B\) has score $0.1$, with a future menu \(\{A,B\}\). A rule that averages outcomes selects \(A\). After outcome reassignment (swapping the values so that \(A\) receives $0.1$ and \(B\) receives $0.9$), it selects \(B\). Pair transport moves the intact pairs to new record slots without breaking the key--score binding, so the same averaging rule still selects \(A\). A first-record rule, however, may switch to \(B\) because position has changed. Consistent renaming replaces \(A,B\) with fresh labels in both the history and future menu and preserves the executable choice after decoding. Key-slot reassignment moves names over unchanged scores, changing both attachment and position simultaneously. Each colored panel in the figure changes one relationship while preserving those named in its caption.

\subsection{Trace, intervention, and readout}
\label{sec:assignment-interface}

We formalize the audit boundary by specifying what each intervention preserves. A \emph{writer} converts (task, action, outcome) records into persistent state, and a \emph{policy} reads that state to choose a future action. Each intervention changes only the records supplied to the writer; the policy, evaluator, future tasks, and future interface remain fixed. The comparison is therefore ``same agent, same future question, different history arrangement.''

Let $D=((x_t,a_t,y_t))_{t=1}^{T}$ be a history of task inputs, selected actions, and observed utilities. Writer $W_m$ maps records to persistent state, and policy $\pi_m(s,x)$ returns an action for future task $x$. An intervention is specified by a record condition $q$ (which relationship to change) and an action-label permutation $\rho$ (which relabeling to apply). Together they produce a transformed history $D_q(\rho)$, yielding
\begin{equation}
S_{m,q}(\rho)=W_m(D_q(\rho)),\quad
a_{m,q}(x;\rho)=\pi_m(S_{m,q}(\rho),x).
\label{eq:main-state-interface}
\end{equation}
Each comparison fixes the initial state, writer parameters, and label bindings. Slot permutations stay within complete blocks sharing a task family and context. Write $k_t^\rho$ for the key sent to the writer when $\rho$ relabels action $a_t$. The mapping from displayed keys to executable actions remains fixed. Appendix~\ref{app:formal-interface} gives the complete key--utility map for each condition.

The baseline is the \emph{aligned} condition ($\rho=\mathrm{id}$), which feeds the original history unchanged. Each intervention departs from this baseline in a specific way. Value reassignment keeps keys in place but shuffles their scores. Pair transport moves intact key--score pairs to new slots. Key-slot reassignment moves keys over unchanged scores. The primary synthetic and joint black-box protocols use the same permutation for value and pair cells, so both see the same utility sequence and only the key--score pairing differs. Appendix~\ref{app:formal-interface} gives the full permutation laws.

\noindent\textbf{Consistent renaming and reassignment.} Renaming both history and future menu labels while decoding back should leave a policy that is equivariant to labels unchanged. The policy sees a relabeled world, but the mapping back to executable actions compensates. Renaming only history keys while fixing the menu instead changes which key receives each outcome, acting as a form of reassignment rather than a neutral relabeling. The black-box experiment crosses these two choices to distinguish the cases (Appendix~\ref{app:global-renaming}).

\subsection{Response measurements}

We use concrete readouts at the replay boundary. The primary readout is \emph{action disagreement}, the fraction of future tasks that receive a different action after the intervention. Let $d(a,b)$ denote this fraction and let $X^+$ be the future task set. The remaining readouts separate action changes from utility changes.
\begin{equation}
d(a,b)=\frac{\sum\nolimits_{x\in X^+}\mathbf{1}[a(x)\ne b(x)]}{|X^+|},\quad
A_q(\rho)=d(a_q^\rho,a_q^{\mathrm{id}}),\quad
D_A^q(\rho)=d(a_q^\rho,a_{\mathrm{aligned}}^\rho).
\label{eq:main-action-readouts}
\end{equation}
Here $A_q(\rho)$ measures how much relabeling actions (via map $\rho$) changes decisions within condition $q$, and $D_A^q(\rho)$ compares condition $q$ against the aligned baseline under the same map. The producer-key response $A_{\mathrm{rekey}}$ averages $A_{\mathrm{aligned}}(\rho)$ over nine derangements. We also score each future choice with the same task evaluator across conditions. Mean utility $U$ and its paired differences quantify task consequences. A changed choice may be better or worse, so consequences are evaluated separately from action disagreement. Oracle-agreement and interaction readouts are defined in Appendix~\ref{app:interaction-readouts}.

\noindent\textbf{Correcting for decoding noise.} Hosted models may return different outputs for the same input across calls. We therefore subtract within-condition disagreement from between-condition disagreement, using two fresh calls per cell with decoded actions $Y_{c1},Y_{c2}$. Intuitively, $T_{\mathrm{gr}}$ measures whether decision differences between two conditions exceed the random variation within either condition. For distinct cells $c,e$,
\begin{equation}
T_{\mathrm{gr}}(c,e)=\frac14\sum\nolimits_{r,s=1}^{2}\mathbf{1}[Y_{cr}\ne Y_{es}]
-\frac12\sum\nolimits_{j\in\{c,e\}}\mathbf{1}[Y_{j1}\ne Y_{j2}].
\label{eq:main-renaming-response}
\end{equation}
If the two conditions have the same action distribution, $T_{\mathrm{gr}}$ is zero in expectation; a positive value indicates a between-condition shift beyond the within-condition variation. Finite-sample estimates can be negative. Details and raw disagreements are in Appendix~\ref{app:joint-blackbox-results}.

\subsection{Interpretation and resolution}

Two results provide the audit's theoretical basis: replay strictly refines conventional evaluation, and the pair-transport zero law supplies a falsifiable calibration.

\noindent\textbf{Endpoint-matched separation.} For any stateful writer, a lookup table can copy its aligned actions on the declared future support, matching endpoint statistics including the memory-on versus memory-off contrast. The lookup remains invariant to replay, whereas the stateful writer may respond. The replay profile therefore separates writers that endpoint measurements cannot distinguish (~\ref{app:endpoint-matched-witness} and~\ref{app:diagnostic-dominance}).

\noindent\textbf{Pair-transport zero law.} If a writer accumulates state by summing a position-independent feature of each (key, score) pair, then reordering intact pairs within a block leaves the sum, and hence every future decision, unchanged~\citep{zaheer2017deepsets}. Formally, if $S_B=s_{0,B}+\sum\nolimits_{t\in B}\phi_B(k_t,u_t)$ with $\phi_B$ independent of slot position, and the global state combines blocks additively, then
\begin{equation}
S_{m,\mathrm{pair}}(\rho)=S_{m,\mathrm{aligned}}(\rho)
\quad\Longrightarrow\quad D_A^{\mathrm{pair}}(\rho)=0.
\label{eq:main-pair-zero-law}
\end{equation}
A nonzero pair-transport response therefore diagnoses a violation of order invariance, indicating that the writer's state depends on \emph{where} records appear, not just \emph{what} they contain. Section~\ref{sec:calibration} tests this prediction against writers with known update laws. For hosted models, repeated calls measure the additional contribution of decoding variation.

\section{Experiments}

The experiments move from known update laws to black-box and closed-loop settings. Synthetic writers establish that replay separates endpoint-equivalent writers (\S\ref{sec:endpoint-separation}) and recovers their known dependence structures (\S\ref{sec:calibration}). We then test black-box LLM interfaces and bounded-memory pipelines with explicit storage budgets (\S\ref{sec:blackbox-assignment}, \S\ref{sec:memory-repair}). Sequential experiment planning, code debugging with hidden tests, and cross-model protocol comparisons connect replay responses to downstream consequences (\S\ref{sec:sequential-agent}, \S\ref{sec:code-debugging}, \S\ref{sec:cross-model}).

\subsection{Experimental setup}

\noindent\textbf{A decision with an externally measured outcome.} Our principal black-box setting asks the model to choose among four sampling strategies for a classification task (coverage, adaptive, replicate, and mixed), each selecting 24 observations from a candidate pool. A fixed ridge-regularized logistic evaluator, hidden from the model, fits on the selected observations and scores them by log loss on a disjoint evaluation pool. Utility is the exponential of negative log loss. The model sees the task family, context, and previously selected action--utility records, but never the unselected outcomes or evaluation labels.

The tasks use the Breast Cancer Wisconsin (Diagnostic) and Wine Quality datasets~\citep{street1993nuclear,cortez2009wine}. A malignant diagnosis and wine quality of at least six define the positive labels. Each family has an independent and identically distributed (i.i.d.) context and a shifted context, where candidate and evaluation pools are separated by the first standardized feature. Appendix~\ref{app:public-data-replication} specifies the pools and evaluator.

\noindent\textbf{Black-box experiments.} The joint challenge uses GPT-6 Astra with max reasoning effort and the ordinary service tier, crossing five ledger forms, four history/menu cells, two namespaces, and two fresh repeats over 48 streams. The provider documentation specifies the \texttt{gpt-6-astra} model alias, supported reasoning settings, context window, and snapshot mechanism; the hosted interface does not expose a parameter count, so we report the model alias and the runtime settings recorded for this evaluation. Each request returns four decisions, one per family--context block. Every 24-record history covers all four actions, constructed from recorded outcomes so that fixed competing reference rules disagree. This is a retrospective test on designed histories (Appendix~\ref{app:method-joint-blackbox}).

The appendices also report archival GPT-5.6 and frozen Qwen3-8B readouts. The finite-memory experiment uses 16 development and 48 test streams with four bounded memory implementations. Development diagnostics select a candidate before test histories are executed (Appendix~\ref{app:method-memory-repair}).

\noindent\textbf{Controlled writers and inference.} Synthetic writer tasks cover source localization, temporal-discount estimation, and survival-risk estimation, with two contexts per family and 48 future tasks per stratum in 48 streams. Each task family provides four actions with utility distributions that make the optimal choice sensitive to how the writer accumulates history. Fitted parameters are frozen before evaluation so that writers do not adapt to the intervention. Interventions are paired within streams, which are the units for bootstrap intervals. The memory-selection test uses two co-primary contrasts with individual 97.5\% intervals, and other reported intervals are 95\%. Full uncertainty, request-level diagnostics, and alternative readouts remain in the appendix.

\subsection{Endpoint equivalence leaves replay dependence unresolved}
\label{sec:endpoint-separation}

This experiment demonstrates that two writers can be indistinguishable by endpoint evaluation on the declared support yet separable by replay. We train two architecturally different writers to the same target, namely selecting the key with the largest sum of centered utilities. The \emph{additive} writer learns a readout over per-key sums, and the \emph{recurrent} writer (16-dimensional hidden state) learns sequential updates. Both are trained on histories with four contiguous eight-record blocks, where the winning block has utilities in $0.78$--$0.94$ and others in $0.06$--$0.22$. Five independently fitted pairs are each evaluated on 96 unseen sequences after freezing parameters. The endpoint gate verifies that aligned and empty-state actions match before writer labels are opened.

Despite matching endpoints, the two writers diverge when records are rearranged. Pair transport changes recurrent decisions in every fit while preserving additive decisions (Table~\ref{tab:endpoint-separation}). The training target depends only on per-key sums, so moving intact pairs preserves all target-relevant information. Yet the recurrent writer has acquired an additional dependence on record arrangement that conventional evaluation cannot detect. The recurrent architecture naturally encodes order, and the training objective leaves this sensitivity unconstrained. Bootstrap intervals are above zero in every recurrent fit (Table~\ref{tab:content-fit-pairs}). A same-architecture control with permutation augmentation reduces pair response but fails the learning-quality gate (Appendix~\ref{app:method-learned-control}). A finite-support witness provides the theoretical counterpart. For any stateful writer one can construct a lookup table that matches its endpoints yet remains unchanged under every replay intervention, proving that endpoint equivalence is strictly weaker than replay equivalence (Appendix~\ref{app:diagnostic-dominance}).

\subsection{Known writer laws identify relation-specific responses}
\label{sec:calibration}

We validate each intervention with writers whose update laws are known analytically (Table~\ref{tab:content-calibration}). A key-only writer counts action appearances without using outcomes, whereas a commutative-pairing writer accumulates key--outcome features. Both respond to key relabeling because both track which keys appear. Value reassignment separates them: with keys fixed, it changes the pairing writer's decisions and leaves the key-only writer unchanged. The pairing writer is invariant to pair transport because its additive update depends on key--outcome \emph{content}, not record \emph{position}. Key-relabeling sensitivity alone therefore cannot distinguish exposure counting from score use.

\begin{table}[ht]
\vspace{-4mm}
\centering
\small
\setlength{\tabcolsep}{3pt}
\caption{Canonical calibration of four writer families. $n_{\rm op}$ counts parameterizations, and Profiles counts distinct response profiles over 48 streams. All parameterizations are classified correctly. Responses average parameterizations, maps, and tasks. Positive scaling or recency settings can produce duplicate profiles (Appendix~\ref{app:hidden-operator-family}).}
\label{tab:content-calibration}
\begin{tabular*}{\linewidth}{@{\extracolsep{\fill}}lrrrrr@{}}
\toprule
\textbf{Family} & $n_{\rm op}$ & $A_{\rm rekey}$ & $D_A^{\rm value}$ & $D_A^{\rm pair}$ & \textbf{Profiles} \\
\midrule
Replay-invariant & 3 & 0.000 & 0.000 & 0.000 & 1 \\
Key-only & 3 & 0.664 & 0.000 & 0.000 & 1 \\
Commutative-pairing & 3 & 0.666 & 0.163 & 0.000 & 1 \\
Order-sensitive & 3 & 0.667 & 0.000 & 0.090 & 2 \\
\bottomrule
\end{tabular*}
\vspace{-2mm}
\end{table}

A family-blind challenge tests the pair-preservation law on hidden constructions using a response threshold fixed before coefficient generation. It identifies every pair-preserving writer and misses one pair-displacing writer (Table~\ref{tab:content-blind}). The missed writer changes state without switching enough decisions to cross the threshold, so the finite query set limits diagnostic resolution. Appendix~\ref{app:blind-relation-challenge} gives the generating laws, threshold, and complete classification results.

Appendices~\ref{app:hidden-operator-holdout}--\ref{app:nonlinear-holdout} report composite updates, alternative readouts, and nonlinear state representations in the calibration suite.

\subsection{Black-box responses separate score attachment from placement}
\label{sec:blackbox-assignment}

We apply the audit to GPT-6 Astra as a black-box interface. On the tested support, changing \emph{which score belongs to which action} changes decisions, while moving \emph{intact action--score pairs} to new slots leaves decisions unchanged (Table~\ref{tab:joint-main-effects}). The two conditions share the same utility sequence, so this contrast isolates score attachment from record position. Consistent renaming of history and menu preserves semantic actions, while history-only relabeling acts as reassignment. Fresh repeats agree, so the pattern remains after within-condition correction.

The designed histories let us compare against fixed reference rules. On full and value-reassigned histories, Astra agrees with the historical-mean rule in both namespaces (Table~\ref{tab:joint-main-reference}). This is a behavioral description on the tested support, not a claim about the model's internal mechanism. Removing utility values changes the pattern. Consistent renaming now produces a response with semantic aliases, showing that outcome values can stabilize naming behavior. We use two display namespaces in this challenge: the \emph{Semantic} group uses readable action labels, whereas the \emph{Opaque} group uses token-matched opaque labels. The groups share the histories, future tasks, and evaluators. Full reference comparisons and archival results are in Appendices~\ref{app:joint-blackbox-results} and~\ref{app:archival-blackbox-results}.

\begin{table}[ht]
\vspace{-4mm}
\centering
\small
\begin{minipage}[t]{0.595\linewidth}
\small
\setlength{\tabcolsep}{0pt}
\captionof{table}{Action responses over 48 streams. Within each \emph{Semantic} or \emph{Opaque} group, \emph{Full} compares the row with the full ledger, \emph{Hist.} rekeys the history while keeping the future menu fixed, and \emph{Hist.+menu} rekeys both history and menu. Two fresh requests per cell. Intervals are in Appendix~\ref{app:joint-blackbox-results}.}
\label{tab:joint-main-effects}
\begin{tabular*}{\linewidth}{@{\extracolsep{\fill}}lrrrrrr@{}}
\toprule
Form & \multicolumn{3}{c}{Semantic} & \multicolumn{3}{c}{Opaque} \\
\cmidrule(lr){2-4}\cmidrule(l){5-7}
 & Full & \shortstack{hist.\\rekey} & \shortstack{hist.+\\menu} & Full & \shortstack{hist.\\rekey} & \shortstack{hist.+\\menu} \\
\midrule
Full & \textemdash{} & 1.000 & 0.000 & \textemdash{} & 1.000 & 0.000 \\
Value reassigned & 0.807 & 1.000 & 0.000 & 0.807 & 1.000 & 0.000 \\
Pair & 0.000 & 1.000 & 0.000 & 0.000 & 1.000 & 0.000 \\
Key-slot & 0.766 & 1.000 & 0.000 & 0.766 & 1.000 & 0.000 \\
Utility-masked & 0.531 & 0.135 & 0.125 & 0.527 & 0.021 & -0.003 \\
\bottomrule
\end{tabular*}
\end{minipage}\hfill
\begin{minipage}[t]{0.39\linewidth}
\small
\setlength{\tabcolsep}{1.5pt}
\captionof{table}{Reference-rule agreement over 48 streams (aligned cell). Columns compare the full ledger and value-reassigned histories. Intervals are in Appendix~\ref{app:joint-blackbox-results}.}
\label{tab:joint-main-reference}
\begin{tabular*}{\linewidth}{@{\extracolsep{\fill}}lrrrr@{}}
\toprule
Rule & \multicolumn{2}{c}{Semantic} & \multicolumn{2}{c}{Opaque} \\
\cmidrule(lr){2-3}\cmidrule(l){4-5}
 & \shortstack{Full\\ledger} & \shortstack{Value\\swap} & \shortstack{Full\\ledger} & \shortstack{Value\\swap} \\
\midrule
Mean & 1.000 & 1.000 & 1.000 & 1.000 \\
Sum & 0.635 & 0.615 & 0.635 & 0.615 \\
Count & 0.182 & 0.276 & 0.182 & 0.276 \\
Latest record & 0.203 & 0.229 & 0.203 & 0.229 \\
Best & 0.453 & 0.474 & 0.453 & 0.474 \\
\bottomrule
\end{tabular*}
\end{minipage}
\vspace{-2mm}
\end{table}

\subsection{Order sensitivity can make a memory migration unsafe}
\label{sec:memory-repair}

The previous experiments used unlimited memory. A practical memory pipeline, however, has a writer, a storage budget, and a retrieval step. If restoring the same event log in a different arrival order changes the agent's decisions, the pipeline is order-sensitive and a simple backup--restore migration is unsafe. We test four bounded-memory implementations (FIFO retention, logical-time retention, an identifier-priority reservoir, and action-descriptor bin means), all sharing a 16-slot limit and an eight-slot retrieval interface. Reordering full events preserves their descriptors, outcomes, and logical times, isolating ingestion-order dependence.

The development audit detects the FIFO change before model readout. Reordering changes its serialized summaries and retrieved text in every block, while the other three writers preserve them (Table~\ref{tab:memory-development}). FIFO retention drops the oldest entries when the buffer is full, so a different arrival order changes which entries survive, producing different summaries even from the same event set. Value reassignment changes all writers' decisions, confirming that every pipeline uses outcome values. The audit and endpoint selectors freeze their choices before test histories are generated.

\begin{table}[ht]
\vspace{-5mm}
\centering
\small
\begin{minipage}[t]{0.50\linewidth}
\small
\setlength{\tabcolsep}{1.5pt}
\captionof{table}{Development diagnostics over 16 streams. $U_2$ and $U_6$ average two and six aligned calls. $T_{\rm in}$ and $T_{\rm value}$ are repeat-corrected responses. Changed counts are out of 64 blocks per row.}
\label{tab:memory-development}
\begin{tabular*}{\linewidth}{@{\extracolsep{\fill}}lrrrrr@{}}
\toprule
Memory & $U_2$ & $U_6$ & $T_{\rm in}$ & $T_{\rm value}$ & Changed \\
\midrule
FIFO & 0.674 & 0.675 & 0.516 & 0.672 & 64 \\
Event time & 0.675 & 0.675 & 0.000 & 0.672 & 0 \\
Reservoir & 0.676 & 0.676 & -0.008 & 0.719 & 0 \\
Binned mean & 0.677 & 0.677 & 0.000 & 0.727 & 0 \\
\bottomrule
\end{tabular*}
\end{minipage}\hfill
\begin{minipage}[t]{0.48\linewidth}
\small
\setlength{\tabcolsep}{-0.1pt}
\captionof{table}{Frozen selection: 192 tasks, 48 streams. $U_{\rm A}$, $U_{\rm R}$, and $U_{\rm mix}$ are aligned, reordered, and ingestion-order averages with two fresh calls per condition. Binned-mean rows share a pipeline.}
\label{tab:memory-test}
\begin{tabular*}{\linewidth}{@{\extracolsep{\fill}}lrrrr@{}}
\toprule
Memory & $U_{\rm A}$ & $U_{\rm R}$ & $U_{\rm mix}$ & $T_{\rm in}$ \\
\midrule
FIFO & 0.673 & 0.668 & 0.670 & 0.570 \\
Binned mean (audit) & 0.673 & 0.673 & 0.673 & 0.003 \\
Binned mean (endpoint) & 0.674 & 0.673 & 0.674 & 0.001 \\
No memory & 0.669 & 0.669 & 0.669 & -0.029 \\
\bottomrule
\end{tabular*}
\end{minipage}
\vspace{-2mm}
\end{table}

On new histories and future tasks, FIFO remains order-sensitive and loses utility under reordered ingestion (Table~\ref{tab:memory-test}), while the selected binned-mean pipeline preserves retrieved text. Both selectors choose the same binned-mean pipeline and therefore have similar endpoint utilities. The audit reveals a FIFO migration hazard that endpoint comparison misses: restoring FIFO memory from an unordered backup changes the retained history.

\subsection{Historical score assignments persist through sequential planning}
\label{sec:sequential-agent}

The previous experiments used fixed histories. Here we test whether the replay pattern survives a closed-loop setting where the agent runs fresh experiments. An Astra agent does sequential model development~\citep{huang2024mlagentbench}. Starting from eight measured trials, it tests three new configurations via tool calls (each result enters the conversation before the next choice) and then submits a final pipeline. The 96 available configurations combine four sampling policies, two preprocessors, two classifiers, three regularization levels, and two label budgets. All submissions are sealed before any test score is computed. The configuration ID is the recorded key, validation utility is the outcome, experiment position is the placement, and the sealed final-test score is the external evaluator.

The four conditions share this workflow and the complete initial record set, changing only the initial utility assignments, intact configuration--utility pairs, or configuration IDs. Value and pair use the same permutation. Each task and condition has two fresh trajectories. A randomized-search reference samples three configurations and submits the best observed one. Appendix~\ref{app:sequential-agent} gives the tools, task pools, complete trajectories, and paired task-level intervals.

\begin{table}[ht]
\vspace{-2mm}
\centering
\begin{minipage}[c]{0.43\textwidth}
\small
\setlength{\tabcolsep}{1.5pt}
\begin{tabular}{@{}lrrr@{}}
\toprule
History / ref. & First & Final & $U$ \\
\midrule
Aligned & \textemdash{} & \textemdash{} & 0.618 \\
Value reassigned & 0.708 & 0.875 & 0.359 \\
Pair transport & 0.031 & 0.000 & 0.617 \\
Consistent renaming & -0.042 & -0.021 & 0.619 \\
Random search (aligned) & \textemdash{} & \textemdash{} & 0.610 \\
Random search (value) & \textemdash{} & \textemdash{} & 0.443 \\
\bottomrule
\end{tabular}
\end{minipage}\hfill
\begin{minipage}[c]{0.54\textwidth}
\setlength{\abovecaptionskip}{0pt}
\caption{First and Final give $T_{\rm gr}$ for the first experiment and submitted pipeline, relative to aligned. Each of 48 tasks has two trajectories per condition and three experiment calls per trajectory. $U$ is final-test utility. The random-search reference rows report $U$ only, as First and Final are not applicable. The dash marks an unavailable comparison. Intervals and intermediate choices appear in Appendix~\ref{app:sequential-agent}.}
\label{tab:sequential-main}
\end{minipage}
\vspace{-2mm}
\end{table}

Value reassignment changes both the first experiment and the final submitted pipeline (Table~\ref{tab:sequential-main}). Pair transport and consistent renaming produce much smaller responses. Final-test utility falls under value reassignment despite three fresh experiments, and the randomized-search reference also has lower utility under reassignment.

\subsection{The audit transfers to code debugging}
\label{sec:code-debugging}

To test whether the audit transfers beyond model selection, we ran a controlled code-debugging agent on 24 bug-fixing tasks with twelve candidate patches each and a four-entry diagnostic history. The agent reads the history of previous patch attempts and their test outcomes, then selects and applies patches through tool calls. GPT-6 Astra runs two trajectories per task and condition (192 trajectories, 768 tool events). Table~\ref{tab:code-debugging-main} reports results. Intervals are in Appendix~\ref{app:code-debugging-agent}.

\begin{table}[ht]
\vspace{-5mm}
\centering
\small
\setlength{\tabcolsep}{3pt}
\caption{Code-debugging replay responses on 24 tasks. First and Final are action-disagreement proportions relative to aligned. Hidden $U$ is held-out test utility and $\Delta U$ is its paired difference from aligned.}
\label{tab:code-debugging-main}
\begin{tabular}{lrrrr}
\toprule
Condition & First & Final & Hidden $U$ & $\Delta U$ \\
\midrule
Aligned & \textemdash{} & \textemdash{} & 1.000 & 0.000 \\
Value reassigned & 0.604 & 0.167 & 0.875 & -0.125 \\
Pair transport & 0.042 & 0.000 & 1.000 & 0.000 \\
Consistent renaming & 0.083 & 0.000 & 1.000 & 0.000 \\
\bottomrule
\end{tabular}
\vspace{-3mm}
\end{table}

Outcome reassignment changes the initial debugging choice most often and is the only condition with a negative held-out utility contrast (full intervals are in Appendix~\ref{app:code-debugging-agent}). Moving complete patch--score pairs or renaming the patch namespace leaves the final submission unchanged on these tasks. These results support preserving patch identities and order during code-log migration and checking changes that reassign scores to patches.

\subsection{Cross-model signatures require protocol-specific reading}
\label{sec:cross-model}
Table~\ref{tab:cross-model-summary} summarizes the typed replay responses across four readout protocols spanning two model families. Within each protocol, key reassignment changes choices (action disagreement 0.865--0.996), and value reassignment exceeds pair transport in every row. The four rows use different evidence tiers---archival GPT-5.6 readouts and a frozen Qwen3-8B bridge---so only within-row contrasts share an estimand; the table reports a recurring within-protocol pattern rather than a pooled ranking. Qwen3-8B has the largest pair response (0.154 versus at most 0.017 for the GPT variants). Because the evidence tiers differ, this difference requires protocol-specific verification and does not identify a model-size effect. Full factorial results, namespace controls, and confidence intervals are in Appendices~\ref{app:joint-blackbox-results}, \ref{app:llm-blackbox-transfer}, and~\ref{app:llm-opaque-control}.

\begin{table}[ht]
\vspace{-5mm}
\centering
\small
\caption{Cross-model typed replay responses by protocol. Each row is a separate evidence tier and readout protocol; only within-row contrasts share an estimand, so rows are not pooled or ranked. GPT-5.6 rows use readable arm keys and 48 replicate streams; the Qwen3-8B row uses the frozen action-key namespace on the same stream count. Value exceeds pair in every row, and Qwen3-8B has the largest pair response on its protocol.}
\label{tab:cross-model-summary}
\begin{tabular}{@{}lrrr@{}}
\toprule
\textbf{Readout} & $A_{\rm rekey}$ & $D_A^{\rm value}$ & $D_A^{\rm pair}$ \\
\midrule
GPT-5.6 Luna & $0.865$ & $0.115$ & $0.016$ \\
GPT-5.6 Sol & $0.996$ & $0.325$ & $0.010$ \\
GPT-5.6 Terra & $0.984$ & $0.247$ & $0.017$ \\
Qwen3-8B & $0.946$ & $0.246$ & $0.154$ \\
\bottomrule
\end{tabular}
\vspace{-3mm}
\end{table}

\section{Discussion and Conclusion}

\method{} identifies which relationship in a stored history carries a behavioral effect. It holds the writer, evaluator, and future interface fixed while varying one relationship at a time, producing a dependence profile that endpoint evaluation cannot resolve. The profile supports scientific analysis and migration decisions.

\noindent\textbf{Findings.} Endpoint-matched writers diverge under pair transport despite identical aligned decisions, showing that history arrangement carries behavioral information beyond endpoint accuracy (\S\ref{sec:endpoint-separation}). Sum-based writers satisfy the pair-transport zero law, while prefix-causal writers can violate it (\S\ref{sec:calibration}). On black-box LLM interfaces, outcome reassignment changes decisions while pair transport preserves them, separating score attachment from record order (\S\ref{sec:blackbox-assignment}). Bounded-memory storage exposes ingestion-order sensitivity, and sequential planning plus code debugging show that altered scores can redirect choices and reduce final utility despite fresh measurements (\S\ref{sec:memory-repair}, \S\ref{sec:sequential-agent}, \S\ref{sec:code-debugging}).

\noindent\textbf{Practical guidelines.} Use the corresponding replay cell before each migration: pair transport tests record reordering, Key-slot reassignment tests identifier changes, outcome reassignment tests scores moved across actions, and consistent renaming tests simultaneous relabeling of logs and menus. A nonzero action response flags a compatibility risk, and the paired utility contrast measures its downstream consequence. The framework covers systems whose decisions depend on structured logs with identifiable keys, measured outcomes, and ordered records.

\noindent\textbf{Scope and limitations.} We evaluate four discrete sampling actions and a twelve-candidate code-debugging menu. The audit requires a logged key, an observed outcome, and a declared placement; continuous action spaces require application-specific map families. Hosted calls may drift with provider updates, while frozen prompts and archival protocols limit this source of variation. The current evidence covers these measured interfaces; broader action spaces, longer-horizon trajectories, and continuous-integration workflows remain extensions.

\bibliographystyle{iclr2027_conference}
\bibliography{references}

\appendix

\clearpage
\makeatletter
\let\replaylens@appendixsection\section
\let\replaylens@appendixsubsection\subsection
\let\replaylens@appendixsubsubsection\subsubsection
\replaylens@appendixsection*{Appendix directory}
\renewcommand{\section}[1]{%
\replaylens@appendixsection{#1}%
  \addcontentsline{appx}{section}{\protect\numberline{\thesection}#1}}
\renewcommand{\subsection}[1]{%
\replaylens@appendixsubsection{#1}%
  \addcontentsline{appx}{subsection}{\protect\numberline{\thesubsection}#1}}
\renewcommand{\subsubsection}[1]{%
\replaylens@appendixsubsubsection{#1}%
  \addcontentsline{appx}{subsubsection}{\protect\numberline{\thesubsubsection}#1}}
\makeatother
\begingroup
\setcounter{tocdepth}{2}
\small
\makeatletter
\renewcommand*\l@section[2]{%
  \ifnum \c@tocdepth >\z@
    \addpenalty\@secpenalty
    \addvspace{0.35em \@plus\p@}%
    \setlength{\@tempdima}{1.5em}%
    \begingroup
      \parindent\z@ \rightskip\@pnumwidth
      \parfillskip -\@pnumwidth
      \leavevmode
      \advance\leftskip\@tempdima
      \hskip -\leftskip
      \bfseries #1\normalfont\nobreak\dotfill
      \nobreak\hb@xt@\@pnumwidth{\hss #2\kern-\p@\kern\p@}\par
    \endgroup
  \fi}
\@starttoc{appx}
\makeatother
\endgroup
\noindent\textbf{Reader's guide.} The essential appendices for the central replay claim are ``Replicate and condition specification,'' ``Structural-control details,'' ``Inference and condition diagnostics,'' ``Joint history and action-menu renaming,'' and ``Joint black-box results and recovery provenance.'' The finite-memory, sequential-planning, and code-debugging appendices are essential for the corresponding external-validity experiments. The remaining appendices are supplementary checks covering alternative writer families, representation controls, robustness analyses, and item-level provenance.
\clearpage

\section{Replicate and condition specification}
\label{app:design-details}
\label{app:extended}
\label{app:formal-interface}

The controlled writer comparisons use a hidden-world scientific decision substrate. The executor owns hidden world parameters, outcome noise, and diagnostics. Dynamic controllers see public task metadata and their persistent state before selecting one of four arms, then receive only the selected arm's outcome. \staticpred{} is an explicit privileged control that receives all four development-arm counterfactual means. The secondary \staticselected{} control receives one selected-arm outcome per development task and no counterfactual development outcome. Counterfactual outcomes for all other conditions are executor-side counterfactual reference data.

The controller matrix uses 48 independent replicate streams and 7 conditions: Outcome-Calibrated Recursive Research (OCRR), \noupdate{}, \memory{}, \skill{}, \shuffle{}, \staticpred{}, and \extrasampling{}. Within this controller matrix, OCRR versus \noupdate{} is the sole confirmatory contrast; \extrasampling{} is treated as a secondary efficiency analysis. The separate secondary matrix adds \staticselected{}, \contextucb{}, and \replay{} as structural-control conditions. It contains policy-state comparisons on the same task and outcome sequence. The analysis is defined on the same selected-task trace and counterfactual future for every state comparison.

The primary contrast is a controller-level state-use comparison. In the fixed substrate, adaptation actions and outcomes do not alter future task generation or evaluation. The trace-matched no-state branch applies the OCRR rule at its initial empty state. With no local or global evidence, its deterministic tie rule selects \texttt{replicate}; it therefore isolates an empty initial state from the fixed mixed \noupdate{} controller. Every replay cell preserves the selected-outcome trace, future keys, and future evaluator inputs.

The nine derangements are coupled fixed-label perturbations of~those~traces.

All task, intervention, and fit draw laws are fixed before evaluation. The primary synthetic representation controls use shared within-stream draws, and the value and pair cells use the same blockwise permutation draw. The learned commutative writer uses an order-invariant leave-one-out objective. These conventions fix the intervention coupling while leaving replicate streams as the sampling units. The two cross-policy fits use disjoint 48-stream training and evaluation sets in each direction, use the same order-invariant objective, and freeze the state parameters before either evaluation split.

\noindent\textbf{Organization.} The formal interface below specifies the state boundary, replay cells, readouts,~and~estimands.

\subsection{Formal interface, replay cells, and estimands}

The primary experiment contains 48 replicate streams. Within a stream, each public family--context pair has a development stage and a future stage; the two learning families also have an adaptation stage. The adaptation trace is the object replayed below. The future is held fixed after adaptation, so replay changes the state presented to the future policy without changing the future task distribution or~its~evaluator.

\begin{table}[ht]
\vspace{-4mm}
\centering
\small
\setlength{\tabcolsep}{3pt}
\caption{Complete key--utility map for the replay cells. At slot $t$, the blockwise permutations stay fixed across label maps within a comparison, and public task inputs remain at their original slots.}
\label{tab:typed-cells}
\begin{tabular*}{\linewidth}{@{\extracolsep{\fill}}lcccc@{}}
\toprule
 & \textbf{Aligned} & \textbf{Value} & \textbf{Pair} & \textbf{Key--slot} \\
\midrule
\textbf{Key} $\ell_t^q$ & $k_t^\rho$ & $k_t^\rho$ & $k_{\sigma_{\mathrm p}(t)}^\rho$ & $k_{\sigma_{\mathrm t}(t)}^\rho$ \\
\textbf{Utility} $u_t^q$ & $y_t$ & $y_{\sigma_{\mathrm v}(t)}$ & $y_{\sigma_{\mathrm p}(t)}$ & $y_t$ \\
\bottomrule
\end{tabular*}
\vspace{-2mm}
\end{table}

\noindent\textbf{Typed interface and state assembly.} Let $\mathcal X$ be the public task space, $\mathcal A=\{\mathrm{coverage}, \mathrm{adaptive},\mathrm{replicate},\mathrm{mixed}\}$ the arm space, $\mathcal Z$ the support of executor draws, $\mathcal O$ its structured outcome space, and $\mathcal Y=[0,1]$ the scalar utility space. The common executor and utility projection have types
\begin{equation}
g:\mathcal X\times\mathcal A\times\mathcal Z\to\mathcal O,
\quad h:\mathcal O\to\mathcal Y,
\quad g_R:=h\circ g|_{\mathcal X_R\times\mathcal A\times\mathcal Z_R}.
\label{eq:typed-executor}
\end{equation}
For writer $m$, let $\mathcal S_m$ be its complete persistent state and $\mathcal K_m$ its producer-key space. The shared-key case has $\mathcal K_m=\mathcal A$; a disjoint-key writer uses a producer-key copy $\mathcal K_m=\mathcal K_P$. Its one-step transition and future score have types
\begin{equation}
\mathcal{T}_m:\mathcal S_m\times\mathcal X\times\mathcal K_m\times\mathcal Y\to
\mathcal S_m,
\quad
r_m:\mathcal S_m\times\mathcal X\times\mathcal E_m\to\mathbb R.
\label{eq:typed-writer}
\end{equation}
The future policy is $\pi_m(s,x)=\operatorname{TieBreak}_{\prec_{\rm pol}} \arg\max\nolimits_{a\in\mathcal A}r_m(s,x,\eta_{A,m}(a))$. The structural writers use the lexicographically largest canonical arm on an exact policy-score tie; the fixed $\noupdate{}$ policy always returns \texttt{mixed}, and the stochastic \extrasampling{} policy consumes its fixed per-stream draw. The executor oracle uses the canonical tuple order $(\mathrm{coverage},\mathrm{adaptive}, \mathrm{replicate},\mathrm{mixed})$ and returns the first entry on an exact utility tie, hence \texttt{coverage}. These tie rules are part of the estimand.

The interface also admits distinct producer and future-action namespaces. In the disjoint case, let $\mathcal K_P$ and $\mathcal K_A$ be disjoint copies of $\mathcal A$, with bijections $\phi:\mathcal A\to\mathcal K_P$ and $\chi:\mathcal A\to\mathcal K_A$. Let $\mathcal E_m$ be the common feature space in which the writer and future reader interpret the two bindings. Fixed bindings $\psi_P:\mathcal K_P\to\mathcal E_m$ and $\psi_A:\mathcal K_A\to\mathcal E_m$ obey $\psi_P(\phi(a))=\psi_A(\chi(a))$; the writer receives the producer key and the future reader receives the action key. The shared namespace is the identified special case $\mathcal K_P=\mathcal K_A=\mathcal A$, $\phi=\chi=\operatorname{id}$, and $\psi_P=\psi_A$. Define the producer binding $\eta_{P,m}=\psi_P\circ\phi$ and action binding $\eta_{A,m}=\psi_A\circ\chi$; in the shared case both reduce to the identity binding. The update receives $\kappa_m(a)=\phi(a)$ in the disjoint case and $\kappa_m(a)=a$ in the shared case, while the future reader receives $\eta_{A,m}(a)$. This typed extension matters for representation transport: the protocol follows the declared binding from the selected action to the state and from the state to the future query, so a renamed identifier remains distinguishable from a changed future-action interface.

The ledger state is a product of a shared global component and one component for each family--context block. In canonical stream order, an update changes the active block and the shared component. If $b(R)$ is the last adaptation block whose state feeds future stratum $R$, define $I_{r,R}=\{t:\operatorname{block}(t)\le b(R)\}$ and $T_{r,R}=\max I_{r,R}$; survival uses the prefix after the four learning blocks and has no survival update. Fitted associative and recurrent controls instead use a zero-initialized state per family--context block with parameters shared across blocks.

Let $D_r=((x_t,a_t,y_t))_{t=1}^{T_r}$ denote the full selected adaptation trace in replicate $r$. Index its prefix from $t=1$, initialize $s_0$, and let a state writer have the stationary one-step update
\begin{equation}
 s_t=\mathcal{T}_m(s_{t-1},x_t,\ell_t^q,u_t^q),\quad t=1,\ldots,T_{r,R},\quad
 \hat a_j=\pi_m(s_{T_{r,R}},x_j),
 \end{equation}
where $(\ell_t^q,u_t^q)$ is the typed label--utility pair supplied by cell $q$ with $\ell_t^q\in\mathcal K_m$; for a cross-namespace writer, $\ell_t^q$ is the producer key bound by $\phi$. Write $W_m^{(T_{r,R})}$ for the corresponding prefix composition from the fixed initial state. The future evaluator returns the counterfactual utility of $\hat a_j$; it is not exposed to $\mathcal{T}_m$ or $\pi_m$.

To place all replay cells in one functional, let $\sigma=(\sigma_{r,b})_{r,b}$ denote the collection of fixed within-block slot permutations and let $q$ denote the state-boundary cell. For a slot $t$ in block $b$ of replicate $r$, write $\sigma(t)=\sigma_{r,b}(t)$. The label--utility pair supplied to the writer is written below in shared-key notation; for a disjoint producer-key space, apply $\kappa_m$ to the first coordinate before the update:
\begin{equation}
(\ell_t^q,u_t^q)=
\begin{cases}
(\rho(a_t),y_t), & q=\mathrm{aligned},\\
(\rho(a_t),y_{\sigma(t)}), & q=\mathrm{value},\\
(\rho(a_{\sigma(t)}),y_{\sigma(t)}), & q=\mathrm{pair},\\
(\rho(a_{\sigma(t)}),y_t), & q=\mathrm{time},
\end{cases}
\quad
 S_{r,R}^{m,q}(\rho,\sigma)=W_m^{(T_{r,R})}\!\left(\big((x_t,\ell_t^q,u_t^q)\big)_{t\in I_{r,R}}\right).
\label{eq:cell-functional}
\end{equation}
For a declared protocol $\mathcal P$, the action-valued replay signature of writer $W_m$ is the family of future action tables
\begin{equation}
\mathsf S_{\mathcal P}(W_m)=
\left\{a^+_{r,R,m,q}(\cdot;\rho,\sigma)\right\}_{(r,R,q,\rho,\sigma)\in\mathcal P}.
\label{eq:full-replay-signature}
\end{equation}
The declared protocol includes the aligned baseline $\rho=\mathrm{id}$; the nine nonidentity maps in $\mathcal D_4$ are the replay interventions. Thus the ordinary endpoint is a member of the signature, while structural expectations below average only over $\mathcal D_4$. Here $a^+_{r,R,m,q}(x;\rho,\sigma)= \pi_m(S_{r,R}^{m,q}(\rho,\sigma),x)$ is the future action function induced by the terminal state in Eq.~\eqref{eq:cell-functional}. The scalar readouts below are reported projections of this family. We write $W_m\equiv_{\mathrm{end}}W_{m'}$ when the aligned action tables agree pointwise on the declared future support. This pointwise endpoint notion is stronger than equality of an aggregate future score, which may hide task-level differences. The four cells use the same realized trace, future task set, evaluator, and outer map family. Their declared permutation laws may differ, while each law is fixed before the readout. The prefix index $T_{r,R}$ ensures that the future for $R$ never receives a later block's outcome. The task input $x_t$ remains in its original slot. Pair transport therefore has a pair-only interpretation under the commutative conditions stated below; for a general sequential writer it is a joint record, position, and exposure intervention. Let $U^{m,q}_{r,R}(\rho,\sigma)$ be future utility, $A^{m,q}_{r,R}(\rho,\sigma)$ the fraction of future actions that differ between the identity and mapped states, and $O^{m,q}_{r,R}(\rho,\sigma)$ executor-relative oracle agreement. With future randomness coupled within a replicate, orient the three readouts as
\begin{equation}
\resizebox{0.96\linewidth}{!}{$
\begin{aligned}
\bigl(\delta_U^{m,q},\delta_A^{m,q},\delta_O^{m,q}\bigr)(\rho,\sigma)
&=\bigl(U^{m,q}_{r,R}(\mathrm{id},\sigma)-U^{m,q}_{r,R}(\rho,\sigma),
A^{m,q}_{r,R}(\rho,\sigma),
O^{m,q}_{r,R}(\rho,\sigma)-O^{m,q}_{r,R}(\mathrm{id},\sigma)\bigr),\\
\overline{\Gamma}_y^{m,q}&=
\frac{1}{48}\sum\nolimits_{r=1}^{48}\frac{1}{|\mathcal R|}
\sum\nolimits_{R\in\mathcal R}\frac{1}{9}\sum\nolimits_{\rho\in\mathcal D_4}
\mathbb{E}_{\sigma\sim P_q}\left[\delta_y^{m,\mathrm{aligned}}(\rho,\sigma)-
\delta_y^{m,q}(\rho,\sigma)\right].
\end{aligned}
$}
\label{eq:formal-gamma}
\end{equation}
For the primary synthetic protocol, $\mathcal R$ is the set of six family--context strata and $P_q$ is the declared within-block permutation law for cell $q$; the pair-preserving and value-null cells use the same paired law, $P_{\mathrm{pair}}=P_{\mathrm{value}}$. More explicitly, let $\mathcal B$ be the four learning blocks per stream, each with 48 selected slots, and let $\mathfrak S_{48}$ and $\mathfrak D_{48}=\{\sigma\in\mathfrak S_{48}:\sigma(t)\ne t\ \forall t\}$ denote the permutation and derangement supports. The time placebo uses the product of uniform laws on $\mathfrak S_{48}$, while the value and pair cells use the product of uniform laws on $\mathfrak D_{48}$, with the pair cell sharing the value-cell draw. The outer map $\rho$ is averaged uniformly over the nine fixed-point-free maps in $\mathcal{D}_4$. The formal estimand averages over this declared law, while the numerical suite fixes one permutation per replicate--block according to each cell's declared law before computing the readout. We write the resulting collection as $\sigma^\star$. The reported cluster summaries and fixed-draw estimator condition on $\sigma^\star$. The five refit summaries vary parameter initialization under the order-invariant objective, not this permutation draw. For an observed collection $\sigma^\star$, let $\widehat{\delta}_{y,r,R,x}^{m,q} (\rho,\sigma^\star)$ denote the readout difference for future task $x$ and let $\widehat{\delta}_{y,r,R}^{m,q}(\rho,\sigma^\star)$ denote its average over the future tasks in stratum $R$. The reported fixed-draw estimator is
\begin{equation}
\widehat{\Gamma}_{y}^{m,q}(\sigma^\star)=
\frac{1}{48}\sum\nolimits_{r=1}^{48}\frac{1}{6}\sum\nolimits_R\frac{1}{9}
\sum\nolimits_{\rho\in\mathcal{D}_4}
\left[\widehat{\delta}_{y,r,R}^{m,\mathrm{aligned}}(\rho,\sigma^\star)-
\widehat{\delta}_{y,r,R}^{m,q}(\rho,\sigma^\star)\right].
\label{eq:locked-gamma}
\end{equation}
The interval resamples complete replicate clusters with $\sigma^\star$, fitted parameters, maps, and stratum weights held fixed; it is a percentile interval over replicate-cluster bootstrap draws. The utility and oracle readouts are consequence measures within the declared substrate; semantic interpretation of arm names is outside these readouts. The ordered sequence in Eq.~\eqref{eq:cell-functional} matters for order-sensitive writers, whereas the commutative proposition below assumes a slot-independent initial state, position-free feature map, label-only outer map, and a readout that depends only on the terminal state and~held-fixed~future.

\noindent\textbf{Slot-law variants.} The laws above describe the primary synthetic protocol. Within each public measured-outcome branch, value and pair share a 24-slot derangement, and time uses a separately seeded derangement. The structural and recurrent branches generate their assignments separately. The changed-substrate selected-only calibration shares one 48-slot derangement across value, pair, and time. The textual-memory experiments generate each cell's 24-slot derangement with a condition-specific seed formula; these formulas coincide across cells for replicate 0, and do not enforce shared value--pair order across the remaining streams. Their archived time-labelled implementation moves complete pairs, as documented in Appendix~\ref{app:textual-time-audit}; it does not implement the key-only time cell above. Their reported responses compare each transformed cell with the aligned cell under the identity key map. These branch-specific estimates condition on their own slot draws and are reported separately. A shared-order pair--value interpretation requires $\sigma_{\mathrm v}=\sigma_{\mathrm p}$; the complete-pair zero law itself holds for every permitted pair permutation under its stated~writer~assumptions.

In the primary synthetic protocol, the value interaction is the primary descriptive structural estimand and is separate from the transfer endpoint; it does not determine transfer efficacy. The value-null cell receives the same labels and update slots as the aligned writer, but the selected utilities are assigned to slots by a within-block derangement. The time placebo changes the stored labels while retaining the utility sequence. The pair-preserving cell uses the same utility derangement as the value null and carries each producing label with its utility. Thus the pair and value cells share utility order, and differ in whether the label--utility pairing is preserved. For an order-sensitive writer, this crossed construction remains a joint pairing-and-exposure comparison; the commutative writer provides the cleaner pairing-specific check.

The trace-matched state-readout estimand separates this structural measurement from the closed-loop controller comparison. Let $W_0$ use the same OCRR choice rule and policy clock as $W_{\mathrm{OCRR}}$, set its persistent state to the initial empty state, and discard every adaptation outcome before the fixed future. This is a distinct empty-state control, not the fixed mixed $\noupdate{}$ rule. With no local or global evidence, the deterministic tie rule selects \texttt{replicate}; this is the action used by $W_0$ at the future boundary. We include the policy-clock value in the future-policy input $x$. Thus $W_0$ is the same OCRR rule with its state writes removed, whereas $\noupdate{}$ is a separate fixed-action controller. The coupling construction preserves the selected-outcome trace and future inputs in every cell; replay makes no future-state update. For a future set $\mathcal{X}_r^+$, write $U(W,D_r)$ for future utility and $O(W,D_r)$ for executor-relative oracle agreement. Here $\pi_W(x;D_r)$ denotes the future action after writer $W$ has processed $D_r$. Define the direct action mismatch
\begin{equation}
D_A(W_1,W_0;D_r)=\frac{1}{|\mathcal{X}_r^+|}
\sum\nolimits_{x\in\mathcal{X}_r^+}
\mathbf{1}\!\left[\pi_{W_1}(x;D_r)\ne\pi_{W_0}(x;D_r)\right].
\label{eq:direct-action-mismatch}
\end{equation}
With these readouts,
\begin{equation}
\begin{aligned}
\Theta_U&=\mathbb{E}_{r}\!\left[U(W_{\mathrm{OCRR}},D_r)-U(W_0,D_r)\right],\\
\Theta_A&=\mathbb{E}_{r}\!\left[D_A(W_{\mathrm{OCRR}},W_0;D_r)\right],\\
\Theta_O&=\mathbb{E}_{r}\!\left[O(W_{\mathrm{OCRR}},D_r)-O(W_0,D_r)\right].
\end{aligned}
\end{equation}
This estimand is conditional on the observed adaptation trace; the closed-loop OCRR--\noupdate{} comparison remains a separate controller comparison. The arm-omitting state supplies a formal negative control for $\Gamma$. If a writer's update and future rule depend on each block only through its utility multiset and public family--context key, a within-block utility permutation leaves its terminal state unchanged. Its aligned and matched-null readouts are therefore identical and $\Gamma=0$. The observed zero response of the label-invariant control calibrates this implication before interpreting a nonzero response of an arm-indexed writer.

\noindent\textbf{Commutative paired-state proposition.} Consider a writer whose terminal state on one complete block is $h=h_0+\sum\nolimits_t\varphi(\ell_t,u_t)$, where $\ell_t$ is the mapped producer label and $u_t$ is the exposed utility. The initial state is slot-independent, and $\varphi$ has no dependence on $x_t$, block position, or time. The future readout depends only on $h$, the held-fixed future, and a coupled tie rule; the outer map acts only on the producer label. The statement is blockwise, and a global writer must combine complete block states additively for the same argument. The pair intervention covers every selected slot in that block. Moving complete $(\ell_t,u_t)$ pairs between slots leaves $h$ and every future readout unchanged. Define pairing sensitivity at readout $y$ by a nonzero value of $\delta_y^{m,\mathrm{aligned}}-\delta_y^{m,\mathrm{value}}$ under the declared permutation law. A changed multiset of feature terms alone is insufficient: the terminal sum and the future readout must also change. Under this explicit state-and-readout condition, the aligned--pair-preserving difference is exactly zero and the aligned--value-null interaction is a conditional pairing contrast. A nonzero pair residual therefore diagnoses a violation of at least one listed condition, such as order- or position-dependence, a non-additive global combination, an uncoupled readout input, or incomplete pair transport.

\noindent\textbf{Proof of finite-support observational nonidentification.} Let $\mathcal C$ be a controller class that contains stateful writers and is closed under finite-support, stream-conditioned lookup writers. For one replicate stream and one adaptation prefix, let $W\in\mathcal C$ and let $f_W(x)=\pi_W(W(D_{r,R}),x)$ be the future action induced by $W$ on a fixed adaptation prefix, with finite future support $\mathcal X^+_{r,R}$. Closure under lookup gives $W_0\in\mathcal C$ with constant state $s_0$ and $\pi_0(s_0,x)=f_W(x)$ for every $x\in\mathcal X^+_{r,R}$. The aligned future actions and scalar future-value score of $W_0$ therefore equal those of $W$. Since the complete aligned action table is equal, every endpoint statistic that is a function of that table and the fixed executor is equal as well, while every label replay leaves $W_0$ unchanged. Whenever $W$ has a nonzero replay response, the two members of the same class are observationally equivalent for the aligned endpoint and separated by replay. This proves finite-support score nonidentification. The closure is invoked separately for each observed stream and prefix: $W_0$ may be a stream-conditioned support lookup, and the theorem does not assert that one lookup table is learned across replicates. It makes no claim about unseen futures, where coverage determines the resolution of the~state-boundary~readout. Here and below, ``endpoint'' denotes the ordinary identity endpoint $q=\mathrm{aligned}$, $\rho=\mathrm{id}$, with no replay intervention. The replay-derived action response is a separate readout of the same writer.

\noindent\textbf{Proof of strict refinement of aligned endpoint equivalence.}  If two writers are equivalent under $\mathcal P$, their full action-valued signatures agree on every declared cell. In particular, their ordinary identity-endpoint action tables agree pointwise, so $W\equiv_{\mathrm{end}}W'$. For the reverse containment, choose the stateful writer and its support lookup from the finite-support construction. They agree on the ordinary endpoint $(q=\mathrm{aligned},\rho=\mathrm{id})$ and hence on every endpoint statistic computed from this identity action table, while the stateful writer has a nonzero replay response and the lookup has zero response. Their full replay signatures therefore differ, proving the strict refinement of pointwise~endpoint~equivalence.

\noindent\textbf{Proof of the support-separation criterion.}  For each $\tau\in\mathcal P$, let $\mathcal O_\tau$ be the terminal-state orbit and let $F_\tau(s)=[x\mapsto\pi(s,x)]_{x\in\mathcal X^+_\tau}$. Assume that $F_\tau$ is injective on $\mathcal O_\tau$. Hence $S_\tau^W=S_\tau^{W'}$ for every $\tau$ gives equal action tables and therefore equal indexed signatures. Conversely, equal signatures give $F_\tau(S_\tau^W)=F_\tau(S_\tau^{W'})$ for each $\tau$; injectivity then gives $S_\tau^W=S_\tau^{W'}$. The premise is stronger than that used by the empirical assay: when it fails, the action signature remains observable, but a zero action contrast does not imply terminal-state equality.

\subsection{Structural interaction readouts}
\label{app:interaction-readouts}

Let $\widetilde g(x,a,z_x)$ evaluate utility with a shared draw $z_x$, and let $a^*(x)$ maximize this utility under a fixed tie rule. Mean utility and oracle agreement are
\begin{equation}
U_q(\rho)=\frac{1}{|X^+|}\sum\nolimits_{x\in X^+}\widetilde g(x,a_q^\rho(x),z_x),\quad
O_q(\rho)=\frac{1}{|X^+|}\sum\nolimits_{x\in X^+}\mathbf{1}[a_q^\rho(x)=a^*(x)].
\label{eq:main-consequence-readouts}
\end{equation}
Direct consequence contrasts are $\Delta U_q(\rho)=U_q(\rho)-U_{\mathrm{aligned}}(\rho)$ and $\Delta O_q(\rho)=O_q(\rho)-O_{\mathrm{aligned}}(\rho)$; positive values indicate higher utility and oracle agreement, respectively.

For structural rekey contrasts, set $\delta_U^q=U_q(\mathrm{id})-U_q(\rho)$, $\delta_A^q=A_q(\rho)$, and $\delta_O^q=O_q(\rho)-O_q(\mathrm{id})$. Positive $\delta_U^q$ denotes a utility loss under rekeying, whereas positive $\delta_O^q$ denotes increased oracle agreement. Write $P$ and $V$ for pair and value. At the realized intervention draws, the aligned-minus-condition and pair-minus-value interactions are, for $y\in\{U,A,O\}$,
\begin{equation}
\widehat{\Gamma}_y^q=\left\langle\delta_y^{\mathrm{aligned}}-\delta_y^q\right\rangle,\quad
\widehat{\Pi}_y^{P-V}=\left\langle\delta_y^{\mathrm{pair}}-\delta_y^{\mathrm{value}}\right\rangle.
\label{eq:main-interactions}
\end{equation}
Positive $\widehat{\Gamma}_A^q$ means that rekeying changes more decisions under aligned records than under condition $q$; positive $\widehat{\Pi}_A^{P-V}$ means that this response is larger under pair transport than under value reassignment. A zero action interaction can coexist with a nonzero direct response: if value reassignment switches the winning key while both conditions have rekey response one, then $\widehat{\Gamma}_A^{\mathrm{value}}=0$ and $D_A^{\mathrm{value}}(\mathrm{id})=1$. The primary synthetic $\widehat{\Pi}_y^{P-V}$ contrasts use $\sigma_{\mathrm v}=\sigma_{\mathrm p}$, holding utility order fixed. Brackets average paired responses over the nine maps and six equally weighted family--context strata within each stream, then equally over the 48 streams in the primary synthetic protocol. Structural confidence intervals resample whole streams with writer parameters and intervention draws fixed; they quantify stream variation conditional on those draws. Action and oracle rates and their differences use the proportion scale. Sign and exact-zero claims use unrounded results. Appendix~\ref{app:formal-interface} gives the full indexing and distinguishes fixed-draw estimates from expectations over the permutation law.

\subsection{Observation-map boundary}
\label{app:observation-map-boundary}

Let $D$ be the realized typed stream and let $a_W^+(x;D)$ denote the future action of writer $W$ on the original aligned stream. For any statistic $\phi$ of the aligned future action table and its held-fixed executor readout, define
\begin{equation}
\mathcal E_\phi(W;D)=
\phi\!\left(\{a_W^+(x;D)\}_{x\in\mathcal X^+},
\{g(x,a_W^+(x;D),z_x)\}_{x\in\mathcal X^+}\right).
\label{eq:aligned-observation-map}
\end{equation}
The endpoint, future utility, and executor-relative oracle summaries used in the assay are members of this aligned observation family. The replay map for a cell $q$ instead evaluates the writer on the transformed stream $D_q$ while holding $\mathcal X^+$, $g$, and $z_x$ fixed. Here $D_q$ includes the declared map $\rho$ and slot draw $\sigma$, suppressed in $\mathcal R_q$ for brevity:
\begin{equation}
\mathcal R_q(W;D)=\{a_W^+(x;D_q)\}_{x\in\mathcal X^+}.
\label{eq:replay-observation-map}
\end{equation}
For the endpoint-matched pair, the identity action tables agree, while at least one declared replay cell and map separate the writers:
\begin{equation}
\mathcal E_\phi(W^{\mathrm R};D)=
\mathcal E_\phi(W^{\mathrm L};D)\quad\text{for every }\phi,
\quad
\mathcal R_q(W^{\mathrm R};D)\ne
\mathcal R_q(W^{\mathrm L};D).
\label{eq:observation-map-separation}
\end{equation}
\subsection{Action margins and finite-query resolution}
\label{app:margin-resolution}

Suppose a deterministic readout chooses the largest score $f_a(S,x)$, with a fixed tie rule. Write $D_A(S,S')$ for the fraction of changed actions on $X^+$. Let $a_x$ be the aligned winner and let $m_x=f_{a_x}(S,x)-\max_{b\ne a_x}f_b(S,x)$ be its margin. Assume each action score is $L_x$-Lipschitz along the state perturbation under consideration. For $S'=S+\varepsilon H$, the change in any winner--competitor score difference has magnitude at most $2L_x|\varepsilon|\lVert H\rVert$. Consequently,
\begin{equation}
D_A(S,S')\leq \frac{1}{|X^+|}\sum\nolimits_{x\in X^+}
\mathbf{1}[m_x\leq 2L_x|\varepsilon|\lVert H\rVert].
\label{eq:margin-resolution}
\end{equation}
The inequality follows by applying the Lipschitz bound to both action scores: if the aligned margin exceeds that bound, its winner cannot change. With strictly positive margins at a fixed baseline $S$ and uniform finite Lipschitz constants in a neighborhood of $S$, a sufficiently small perturbation along a fixed direction produces zero action response on the finite query set. For the family $S_\varepsilon(D)=S_{\rm pair}(D)+\varepsilon H_{\rm order}(D)$, aligned and pair-replayed states converge to the same order-invariant state as $\varepsilon\to0$. If all query margins at that common zero-order state are strictly positive and the scores are locally Lipschitz there, both states retain its actions for sufficiently small $|\varepsilon|$. Positive margins at each nonzero $\varepsilon$ alone do not give this guarantee. This is a resolution bound; it supplies no empirical estimate of the margins or perturbation sizes in a hosted model.

\subsection{Endpoint-matched finite-support witness}
\label{app:endpoint-matched-witness}

The finite-support construction is instantiated separately within each replicate stream of the fixed primary trace. The relation writer is a prefix-causal commutative accumulator with one coordinate per producer key,
\begin{equation}
s_{t,a}=s_{t-1,a}+(u_t-1/2)\,\mathbf{1}[\ell_t=a],\quad
\pi(s)=\operatorname{TieBreak}_{\prec_{\rm src}}
\arg\max\nolimits_{a\in\mathcal A}s_a .
\label{eq:endpoint-witness-writer}
\end{equation}
It receives the selected outcome at each adaptation slot and is evaluated after the adaptation prefix. The general future input $x$ may contain public metadata such as a policy clock. This endpoint-matched witness instantiates the reduced future query $\tilde{x}=(f,c)$ given by the family--context key, and passes this same reduced query to both writers. The support lookup therefore has six entries per replicate, one for each family--context stratum; the future task identifier indexes only the held-fixed executor outcome and is not a lookup key. In the general finite-support construction, a lookup requires one value for every distinct public query on the declared support. The six-entry implementation is thus a property of this witness interface, not a restriction on the theorem. The support lookup has constant state, receives no adaptation outcome, and is frozen before any rekey replay. Both writers use the same future task set, counterfactual executor table, first-entry source-order tie rule $(\mathrm{coverage},\mathrm{adaptive},\mathrm{replicate},\mathrm{mixed})$, and nine nonidentity producer-key maps. This local tie rule is coupled for the two witness writers; the primary ledger's lexicographically-largest tie rule is a separate controller convention. On the aligned identity cell, every future action selected by the relation writer is copied by the support lookup, so the induced future utility and executor-relative oracle agreement also agree. Under a nonidentity map, the relation writer receives a different producer-key relation and the lookup remains unchanged. The direct action response is therefore expected to separate the writers even though the aligned endpoint cannot. All 48 replicate streams, six future strata, 48 future tasks per stratum, and nine maps are evaluated; maps and future tasks remain paired within replicate streams.

\begin{table}[ht]
\vspace{-4mm}
\centering
\small
\caption{Endpoint-matched finite-support witness. The aligned checks compare the two writers on 13,824 future decisions. Replay responses average over 432 paired replicate--map summaries, with the mapped-minus-identity utility change shown for the same future support. Action and oracle quantities are proportions; utility quantities are normalized units. The relation writer changes all 432 map--stream pairs, whereas the support lookup changes none. The support lookup is frozen before replay and has no adaptation state.}
\label{tab:endpoint-matched-witness}
\begin{tabular}{@{}lrr@{}}
\toprule
Quantity & Relation writer & Support lookup \\
\midrule
Aligned action mismatch (proportion) & 0.000 & 0.000 \\
Maximum aligned utility gap & 0.000 & 0.000 \\
Maximum aligned oracle gap (proportion) & 0.000 & 0.000 \\
Mean replay action response (proportion) & 0.666 & 0.000 \\
Mean replay utility change & $-0.164$ & 0.000 \\
\bottomrule
\end{tabular}
\vspace{-2mm}
\end{table}

The exact aligned match and complete zero response of the support lookup instantiate the finite-support construction, while the relation writer provides replay separation on the declared support. This construction-level witness uses the per-stream closure in the theorem; the lookup is defined on the declared support, establishing the stated finite-support nonidentification result.

\subsection{Diagnostic dominance under endpoint matching}
\label{app:diagnostic-dominance}

We next compare two writers matched on the state-on--state-off endpoint contrast. Let $W^{\mathrm R}$ denote the relation writer and let $q_{\mathrm{on}}(f,c)$ be its aligned future action for family--context key $(f,c)$. Let $q_{\mathrm{off}}(f,c)$ be the action selected by the same writer from its initial state. The support lookup is defined on the same public keys by $L_{\mathrm{on}}(f,c)=q_{\mathrm{on}}(f,c)$ and $L_{\mathrm{off}}(f,c)=q_{\mathrm{off}}(f,c)$. It receives no adaptation outcome and keeps both tables fixed throughout replay. Consequently, the two writers have the same aligned action table and the same conventional endpoint contrast,
\begin{equation}
\mathcal C(W)=\left(\Delta A_{\mathrm{on-off}},\Delta U_{\mathrm{on-off}},\Delta O_{\mathrm{on-off}}\right),\quad
\mathcal C(W^{\mathrm R})=\mathcal C(L),
\label{eq:diagnostic-dominance-match}
\end{equation}
Here $\Delta A_{\mathrm{on-off}}=d(a_{\mathrm{on}},a_{\mathrm{off}})$, $\Delta U_{\mathrm{on-off}}=U_{\mathrm{on}}-U_{\mathrm{off}}$, and $\Delta O_{\mathrm{on-off}}=O_{\mathrm{on}}-O_{\mathrm{off}}$ use the common future task set, executor, and coupled draw. The evaluation mode selects one of the two lookup tables fixed before replay; no adaptation record changes either table. The equality is checked at the stream level before replay signatures are decoded.

\noindent\textbf{Extension to endpoint and state-use equivalence.} Define $W\equiv_{\mathrm{end+off}}W'$ when their identity-endpoint action tables and $\mathcal C$ contrasts agree on the declared finite support. Appending the replay profile preserves both measurements, so equality of the augmented observation implies $\equiv_{\mathrm{end+off}}$. In the two-mode lookup construction above, the relation writer and lookup are $\equiv_{\mathrm{end+off}}$, while the former has a nonzero producer-key response and the lookup has zero response. Their augmented observations differ. This establishes strict refinement for a class containing these frozen two-mode lookups, conditional on the fixed trace and future support. The test separates information used for matching from information used for classification. The writer labels are permuted after the two endpoint matches are formed. A fixed decoder calls a writer replay-sensitive when its mean direct action response on maps 1--4 and future indices 0--23 exceeds zero; otherwise it calls the writer replay-invariant. The prediction is then evaluated on maps 5--9 and future indices 24--47. No endpoint value, state-on--state-off contrast, utility response, oracle response, or opened writer name enters the decoder. Across 48 replicate streams, the two aligned future tables have zero action mismatch and zero utility and oracle gaps. Their state-on--state-off contrasts are identical: 37.153\% of future actions change, the utility change is $+0.008$, and the oracle change is $+2.901$ percentage points. The calibration block gives action responses of 66.667\% and 0.000\% for the relation writer and lookup, respectively; the disjoint holdout gives 66.528\% and 0.000\%. The blinded decoder therefore classifies both writers correctly on calibration and predicts both held-out classes correctly. The complete block-level readouts are in Table~\ref{tab:diagnostic-dominance-appendix}. The test holds a conventional state-use endpoint fixed while replay separates the observable relation class. The result is conditional on the declared writer interface and future support.

\begin{table*}[ht]
\vspace{-4mm}
\centering
\small
\caption{Held-out replay signatures in the diagnostic-dominance test. The relation writer and support lookup are matched on the aligned endpoint and on the state-on--state-off endpoint contrast before labels are opened. Calibration uses maps 1--4 and future indices 0--23; holdout uses maps 5--9 and future indices 24--47. $A$ is changed-action proportion, $U$ is mapped-minus-identity future utility in normalized units, and $O$ is mapped-minus-identity executor-relative oracle-agreement difference in proportion units. Subscripts $R$ and $L$ denote the relation and lookup writers. The final column is decoder accuracy as a proportion.}
\label{tab:diagnostic-dominance-appendix}
\begin{tabular*}{\textwidth}{@{\extracolsep{\fill}}lrrrrrrrrr@{}}
\toprule
\textbf{Block} & \textbf{Maps} & \textbf{Future} & $\boldsymbol{A_R}$ & $\boldsymbol{A_L}$ & $\boldsymbol{\Delta U_R}$ & $\boldsymbol{\Delta U_L}$ & $\boldsymbol{\Delta O_R}$ & $\boldsymbol{\Delta O_L}$ & $\boldsymbol{B_{\mathrm{dec}}}$ \\
\midrule
Calibration & 4 & 27{,}648 & 0.667 & 0.000 & $-0.180$ & 0.000 & $-0.297$ & 0.000 & 1.000 \\
Holdout & 5 & 34{,}560 & 0.665 & 0.000 & $-0.151$ & 0.000 & $-0.268$ & 0.000 & 1.000 \\
\bottomrule
\end{tabular*}
\vspace{-2mm}
\end{table*}


The finite-support construction gives the corresponding observational nonidentification result. The Lookup writer stores the aligned action for every future query, while the Relation writer reconstructs state from the records. A second set of lookup decisions matches the empty-state actions. Both writers consequently have the same endpoint and memory-on/off measurements (Table~\ref{tab:content-endpoint}), while producer-key replay changes 66.528\% of Relation-writer holdout actions and leaves Lookup unchanged. This construction states why endpoint evidence alone cannot resolve the dependence; the independently fitted pairs show the gap without copying an endpoint action table.

\begin{table*}[ht]
\vspace{-4mm}
\centering
\small
\caption{Endpoint-matched writers. $A_0$ and $U_0$ compare aligned actions and utility between writers. $A_{\varnothing}$ is written-versus-empty action disagreement; $U_{\varnothing}$ and $O_{\varnothing}$ are written-minus-empty utility and oracle agreement. $A_{\rm cal}$ and $A_{\rm ho}$ report rekey responses for Relation/Lookup and pair responses for Recurrent/Additive; $B_{\rm dec}$ is decoder accuracy. Action and oracle quantities are proportions; utility is normalized. Relation/Lookup uses 48 streams; Recurrent/Additive averages five fitted pairs. Bold and underlined entries distinguish the paired writers.}
\label{tab:content-endpoint}
\begin{tabular*}{\textwidth}{@{\extracolsep{\fill}}lrrrrrrrr@{}}
\toprule
\textbf{Writer} & $A_0$ & $U_0$ & $A_{\varnothing}$ & $U_{\varnothing}$ & $O_{\varnothing}$ & $A_{\rm cal}$ & $A_{\rm ho}$ & $B_{\rm dec}$ \\
\midrule
Relation & 0.000 & 0.000 & 0.372 & 0.008 & 0.029 & \textbf{0.667} & \textbf{0.665} & 1.000 \\
Lookup & 0.000 & 0.000 & 0.372 & 0.008 & 0.029 & \underline{0.000} & \underline{0.000} & 1.000 \\
\midrule
Recurrent & 0.000 & 0.000 & 0.750 & $-0.001$ & $-0.004$ & \textbf{0.675} & \textbf{0.682} & 1.000 \\
Additive & 0.000 & 0.000 & 0.750 & $-0.001$ & $-0.004$ & \underline{0.000} & \underline{0.000} & 1.000 \\
\bottomrule
\end{tabular*}
\vspace{-2mm}
\end{table*}

\subsection{Independent learned endpoint pair}
\label{app:independent-endpoint-pair}

We next test the same boundary with two writers that are fitted independently and never exchange endpoint decisions. The adaptation sequence contains four producer keys and 32 selected slots, with the 32 slots presented in a fixed source order during fitting and evaluation. For a sequence $D=((\ell_t,u_t))_{t=1}^{32}$, the common endpoint target is the winner
\begin{equation}
a^\dagger(D)=\operatorname*{arg\,max}_{a\in\mathcal A}
\sum\nolimits_{t=1}^{32}(u_t-1/2)\mathbf 1[\ell_t=a],
\label{eq:independent-endpoint-target}
\end{equation}
The finite-support margin and fitting rule are fixed before evaluation. Each sequence contains four producer keys in four contiguous blocks of eight slots; the winning block utilities are drawn independently from $\mathsf{Uniform}(0.78,0.94)$ and the other blocks from $\mathsf{Uniform}(0.06,0.22)$. The target vector assigns $0.95$ to the winner and $0.05$ to every other key, with winner classes balanced within each training stream. The additive writer uses the learned readout $s_a^{\mathrm{add}}=q_a^{\mathsf T}h^{\mathrm{add}}$ over the exact key-specific sum $h^{\mathrm{add}}_a=\sum\nolimits_t(u_t-1/2)\mathbf 1[\ell_t=a]$. The recurrent writer instead uses
\begin{equation}
h_t^{\mathrm{rec}}=\tanh\!\left(W h_{t-1}^{\mathrm{rec}}+p_{\ell_t}
+(u_t-1/2)v_{\ell_t}\right),\quad
s_a^{\mathrm{rec}}=(q_a^{\mathrm{rec}})^{\mathsf T}h_{32}^{\mathrm{rec}}.
\label{eq:independent-endpoint-recurrent}
\end{equation}
Both writers are fitted to this common endpoint target from independent 512-sequence streams for each writer and for each of five pair fits; no endpoint action table is shared between the fits. The recurrent state has dimension 16, starts at zero, and uses Gaussian parameter initialization with scale $0.05$; the additive readout is initialized with scale $0.08$. Both fits use full-batch binary cross-entropy for exactly 4800 updates, learning rate $0.015$, $\ell_2$ coefficient $10^{-4}$, and a global gradient-norm cap of $5$; the first- and second-moment coefficients are $0.9$ and $0.999$ with numerical constant $10^{-8}$. The four recurrent restart offsets are prespecified as $\{0,1009,2017,3023\}$; the candidate with the lowest final training loss is retained before evaluation readout. There is no early stopping. Final training binary cross-entropy rounds to 0.199 for each of the ten retained writers. The records do not contain a separate held-out target-accuracy measure or a target-quality gate; the declared evaluation gate tests agreement between the fitted endpoints.

\begin{table*}[ht]
\vspace{-4mm}
\centering
\small
\caption{Per-fit endpoint contrasts for the independently learned pair. $A_0$ and $A_{\varnothing}$ are action-disagreement proportions; $U_{\varnothing}$ is a normalized utility contrast; and $O_{\varnothing}$ is an oracle-agreement contrast reported as a proportion. Values are rounded to three decimals; nonzero claims use unrounded values. The five fits are paired by fit index.}
\label{tab:independent-endpoint-pair-endpoint}
\begin{tabular*}{\textwidth}{@{\extracolsep{\fill}}rrrrrrrrr@{}}
\toprule
 & \multicolumn{4}{c}{\textbf{Additive}} & \multicolumn{4}{c}{\textbf{Recurrent}} \\
\cmidrule(lr){2-5}\cmidrule(lr){6-9}
\textbf{Fit} & $A_0$ & $A_{\varnothing}$ & $U_{\varnothing}$ & $O_{\varnothing}$ & $A_0$ & $A_{\varnothing}$ & $U_{\varnothing}$ & $O_{\varnothing}$ \\
\midrule
1 & 0.000 & 0.750 & 0.000 & 0.001 & 0.000 & 0.750 & 0.000 & 0.001 \\
2 & 0.000 & 0.750 & $-0.002$ & $-0.008$ & 0.000 & 0.750 & $-0.002$ & $-0.008$ \\
3 & 0.000 & 0.750 & $-0.003$ & $-0.006$ & 0.000 & 0.750 & $-0.003$ & $-0.006$ \\
4 & 0.000 & 0.750 & $-0.002$ & $-0.005$ & 0.000 & 0.750 & $-0.002$ & $-0.005$ \\
5 & 0.000 & 0.750 & 0.002 & 0.000 & 0.000 & 0.750 & 0.002 & 0.000 \\
\bottomrule
\end{tabular*}
\vspace{-2mm}
\end{table*}

The evaluation contains 96 unseen sequences per fit and 48 future evaluator utility rows per sequence. Each writer predicts one action per sequence and condition; these utility rows score that action without entering the policy input. Future action utilities are generated once per fit and held fixed for both writers. The endpoint gate is closed before writer labels are opened: it compares all aligned future actions and the same future actions read from the empty state. Replay then transports complete key--utility pairs under nine fixed permutations. The first 48 evaluation episodes and first four permutations form protocol calibration; the remaining 48 episodes and five permutations form holdout. The decoder assigns a writer to the replay-sensitive class exactly when its calibration pair-action response exceeds the fixed zero threshold. It is evaluated on new episodes and permutations under the same frozen pair, so the split measures protocol transfer rather than population generalization across writer families. The writers are frozen before replay, and no evaluation target is supplied to either fit. Permutations are repeated interventions within an episode, and future utility rows are repeated scoring contexts; all 95\% intervals use the episode as the sampling unit. The prespecified suite contains five additive fits and twenty recurrent restart fits, yielding five retained pairs. A failed endpoint gate is recorded as a failed condition and cannot trigger a replacement fit, an evaluation-based selection, or a rerun. The aligned action mismatch and the state-on--state-off contrast gap are zero for all five fits. The additive writer has zero pair response by its terminal state law. The recurrent writer changes 67.500\% of future actions on average in calibration and 68.167\% on the disjoint holdout; every per-fit episode bootstrap interval remains strictly above zero. A threshold decoder that sees only calibration pair-action response classifies both writers correctly on calibration and predicts both holdout classes correctly. Per-fit point estimates and episode-bootstrap limits are in Tables~\ref{tab:independent-endpoint-pair-endpoint} and \ref{tab:independent-endpoint-pair-replay}; classification accuracies are in Table~\ref{tab:independent-endpoint-pair-classification}. The learned pair reproduces the endpoint-equivalence construction without a lookup table under the finite-support synthetic interface. The result is scoped to the canonical order and prescribed utility margin; semantic-memory behavior lies outside this experiment.

\begin{table*}[ht]
\vspace{-4mm}
\centering
\small
\caption{Per-fit replay responses for five independently fitted endpoint pairs. Each additive--recurrent pair is fitted on separate sequence streams, with parameters frozen before evaluation. $A_{\rm cal}$ and $A_{\rm ho}$ are complete-pair action response proportions on calibration and holdout episodes. Lower and Upper give the endpoints of 95\% bootstrap confidence intervals; episodes are resampled within each frozen fit. Bold and underlined estimates denote recurrent and additive writers, respectively.}
\label{tab:content-fit-pairs}
\label{tab:independent-endpoint-pair-replay}
\begin{tabular*}{\textwidth}{@{\extracolsep{\fill}}lrrrrrrr@{}}
\toprule
\textbf{Writer} & \textbf{Fit} & \multicolumn{3}{c}{$A_{\rm cal}$} & \multicolumn{3}{c}{$A_{\rm ho}$} \\
\cmidrule(lr){3-5}\cmidrule(l){6-8}
 & & \textbf{Estimate} & \textbf{Lower} & \textbf{Upper} & \textbf{Estimate} & \textbf{Lower} & \textbf{Upper} \\
\midrule
Additive & 1 & \underline{0.000} & 0.000 & 0.000 & \underline{0.000} & 0.000 & 0.000 \\
Recurrent & 1 & \textbf{0.703} & 0.635 & 0.776 & \textbf{0.738} & 0.671 & 0.804 \\
Additive & 2 & \underline{0.000} & 0.000 & 0.000 & \underline{0.000} & 0.000 & 0.000 \\
Recurrent & 2 & \textbf{0.646} & 0.531 & 0.750 & \textbf{0.538} & 0.431 & 0.642 \\
Additive & 3 & \underline{0.000} & 0.000 & 0.000 & \underline{0.000} & 0.000 & 0.000 \\
Recurrent & 3 & \textbf{0.682} & 0.594 & 0.766 & \textbf{0.654} & 0.571 & 0.733 \\
Additive & 4 & \underline{0.000} & 0.000 & 0.000 & \underline{0.000} & 0.000 & 0.000 \\
Recurrent & 4 & \textbf{0.609} & 0.536 & 0.677 & \textbf{0.683} & 0.617 & 0.750 \\
Additive & 5 & \underline{0.000} & 0.000 & 0.000 & \underline{0.000} & 0.000 & 0.000 \\
Recurrent & 5 & \textbf{0.734} & 0.661 & 0.807 & \textbf{0.796} & 0.738 & 0.854 \\
\bottomrule
\end{tabular*}
\vspace{-2mm}
\end{table*}

\begin{table*}[ht]
\vspace{-4mm}
\centering
\small
\caption{Per-fit blind classification accuracies for the independently learned pair. $B_{\rm cal}$ and $B_{\rm ho}$ are calibration and holdout accuracies from a decoder fixed on calibration replay; the five fit pairs are the same as in the endpoint and replay tables. Accuracies are proportions and Fit is an integer index.}
\label{tab:independent-endpoint-pair-classification}
\begin{tabular*}{\textwidth}{@{\extracolsep{\fill}}rrrrr@{}}
\toprule
 & \multicolumn{2}{c}{\textbf{Additive}} & \multicolumn{2}{c}{\textbf{Recurrent}} \\
\cmidrule(lr){2-3}\cmidrule(lr){4-5}
\textbf{Fit} & $B_{\rm cal}$ & $B_{\rm ho}$ & $B_{\rm cal}$ & $B_{\rm ho}$ \\
\midrule
1 & 1.000 & 1.000 & 1.000 & 1.000 \\
2 & 1.000 & 1.000 & 1.000 & 1.000 \\
3 & 1.000 & 1.000 & 1.000 & 1.000 \\
4 & 1.000 & 1.000 & 1.000 & 1.000 \\
5 & 1.000 & 1.000 & 1.000 & 1.000 \\
\bottomrule
\end{tabular*}
\vspace{-2mm}
\end{table*}


\subsection{Operator factorization and falsification}
\label{app:operator-factorization}

The four state-boundary cells form a matched intervention on the label and utility coordinates supplied to the writer. Write $\boldsymbol\sigma=(\sigma_{\rm v},\sigma_{\rm p},\sigma_{\rm t})$ for the three cell-specific slot permutations. For a fixed outer map $\rho$, the coordinates carried by the current slot are
\begin{equation}
\begin{array}{c|cc}
\text{cell} & \text{producer label} & \text{exposed utility}\\
\hline
\mathrm{aligned} & \rho(a_t) & y_t\\
\mathrm{value} & \rho(a_t) & y_{\sigma_{\rm v}(t)}\\
\mathrm{pair} & \rho(a_{\sigma_{\rm p}(t)}) & y_{\sigma_{\rm p}(t)}\\
\mathrm{time} & \rho(a_{\sigma_{\rm t}(t)}) & y_t .
\end{array}
\label{eq:cell-factorization}
\end{equation}
The value cell changes the association between a label and a utility while holding the label sequence, update slots, and utility multiset fixed. The pair cell changes the temporal position of complete pairs. The time cell changes the stored label at a slot while retaining the utility sequence. For a writer with a known algebra, this factorization isolates the intended comparisons; for a general sequential writer, the joint contrast remains explicit. The task input $x_t$ remains attached to its original slot in all four cells; a pair-only interpretation therefore requires the commutative, position-free conditions~stated~below.

For each future readout $y$, define the pair-transport residual and the diagonal value--time contrast at a fixed $(r,R,\rho,\boldsymbol\sigma)$ by
\begin{equation}
\begin{aligned}
P_{y,r,R}^{m}(\rho,\boldsymbol\sigma)
  =\delta_y^{m,\mathrm{pair}}(\rho,\boldsymbol\sigma)
    -\delta_y^{m,\mathrm{aligned}}(\rho,\boldsymbol\sigma),E_{y,r,R}^{m}(\rho,\boldsymbol\sigma)
  =\delta_y^{m,\mathrm{value}}(\rho,\boldsymbol\sigma)
    -\delta_y^{m,\mathrm{time}}(\rho,\boldsymbol\sigma).
\end{aligned}
\label{eq:local-factor-contrasts}
\end{equation}
The first contrast asks whether transporting complete label--utility pairs changes the readout relative to the aligned state. The second is a diagonal value--time contrast: both the exposed label sequence and the exposed utility sequence differ between its two cells, so it is not a label-only exposure effect. In the primary synthetic protocol, the pair--value contrast uses the same realized slot map in both cells, $\sigma_{\rm p}=\sigma_{\rm v}$; contrasts involving time use its separately declared label-permutation law. Their averages are descriptive factor contrasts under these stated couplings. Neither contrast creates a new sampling unit: $(r,R)$ remains nested in the replicate stream and $\rho,\boldsymbol\sigma$ are coupled interventions.

\noindent\textbf{Proposition (pair transport under commutativity).} Suppose the state update on every block has the form
\begin{equation}
s_T=s_0+\sum\nolimits_{t=1}^{T}\varphi(\ell_t,u_t).
\label{eq:commutative-terminal-state}
\end{equation}
where $s_0$ is independent of slot position, $\varphi$ is independent of position and time, and the future readout is a deterministic function of the terminal state and the held-fixed future. If the pair cell transports $(\ell_t,u_t)$ jointly, then
\begin{equation}
s_T^{\mathrm{pair}}(\rho,\boldsymbol\sigma)
 =s_T^{\mathrm{aligned}}(\rho,\mathrm{id}),\quad
P_{y,r,R}^{m}(\rho,\boldsymbol\sigma)=0
\quad\text{for }y\in\{U,A,O\}.
\label{eq:pair-zero-law}
\end{equation}

\noindent\textbf{Proof.} The pair cell is a permutation of the complete summands $\varphi(\rho(a_t),y_t)$. Commutativity of addition makes their sum equal to the aligned sum. The future task set, executor draws, and tie rule are coupled, so the deterministic readout receives identical inputs. Equality holds pointwise for each replicate, stratum, map, and admissible permutation; averaging gives the stated zero law. A pair contrast that exceeds the declared numeric tolerance for deterministic checks, or the pre-specified cluster-level decision criterion when inference is used, falsifies at least one declared condition: the update is order-dependent, the state contains an unmodeled position component, the readout receives an uncoupled input, or the intervention was not applied to the complete pair.

The same argument gives a negative-control law. If $s_T=s_0+\sum\nolimits_t\psi(u_t)$ and the future rule does not inspect producer labels, then every outer map leaves the terminal state unchanged. Thus
\begin{equation}
\delta_y^{m,\mathrm{aligned}}(\rho,\boldsymbol\sigma)
 =\delta_y^{m,\mathrm{time}}(\rho,\boldsymbol\sigma)=0
\quad\Longrightarrow\quad
\Gamma_y^{m,\mathrm{time}}=0 .
\label{eq:label-omitting-zero-law}
\end{equation}
This is a calibration prediction for the intervention, not an assumption that all useful state is label-invariant. The arm-indexed writers are informative only to the extent that their observable response separates from this calibrated null under the same future.

\noindent\textbf{Falsification criteria.} The operator interpretation is rejected by a nonzero response of the arm-omitting state or by a nonzero pair response of a writer declared commutative. The blinded recovery rule is evaluated as a separate calibration claim: a failure rejects that predeclared decoder on the tested draw, while a low response can mark a resolution boundary without falsifying the underlying state displacement. The transport interpretation is rejected when the same writer class changes the declared utility--oracle ordering after the policy or substrate is changed while the information boundary is held fixed, or when a commutative writer violates the pair-zero law. The action response is interpreted as a readout diagnostic. A change in a pooled efficacy endpoint alone does not establish or falsify the operator claim, because that endpoint is a different functional of the future readout.

\subsection{Pooling, clustering, and finite-support inference}
\label{app:pooling-inference}

The reported structural quantities use a fixed hierarchy. Let $d_{y,r,R,x}^{m,q}(\rho,\sigma^\star)$ be the readout difference for future task $x$ in stratum $R$ of replicate $r$. The task average within a stratum and the map-averaged replicate summary are
\begin{equation}
\bar d_{y,r,R}^{m,q}(\rho,\sigma^\star)
 =\frac{1}{48}\sum\nolimits_{x\in\mathcal X^+_{r,R}}
 d_{y,r,R,x}^{m,q}(\rho,\sigma^\star),\quad
\bar d_{y,r}^{m,q}(\sigma^\star)
 =\frac{1}{6}\sum\nolimits_{R\in\mathcal R}
  \frac{1}{9}\sum\nolimits_{\rho\in\mathcal D_4}
  \bar d_{y,r,R}^{m,q}(\rho,\sigma^\star).
\label{eq:replicate-pooling}
\end{equation}
The fixed-draw estimate is
\begin{equation}
\widehat{\Gamma}_{y}^{m,q}(\sigma^\star)
 =\frac{1}{48}\sum\nolimits_{r=1}^{48}
 \left(\bar d_{y,r}^{m,\mathrm{aligned}}
       -\bar d_{y,r}^{m,q}\right).
\label{eq:cluster-estimator}
\end{equation}
Thus the six strata receive equal weight, each of the nine maps is averaged within a replicate, and future tasks are averaged inside a stratum before cluster-level inference. The complete primary future matrix contains $48\times6\times48=13{,}824$ future decisions per cell. This count describes the finite support of the readout; it is not the number of independent scientific observations. The inferential sample size is 48 replicate streams.

For a bootstrap draw $b$, sample replicate indices $r_1^{(b)},\ldots,r_{48}^{(b)}$ with replacement and recompute
\begin{equation}
\widehat{\Gamma}_{y}^{m,q\,*(b)}
 =\frac{1}{48}\sum\nolimits_{i=1}^{48}
 \left(\bar d_{y,r_i^{(b)}}^{m,\mathrm{aligned}}
       -\bar d_{y,r_i^{(b)}}^{m,q}\right).
\label{eq:cluster-bootstrap}
\end{equation}
The future task set, map set, realized $\sigma^\star$, fitted parameters, and stratum weights are fixed inside this resampling. Percentiles of the 20,000 complete-cluster draws give the reported intervals. For the sign-flip calibration, define the replicate-level interaction $c_{y,r}^{m,q}=\bar d_{y,r}^{m,\mathrm{aligned}}- \bar d_{y,r}^{m,q}$. The test replaces this complete interaction by $\varepsilon_r c_{y,r}^{m,q}$ with $\varepsilon_r\in\{-1,+1\}$ sampled independently; it therefore tests a cluster-level sign-symmetry null. It does not test independence of tasks, maps, or controller calls. The same hierarchy is used for the cross-policy fits after the fitting streams are removed from the~evaluation~set.

The law-level target and the fixed-draw estimator answer different questions. For the declared permutation law $P_q$, Eq.~\eqref{eq:formal-gamma} defines the law-level target as an average over admissible slot draws:
\begin{equation}
\overline{\Gamma}_y^{m,q}
 =\mathbb E_{\sigma\sim P_q}
   \left[\Gamma_y^{m,q}(\sigma)\right].
\label{eq:law-level-gamma}
\end{equation}
The reported estimator conditions on the single realized collection $\sigma^\star$ and estimates replicate variation around that draw. A draw-sensitivity analysis can reveal whether the observed ordering is unusually dependent on $\sigma^\star$, but it does not convert the fixed draws into additional replicates. For a fitted writer, the same conditioning applies to its frozen parameters: bootstrap variation reflects evaluation replicates, while parameter initialization and the training sample are held at their declared values unless a separate refit analysis is named.

\subsection{Cross-interface signatures and summaries}
\label{app:transport-claim}
\label{app:cross-interface-transport}

The cross-interface analyses use a readout summary of the replay signature as the object compared across interfaces. For writer $m$, executor $g$, and future support $\mathcal X^+$, define
\begin{equation}
\mathcal S(m,g,\mathcal X^+)
 =
 \left(
 \Gamma_U^{m,\mathrm{value}},
 \Gamma_A^{m,\mathrm{value}},
 \Gamma_O^{m,\mathrm{value}},
 \Gamma_U^{m,\mathrm{pair}},
 \Gamma_A^{m,\mathrm{pair}},
 \Gamma_O^{m,\mathrm{pair}}
 \right).
\label{eq:transport-signature}
\end{equation}
This summary is an observable vector at the state boundary. The action component depends on the future query and executor readout, so it is reported as a diagnostic rather than required to keep a fixed sign across interfaces. For the stable part of the signature, define
\begin{equation}
\operatorname{ord}_{UO}(\mathcal S)
 := \left(\operatorname{sign}\Gamma_U^{m,\mathrm{value}},
            \operatorname{sign}\Gamma_O^{m,\mathrm{value}}\right).
\label{eq:transport-order-statistic}
\end{equation}
A cross-interface pattern is compatible when the selected writer class preserves $\operatorname{ord}_{UO}$ under the same outer intervention. For the changed-substrate check, compatibility means that this ordering is observed again after replacing the executor, task generator, and future support while retaining the selected-outcome update boundary; compatibility permits the writer map to vary across substrates. When the writer is declared commutative, the pair components supply the separate zero-law check in Eq.~\eqref{eq:pair-zero-law}.

This definition separates signature-pattern preservation from performance. The first asks whether the state-writing relation remains visible to the same family of interventions. The second asks whether the relation changes a particular future utility functional. A compatible signature can coexist with a zero pooled efficacy contrast when the future support averages opposing strata. A positive efficacy contrast without a compatible signature does not identify which state relation produced it. The cross-policy and changed-substrate analyses therefore report point estimates for the signature components and retain efficacy as a separate endpoint.

For a sequence of interfaces $\mathcal I_0,\ldots,\mathcal I_K$, let $\mathcal T_k$ denote the map that replaces the policy, executor, or future support at step $k$ while preserving the selected-outcome boundary. The pattern-preservation claim is the conjunction
\begin{equation}
\operatorname{ord}_{UO}\!\left(\mathcal S_k\right)
 =
\operatorname{ord}_{UO}\!\left(\mathcal S_0\right)
\quad\text{for each declared }k,
\quad
\mathcal S_k
 =\mathcal S(m_k,g_k,\mathcal X_k^+),
\label{eq:transport-order}
\end{equation}
where $\operatorname{ord}_{UO}$ records the predeclared utility/oracle sign pattern; action and pair responses remain separately reported diagnostics. The equation states a conditional invariance of an observable pattern within the declared synthetic executor. Extending it to open-ended scientific agents would require a separately declared future support, evaluator, and state interface.

The synthetic cross-interface checks keep the value-null signature fixed while changing the policy, fit, or scientific substrate. Fitted rows retain separate fitting and evaluation streams, and the changed-substrate row retains its declared evaluation boundary.

\begin{table}[ht]
\vspace{-4mm}
\centering
\small
\caption{Cross-interface value-null interactions. Entries are fixed-draw aligned-minus-value estimates of $\Gamma_U^{\mathrm{value}}$, $\Gamma_A^{\mathrm{value}}$, and $\Gamma_O^{\mathrm{value}}$ for the stated evaluation setting. Utility is normalized; action and oracle interactions are differences of proportions.}
\label{tab:main-transport-summary}
\begin{tabular}{@{}lrrr@{}}
\toprule
\textbf{Evaluation} & $\boldsymbol{\Gamma_U^{\mathrm{value}}}$ & $\boldsymbol{\Gamma_A^{\mathrm{value}}}$ (proportion) & $\boldsymbol{\Gamma_O^{\mathrm{value}}}$ (proportion) \\
\midrule
Policy-disjoint & 0.022 & $-0.014$ & $-0.064$ \\
Reverse policy & 0.063 & 0.023 & $-0.214$ \\
Independent fit & 0.025 & 0.013 & $-0.094$ \\
Changed substrate & 0.012 & 0.054 & $-0.049$ \\
\bottomrule
\end{tabular}
\vspace{-2mm}
\end{table}


\begin{table*}[ht]
\vspace{-4mm}
\centering
\small
\caption{Fitted-state and changed-substrate transport checks. $n_{\rm fit}$ and $n_{\rm eval}$ count fitting and evaluation streams; $n_{\rm fit}=0$ denotes a fixed analytic binding. $\widehat{\Gamma}$ is the aligned-minus-condition estimate in Eq.~\eqref{eq:main-interactions}, conditional on the intervention draw; the final row pools the value condition across strata. Utility is normalized; action and oracle interactions are differences of proportions. Fitted parameters are frozen before evaluation.}
\label{tab:fitted-transport-summary}
\begin{tabular*}{\textwidth}{@{\extracolsep{\fill}}llrrrrr@{}}
\toprule
\textbf{Writer} & \textbf{Contrast} & $n_{\rm fit}$ & $n_{\rm eval}$ & $\widehat{\Gamma}_U$ & $\widehat{\Gamma}_A$ & $\widehat{\Gamma}_O$ \\
\midrule
Selected-only & Value null & 48 & 48 & 0.024 & 0.003 & $-0.089$ \\
Selected-only & Time placebo & 48 & 48 & 0.024 & 0.005 & $-0.089$ \\
Fitted associative & Value null & 48 & 48 & 0.039 & 0.018 & $-0.094$ \\
Dense continuous & Value null & 0 & 48 & 0.019 & 0.027 & $-0.055$ \\
Cross-namespace & Value null & 0 & 48 & 0.021 & 0.019 & $-0.069$ \\
Changed substrate & Value null & 24 & 24 & 0.012 & 0.054 & $-0.049$ \\
\bottomrule
\end{tabular*}
\vspace{-2mm}
\end{table*}


\section{Numerical substrate specification}
\label{app:substrate-details}
\label{app:synthetic-substrates}

The source-localization executor samples three two-dimensional latent sources and scores a grid/adaptive/replicate/mixed query policy by permutation-invariant mean squared error. The temporal-discount executor samples a latent log discount parameter and scores the fitted parameter after binary choices at selected delays. The survival-risk executor samples a time--metastasis pool, selects labels, fits a logistic risk model, and scores held-out log loss. Contexts modify noise or event prevalence. Development, adaptation, and future phases modify the hidden parameter distribution. Replicate streams define the sampling units, and state variants share the task and outcome sequence within each stream.

For each replicate, task, split, and arm, a deterministic indexed draw fixes the latent world, observations, and counterfactual utilities. These quantities form the executor draw $z_t$ or $z_{r,x}$ in the formal protocol. They are realized once when the task stream is constructed; replay conditions on them and varies only the declared slot permutations and writer readout.

\noindent\textbf{Finite-world determinism.}  The numerical substrate is a finite indexed object rather than a stream of fresh replay randomness. Let $j$ be the canonical global task index, $q$ the local index within its split, and $r$ the replicate. Each executor draw is a deterministic function of these indices and the arm coordinate. Let $\mathcal I_{\mathrm{draw}}$ collect these coordinates:
\begin{equation}
b:\mathcal I_{\mathrm{draw}}\longrightarrow\mathbb N,
\quad b\ \text{fixed and injective}.
\label{eq:substrate-world-index}
\end{equation}
Distinct arm and condition coordinates therefore address distinct draws. Once these draws are realized, every cell in Eq.~\eqref{eq:cell-functional} is deterministic. The value and pair cells share one within-block draw, while the time cell has its declared independent draw; additional realizations are sensitivity conditions and do not create additional sampling units.

The scientific generators use the following fixed parameterization. In source localization, three latent points are drawn from an isotropic Gaussian with phase-dependent standard deviation $(1.15,0.62,0.90)$; observation noise is $0.22$ in the low-noise context and $0.75$ otherwise. Ten queries are made, and utility is $\exp(-\mathrm{MSE}/1.3)$ after permutation-invariant matching. Coverage follows a nine-point grid, replicate perturbs the best observed query by standard deviation $0.04$, adaptive samples around it with radius $0.9$ for five steps and $0.35$ thereafter, and mixed begins with five grid queries. In temporal discount estimation, $\kappa=\log k$ is Gaussian with phase means $-4.25,-3.55,-3.90$ and standard deviation $0.48$, and $k=\exp(\kappa)$. Each query uses fixed values $i_{\mathrm{ref}}=20$ and $d_{\mathrm{ref}}=120$, together with a delay $d$ in days; a binary response has probability $0.01+0.98\,\operatorname{sigmoid}((d_{\mathrm{ref}}/(1+kd)-i_{\mathrm{ref}})/\alpha)$, with context scales $0.55$ and $1.8$ for $\alpha$. With $t=0$ at query index zero, coverage and initial mixed use delays $(1,4,8,16,28,45,70,105,150,210)$; replicate uses $45+3(t\bmod 3)$. After three observations, adaptive uses $d_t=\operatorname{clip}_{[1,260]}(\operatorname{round}((100/\hat{k}_t) [1+0.12(t\bmod 3-1)]))$, where $\hat{k}_t$ minimizes Bernoulli negative log-likelihood on the history over the grid $\{\exp(s):s\in\operatorname{linspace}(-6,-1,360)\}$ and equals $\exp(-4.25)$ before any observation. Utility is $\exp(-|\log\hat{k}-\log k|/0.72)$. In survival risk, each task draws 160 $(t,m)$ pairs with $t\sim\mathrm{Unif}(0,10)$ and metastasis probability $0.46$ (balanced) or $0.16$ (rare); labels follow a logistic model with coefficients $(-2.0,0.42,\beta_3)$, where $\beta_3$ is $1.6$, $0.8$, or $1.1$ by phase. Ten labels are selected and held-out log loss over 260 fresh points is mapped to utility by $\exp(-\mathrm{logloss})$. These equations specify the synthetic scientific substrate; they are not claims about a natural scientific data-generating process.

\section{Controller and replay specification}
\label{app:controller-details}

This appendix fixes the controller implementations and replay controls used in the primary matrix. The summary table gives the headline contrasts; the definitions below specify each state update, tie rule, and control condition.

\begin{table*}[ht]
\vspace{-4mm}
\centering
\small
\caption{Matched replay contrasts on 48 streams. Panel A reports differences in replay responses. Panel B compares OCRR's written and empty states: utility and oracle agreement are written minus empty, whereas action is disagreement. Utility is normalized; action and oracle entries use proportion units. Complete intervals are reported in Appendix~\ref{app:structural-results}.}
\label{tab:content-contrasts}
\begin{tabular*}{\textwidth}{@{\extracolsep{\fill}}llrrr@{}}
\toprule
\multicolumn{5}{l}{\textbf{A. Replay interactions}} \\
\midrule
\textbf{Writer} & \textbf{Comparison} & \textbf{Utility} & \textbf{Action} & \textbf{Oracle} \\
\midrule
OCRR & Aligned $-$ value & 0.304 & $-0.038$ & $-0.412$ \\
OCRR & Aligned $-$ time & 0.323 & $-0.049$ & $-0.427$ \\
OCRR & Pair $-$ value & 0.302 & $-0.003$ & $-0.405$ \\
UCB1 & Aligned $-$ time & 0.129 & $-0.468$ & $-0.113$ \\
Local-only & Aligned $-$ time & 0.241 & $-0.341$ & $-0.310$ \\
Global-only & Aligned $-$ time & 0.298 & 0.014 & $-0.273$ \\
\bottomrule
\end{tabular*}

\smallskip
\begin{tabular*}{\textwidth}{@{\extracolsep{\fill}}lrrr@{}}
\toprule
\multicolumn{4}{l}{\textbf{B. State-use contrast}} \\
\midrule
\textbf{Comparison} & \textbf{Utility gap} & \textbf{Action disagreement} & \textbf{Oracle gap} \\
\midrule
Written $-$ empty & 0.380 & 0.934 & 0.377 \\
\bottomrule
\end{tabular*}
\vspace{-2mm}
\end{table*}

For \texttt{OCRR}, local state keys are \texttt{(family, context, arm)}. When a local key has no evidence, the selector backs off to a global arm prior accumulated from earlier learning families; this cross-family path is part of the specified design. The selector uses the empirical mean, a $0.28/\sqrt{1+n}$ exploration term, and a lower-confidence bonus after four observations. A key is quarantined for six subsequent policy steps when at least six observations show a current utility below 0.32 and a running mean below 0.46. More precisely, for a non-quarantined arm with local count $n$ and mean $\bar u$, the score is
\begin{equation}
 s=\bar u+\frac{0.28}{\sqrt{1+n}}+0.1\max\{0,L\},\quad
 L=\begin{cases}
 \bar u-1.64\sqrt{\max\{\bar u(1-\bar u),0.02\}/n},&n\ge4,\\
 0,&n<4.
 \end{cases}
\label{eq:ocrr-score}
\end{equation}
When a local key is unvisited, $\bar u$ backs off to the corresponding global arm mean, or to $0.5$ when that arm is also unvisited. The canonical source order is \texttt{coverage}, \texttt{adaptive}, \texttt{replicate}, \texttt{mixed}; the selector chooses the largest score, breaking exact ties by the lexicographically largest canonical label. Hence an empty OCRR state selects \texttt{replicate}. Promotion markers do not change selection. \memory{} stores the last successful arm. \skill{} aggregates evidence over contexts within a family. Survival-risk has no local adaptation block. Its future-boundary state can still contain the global arm prior accumulated through earlier learning-family updates, while the survival future itself remains held fixed. This makes the survival strata a boundary case for replay: the intervention tests the arm labels in the carried state rather than a separate survival adaptation phase.

\shuffle{} uses the same selector as \texttt{OCRR}. During adaptation, each family--context key has a past-only utility buffer. The initial update uses a fixed neutral prior of 0.5; subsequent updates sample one utility from the same key's past buffer, and only then append the current utility. Thus every adaptation slot has one state-write opportunity, while the current outcome is never assigned to the current key. This primary control preserves family/context and update-slot structure while changing both producer-label pairing and the evolving evidence distribution.

The secondary \replay{} uses the OCRR adaptation action trace and selected outcomes. It writes each outcome under the fixed cyclic map
\begin{equation}
\texttt{coverage}\mapsto\texttt{adaptive}\mapsto\texttt{replicate}
\mapsto\texttt{mixed}\mapsto\texttt{coverage}.
\label{eq:replay-cyclic-map}
\end{equation}
then evaluates future tasks with that OCRR selector and no future updates. The action trace, outcome multiset, and update count remain fixed. Only the arm label receiving each adaptation outcome changes. \staticselected{} uses a fixed selected-arm schedule on each development task and freezes its per-family/context means before adaptation. \contextucb{} uses the same family--context--arm keys and global arm fallback, but applies a contextual upper-confidence-bound (UCB1) score to selected adaptation outcomes and has no OCRR quarantine rule. At initialization all local and global counts and sums are zero; an unvisited arm receives an infinite score, and ties use the deterministic arm-order tie rule specified for the protocol. After an arm is visited, the score is the empirical mean plus $\sqrt{2\log(\max(2,N))/n}$, with the global arm mean used when the local key is unseen. Because the infinite-score rule resolves every unvisited local arm before the fallback is queried, the global fallback does not select an arm in that case; the survival strata have no adaptation update. The replay contrast is conditional on this explicit arm-indexed representation and measures label sensitivity; action causality lies outside this contrast.

The label-invariant negative control is a by-construction arm-omitting nuisance-key null: it stores one outcome aggregate per family/context and omits arm labels from update and selection. Its expected response to every arm-label map is therefore exactly zero. Using the same 48 replicate traces and future evaluator, aligned and deranged replay have identical states, actions, and utility in all six strata (13,824 future tasks; both responses $0.0$). The time-preserving blockwise shuffle is the complementary label-sensitive decoy; the three expected regimes are null response for the nuisance-key state, generic rekeying response for an arm-indexed state, and an operator contrast only when aligned and placebo differ. These regimes calibrate label sensitivity; the executor-only oracle below evaluates held-out decision quality within the synthetic substrate and does not assign semantic correctness to a nonzero replay response.

\noindent\textbf{Label-equivariance calibration.} The protocol also includes a full global-renaming null. For every permutation of the four arm labels, the stored arm keys and the future candidate interface are renamed together, while the external arm returned to the evaluator is mapped back to the canonical label. This is the declared equivariant transformation of the arm-indexed ledger, so its expected action, utility, regret, and oracle-match differences are zero. The calibration uses the canonical equivariant selector as an algebraic consistency check; it does not exercise the ordinary controller integration path. Across all 48 replicate streams and 13,824 future tasks for each of the 24 permutations, every one of these four checks is exactly zero. The calibration is a known representation null. As an exploratory fixed-label robustness analysis, the same OCRR adaptation trace was replayed under the nine derangements of the four arm labels, the complete set of fixed-point-free permutations. Map strings use source order \texttt{coverage, adaptive, replicate, mixed}; for example, \texttt{ACMR} means C$\to$A, A$\to$C, R$\to$M, and M$\to$R. The nine derangements reuse the same 48 replicate streams, so they are not independent samples. Their pooled contrasts are given in the following table; their changed-substrate counterparts appear below.

\begin{figure}[ht]
\centering
\includegraphics[width=\linewidth]{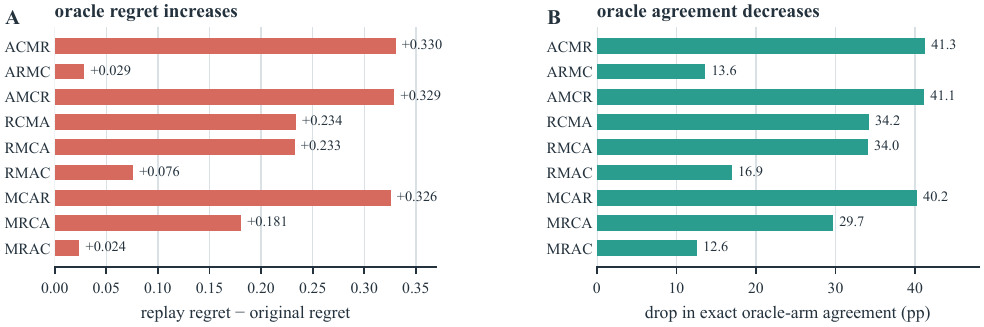}
\vspace{-5mm}
\caption{\textbf{Executor-relative consequences of fixed-label replay.} The two panels give the map-level increase in held-out oracle regret and the corresponding drop in exact oracle-arm agreement. Both readouts use the same future records as the primary fixed-label replay and remain outside the controller state.}
\label{fig:oracle-audit}
\vspace{-3mm}
\end{figure}

The executor also retains an evaluation-only counterfactual oracle for each future task. We use it after replay to compare oracle regret and exact oracle-arm agreement, without exposing either quantity to the controller. The oracle uses the fixed source order coverage, adaptive, replicate, mixed and selects the first entry, coverage, under an exact utility tie, matching the locked evaluator's first-maximum convention. Every primary derangement increases held-out oracle regret and lowers oracle-arm agreement under the synthetic decision criterion. Regret increase is algebraically the identity-minus-replay utility contrast; Figure~\ref{fig:oracle-audit} plots the map-level utility and executor-relative consequences on the same future records. For the canonical cyclic map, the action-change summary averages the 48 replicate-level fractions in each stratum. Each replicate contributes 48 future decisions, so the result is also the task-weighted fraction reported in Table~\ref{tab:action-divergence}. This table reports the descriptive action-change summary and its uncertainty; the nine-map table reports utility contrasts.

\begin{table*}[ht]
\vspace{-4mm}
\centering
\small
\caption{Fraction of future tasks on which the action-sequence-preserving \replay{} selects a different arm from OCRR. The short family labels Source, Temporal, and Survival denote source localization, temporal discount, and survival risk. Each row contains $n=2304$ future tasks from 48 replicate clusters; Estimate, Lower, and Upper are the point estimate and approximate 95\% replicate-cluster interval endpoints, all reported as proportions.}
\label{tab:action-divergence}
\begin{tabular}{@{}llrrrr@{}}
\toprule
Family & Context & $n$ & Estimate & Lower & Upper \\
\midrule
Source & Low & 2304 & 1.000 & 1.000 & 1.000 \\
Source & High & 2304 & 1.000 & 1.000 & 1.000 \\
Temporal & Low & 2304 & 0.875 & 0.771 & 0.958 \\
Temporal & High & 2304 & 0.750 & 0.625 & 0.875 \\
Survival & Balanced & 2304 & 1.000 & 1.000 & 1.000 \\
Survival & Rare & 2304 & 1.000 & 1.000 & 1.000 \\
\bottomrule
\end{tabular}
\vspace{-2mm}
\end{table*}

\section{Structural-control details}
\label{app:structural-control-details}
\label{app:structural-results}

The structural suite analyzes operator behavior under the fixed measurement protocol and remains separate from the efficacy hierarchy. It reuses the 48 replicate streams, selected adaptation actions, selected outcomes, future tasks, counterfactual outcomes, and evaluator. Each of the nine arm derangements (fixed-point-free arm permutations) is applied to the same future tasks for each state variant; these are repeated interventions on the same replicate clusters. No task is regenerated and no future update is allowed. The placebo construction preserves the relevant blockwise outcome and label marginals, and the nine derangements do not add independent replicates.

The suite has five state variants. OCRR is the arm-indexed ledger. UCB1 keeps family--context--arm keys and global arm fallback but replaces the OCRR score and quarantine rule with contextual UCB1 selection. Local-only keeps the OCRR score while removing the carried global arm prior; global-only keeps only arm-level counts and means. The time-preserving placebo keeps the original adaptation actions and selected outcomes at their original positions, then permutes the stored labels across each family--context block. It changes 2,802 of 9,216 labels (30.4\%); repeated labels leave fixed positions. It preserves the block label multiset, outcome sequence, and update count while perturbing the stored-arm label relation at the block level. The label-invariant control is evaluated separately with the same held-fixed future and omits arm identity from its state. Each derived writer is reinitialized immediately before the adaptation trace; development calls are not replayed into the writer. Each result is therefore an adaptation-stage ledger measurement with a held-fixed future and a state interface defined at the adaptation boundary.

The trace-matched empty-state branch isolates the state-readout component of the controller comparison. It reuses the OCRR adaptation actions and selected outcomes, advances the same policy clock, applies the same OCRR choice rule at the initial empty state, and omits the state writes before the held-fixed future. The written-state minus empty-state contrast is $+0.380$ in future utility with interval $[+0.371,+0.389]$ and sign-flip $p<0.0001$; future actions differ on $93.4\%$ of tasks with interval $[91.0,95.8]\%$ and $p<0.0001$, while the executor-relative oracle-match contrast is $+37.7\,\text{pp}$ with interval $[+36.6,+38.8]~\text{pp}$ and $p<0.0001$. The contrast is a conditional state-readout estimand under the fixed synthetic executor. It is separate from the closed-loop OCRR--\noupdate{} endpoint and is not an additional confirmatory test.

The matched-count value-permutation null is a separate derived control. It retains the observed arm labels, every per-key prefix-count trajectory, the update clock, and the utility multiset within each of the 192 family--context blocks. A fixed random derangement assigns those utilities to adaptation slots before the state writer receives them; the future evaluator still uses the original counterfactual table. All 9,216 adaptation utilities change slots, while the label, prefix-count, utility-marginal, and future-task conditions are unchanged. The control therefore changes the label--utility pairing at the state update while matching nominal key counts, the update clock, the utility marginal, and future-task conditions. The reported interaction is conditional on the realized within-block draw $\sigma^\star$; the law-level target is defined in Appendix~\ref{app:formal-interface}. To assess finite-construction sensitivity, Table~\ref{tab:primary-operator-robustness} reports five fixed derangement draws from the same law; every reported stratum retains its direction across those draws. These draws are descriptive checks, not additional randomization populations or inference units.

The intervention law is fully specified at two levels. The outer family contains all nine arm derangements (fixed-point-free permutations) of the four arm labels. Within each 48-slot family--context adaptation block, the value null samples one uniform fixed-point-free permutation of the slots; the pair-preserving cell uses the same blockwise permutation and transports each complete label--utility pair. The time placebo instead draws an independent uniform permutation of the 48 slots within each block and writes the source label from the permuted slot at the current slot. It preserves the utility sequence; repeated labels may leave some projected labels in place. The value and pair cells therefore share the utility order and differ in whether a label travels with its utility. Conditional on these realized permutations, the 48 replicate streams are the sampling units, and the arm and slot permutations are repeated interventions within each unit.

The pair-preserving placebo supplies the complementary pair-transport control. For an order-sensitive writer it remains an order--exposure control; for a commutative writer it is invariant by construction. It uses the same utility permutation as the value null and moves each observed label--utility pair together within its block. The label--utility pairing is retained while temporal placement changes, and the two cells share the same utility order. Relative to the value null, the pair placebo retains a $+0.302$ utility interaction and a $-40.5\,\text{pp}$ oracle interaction, while its action interaction is $-0.3\,\text{pp}$. Both cells also change keyed exposure and update order.

\noindent\textbf{Known-operator calibration.} The structural suite instantiates three writer classes with analytically different responses. A nuisance-key writer has terminal state $h=\sum\nolimits_t\psi(u_t^q)$ and therefore satisfies $\Gamma_y=0$ for every arm-label map, including all nine derangements. A commutative paired writer has $h=\sum\nolimits_t\varphi(\ell_t,u_t)$; it is invariant to the pair-preserving cell, while its value-null interaction can be nonzero precisely when the feature map is pairing-sensitive. An order-sensitive writer has $h_{t+1}=F(h_t,\ell_t,u_t)$; for this class the pair-preserving cell can change the readout, so a nonzero value-null interaction remains a joint pairing--order--exposure contrast. The arm-omitting and nuisance writers provide the invariant calibration class; fitted associative and selected-outcome-only writers instantiate commutative classes, while OCRR and the continuous ledger variants instantiate sequential classes. UCB1 is treated separately as an independently specified count--sum writer whose pair row is a zero-law control. These cells calibrate the protocol's identification logic under the declared executor and state interface.

\begin{table}[ht]
\vspace{-4mm}
\centering
\small
\caption{Matched state variants on the primary fixed trace. Mean utility is the normalized identity-minus-mapped utility gap. Minimum and Maximum give its range over the nine producer-key derangements. Action is mapped-versus-identity disagreement; oracle is mapped-minus-identity agreement. Action and oracle use proportion units. All rows reuse 48 replicate streams. TP denotes the time-preserving placebo.}
\label{tab:content-controls}
\label{tab:representation-controls}
\begin{tabular*}{\textwidth}{@{\extracolsep{\fill}}lrrrrr@{}}
\toprule
\textbf{State variant} & \textbf{Mean utility} & \textbf{Minimum} & \textbf{Maximum} & \textbf{Action} & \textbf{Oracle} \\
\midrule
OCRR & 0.196 & 0.024 & 0.330 & 0.938 & $-0.293$ \\
UCB1 & 0.002 & $-0.138$ & 0.090 & 0.519 & 0.022 \\
Local-only & 0.114 & 0.016 & 0.179 & 0.645 & $-0.176$ \\
Global-only & 0.171 & $-0.024$ & 0.373 & 1.000 & $-0.139$ \\
TP placebo & $-0.127$ & $-0.207$ & 0.022 & 0.986 & 0.134 \\
\bottomrule
\end{tabular*}
\vspace{-2mm}
\end{table}

\begin{figure}[ht]
\centering
\vspace{-2mm}
\includegraphics[width=0.96\linewidth]{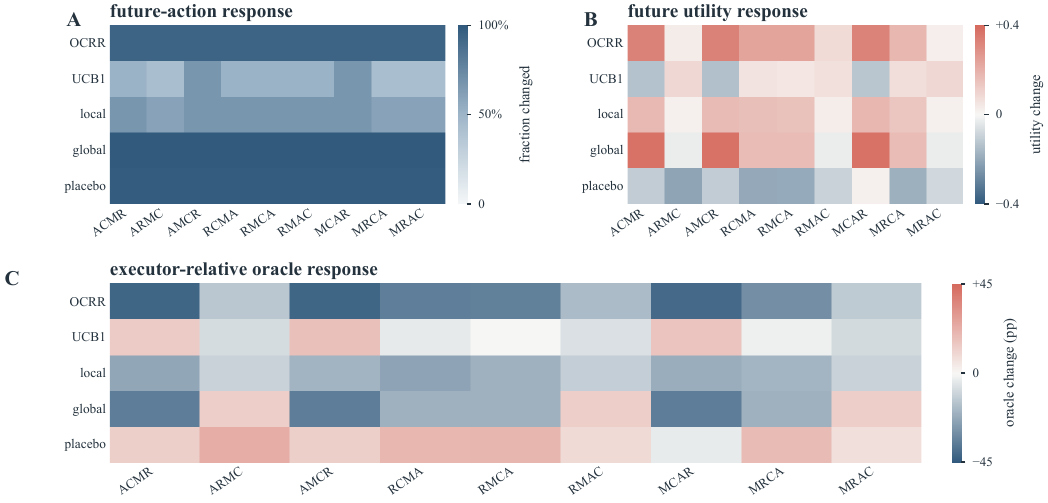}
\vspace{-4mm}
\caption{\textbf{Structural-control responses across replay maps.} Panels show future-action response, future-utility response, and executor-relative oracle response for the arm-indexed ledger, matched state variants, and the time-preserving placebo. The nine columns are paired derangements of the same replicate streams; the heatmaps display pooled point estimates under the fixed measurement protocol.}
\label{fig:representation-controls}
\vspace{-4mm}
\end{figure}

Table~\ref{tab:representation-controls} reports the pooled derangement estimates; Figure~\ref{fig:representation-controls} shows their mapwise pattern. OCRR has a mean identity-minus-replay utility of $+0.196$ and changes 93.8\% of future actions. The time-preserving placebo changes 98.6\% of actions despite keeping every selected outcome in place. Thus the action-change fraction alone cannot identify label--utility pairing. The executor-only oracle changes in opposite directions for OCRR and placebo, $-29.3$ and $+13.4$ percentage points respectively. Their paired interaction is $+0.323$ in future utility and $-42.7\,\text{pp}$ in oracle match over 48 replicate clusters (Table~\ref{tab:representation-interaction}). This difference is an executor-relative operator contrast under the fixed synthetic executor; it does not test~semantic~correctness.

\begin{table}[ht]
\vspace{-4mm}
\centering
\small
\caption{Paired OCRR--time-preserving-placebo interaction. The utility estimate contrasts aligned-minus-map responses, whereas action-change and oracle-match use mapped-minus-aligned responses; each interaction is OCRR minus the corresponding placebo response on the same derangement and replicate. Utility is normalized; action-change and oracle-match are differences expressed in proportion units. Estimate is the point estimate, while Lower and Upper are 95\% percentile replicate-cluster interval endpoints from 20,000 complete block resamples; $p$ is the sign-flip value.}
\label{tab:representation-interaction}
\begin{tabular}{@{}lrrrr@{}}
\toprule
Metric & Estimate & Lower & Upper & $p$ \\
\midrule
Future utility & 0.323 & 0.286 & 0.359 & $<0.001$ \\
Future actions changed & $-0.049$ & $-0.069$ & $-0.028$ & $<0.001$ \\
Oracle-match change & $-0.427$ & $-0.467$ & $-0.389$ & $<0.001$ \\
\bottomrule
\end{tabular}
\vspace{-2mm}
\end{table}


The matched-count value-permutation interaction is $+0.304$ in future utility, $-3.8\,\text{pp}$ in future-action change, and $-41.2\,\text{pp}$ in oracle-match change. The value-null--time-placebo comparison is descriptive across their separately declared intervention laws; its difference is not a shared-law test. The table gives point estimates, while the complete intervals and sign-flip values appear in the~uncertainty~analysis.

\begin{table}[ht]
\vspace{-4mm}
\centering
\small
\caption{Matched value-permutation control on the primary fixed trace. The value-permutation null keeps the observed arm labels, per-key prefix counts, update slots, and blockwise utility multiset, while permuting utilities across adaptation slots. Utility is normalized, and action and oracle entries are differences of proportions. The first two rows are aligned-minus-cell interactions; the last row is the value-null minus time-placebo contrast. Uncertainty intervals and sign-flip values are given in Appendix~\ref{app:statistical-analysis}.}
\label{tab:semantic-null-interaction}
\begin{tabular}{@{}lrrr@{}}
\toprule
Contrast & \textbf{Utility} & \textbf{Action (proportion)} & \textbf{Oracle (proportion)} \\
\midrule
OCRR $-$ value null & 0.304 & $-0.038$ & $-0.412$ \\
OCRR $-$ time placebo & 0.323 & $-0.049$ & $-0.427$ \\
Value null $-$ time placebo & 0.019 & $-0.010$ & $-0.016$ \\
\bottomrule
\end{tabular}
\vspace{-2mm}
\end{table}

\begin{table*}[ht]
\vspace{-4mm}
\centering
\small
\caption{Full uncertainty estimates for the matched value-permutation interactions. Utility is normalized, while action-change and oracle-match entries are differences expressed in proportion units. Estimate is the point estimate, while Lower and Upper are 95\% replicate-cluster bootstrap interval endpoints from 20,000 complete block resamples; $p$ is the two-sided cluster sign-flip value.}
\label{tab:semantic-null-detail}
\begin{tabular}{@{}llrrrr@{}}
\toprule
Contrast & Readout & Estimate & Lower & Upper & $p$ \\
\midrule
OCRR $-$ value null & Future utility & 0.304 & 0.266 & 0.341 & $<0.001$ \\
OCRR $-$ value null & Action change & $-0.038$ & $-0.063$ & $-0.017$ & 0.001 \\
OCRR $-$ value null & Oracle match & $-0.412$ & $-0.454$ & $-0.369$ & $<0.001$ \\
OCRR $-$ time placebo & Future utility & 0.323 & 0.286 & 0.359 & $<0.001$ \\
OCRR $-$ time placebo & Action change & $-0.049$ & $-0.069$ & $-0.028$ & $<0.001$ \\
OCRR $-$ time placebo & Oracle match & $-0.427$ & $-0.467$ & $-0.389$ & $<0.001$ \\
Value null $-$ time placebo & Future utility & 0.019 & $-0.027$ & 0.062 & 0.417 \\
Value null $-$ time placebo & Action change & $-0.010$ & $-0.028$ & 0.007 & 0.458 \\
Value null $-$ time placebo & Oracle match & $-0.016$ & $-0.059$ & 0.029 & 0.501 \\
\bottomrule
\end{tabular}
\vspace{-2mm}
\end{table*}

\begin{table*}[ht]
\vspace{-4mm}
\centering
\small
\caption{Pair-preserving placebo relative to the value-permutation null. The placebo uses the same utility permutation as the value null and moves each observed label--utility pair within its adaptation block, so producer association is retained while temporal placement changes. Utility entries are normalized, while action-change and oracle-match entries are differences expressed in proportion units. Estimate is the point estimate, while Lower and Upper are 95\% replicate-cluster interval endpoints; utility entries are differences of the two utility interactions under the signed identity-minus-replay convention.}
\label{tab:pair-preserving-detail}
\begin{tabular}{@{}llrrrr@{}}
\toprule
Contrast & Readout & Estimate & Lower & Upper & $p$ \\
\midrule
Pair placebo $-$ value null & Future utility & 0.302 & 0.263 & 0.340 & $<0.001$ \\
Pair placebo $-$ value null & Action change & $-0.003$ & $-0.028$ & 0.021 & 0.694 \\
Pair placebo $-$ value null & Oracle match & $-0.405$ & $-0.449$ & $-0.362$ & $<0.001$ \\
\bottomrule
\end{tabular}
\vspace{-2mm}
\end{table*}


\begin{table}[ht]
\vspace{-4mm}
\centering
\small
\caption{Aligned-minus-time interactions for each state writer. Utility is normalized; Lower and Upper delimit its 95\% replicate-cluster bootstrap interval, and $p_U$ is its sign-flip value. Action and oracle interactions are differences of proportions. All rows use the same 48 replicate streams and nine paired derangements.}
\label{tab:representation-writer-interactions}
\begin{tabular*}{\textwidth}{@{\extracolsep{\fill}}lrrrrrr@{}}
\toprule
Writer & \multicolumn{3}{c}{$\widehat{\Gamma}_U$} & $p_U$ & $\widehat{\Gamma}_A$ & $\widehat{\Gamma}_O$ \\
\cmidrule(lr){2-4}
& Estimate & Lower & Upper & & & \\
\midrule
OCRR & 0.323 & 0.286 & 0.359 & $<0.001$ & $-0.049$ & $-0.427$ \\
UCB1 & 0.129 & 0.093 & 0.164 & $<0.001$ & $-0.468$ & $-0.113$ \\
Local-only & 0.241 & 0.206 & 0.274 & $<0.001$ & $-0.341$ & $-0.310$ \\
Global-only & 0.298 & 0.262 & 0.333 & $<0.001$ & 0.014 & $-0.273$ \\
\bottomrule
\end{tabular*}
\vspace{-2mm}
\end{table}


\section{Opaque operator families and signature patterns}
\label{app:hidden-operator-family}

The operator study withholds the update law until the replay signature has been computed. Twelve prefix-causal state writers share one four-dimensional state, the same held-fixed future, executor, nine outer arm derangements, and four state-boundary cells. Their identities are assigned to opaque identifiers before evaluation; family names enter only after the signature has been decoded. Each of the six future strata contributes 48 future tasks per replicate, so a writer is evaluated on 288 future tasks in each of the 48 replicate streams. Averages over derangements and tasks remain paired within a replicate stream, and every numerical entry below is a fixed-draw calibration summary conditional on the realized $\sigma^\star$.

\subsection{Canonical prefix-causal operators}

The four classes are a replay-invariant nuisance state, a key-only state, a commutative label--utility state, and an order-sensitive state. All initialize at $s_0=0$ and use the same deterministic future readout \(\pi(s,x)=\operatorname{TieBreak}\arg\max\nolimits_i s_i\), with exact ties resolved by the first arm in the canonical order $(\mathrm{coverage},\mathrm{adaptive},\mathrm{replicate},\mathrm{mixed})$. Let $e_\ell$ be the basis vector of the supplied producer label and $c(u)=u-1/2$. The one-step laws are
\begin{equation}
\begin{aligned}
U_{\mathrm{inv}}(s,\ell,u)&=s+\alpha c(u)\mathbf 1, &
U_{\mathrm{key}}(s,\ell,u)&=s+\alpha e_\ell,\\
U_{\mathrm{pair}}(s,\ell,u)&=s+\alpha c(u)e_\ell, &
U_{\mathrm{order}}(s,\ell,u)&=d s+\bigl(1+0.01c(u)\bigr)e_\ell.
\end{aligned}
\label{eq:hidden-online-operators}
\end{equation}
The invariant, key-only, and commutative classes use $\alpha\in\{0.5,1,2\}$; the order-sensitive class uses $d\in\{0.05,0.20,0.60\}$. The update receives one current pair at each prefix step. It receives no future task, future outcome, or terminal-batch summary.

Positive scaling preserves the deterministic argmax in the first three classes, so their three parameterizations implement the same action rule. In the archived audit, $d=0.05$ and $d=0.20$ also produce identical stream-response profiles; $d=0.60$ differs on 14 of the 48 streams. The four families therefore contribute one, one, one, and two distinct operator response profiles on this support. These counts describe the complete stored stream profiles, not the number of statistically independent writer draws.

\begin{figure}[ht]
\centering
\includegraphics[width=\linewidth]{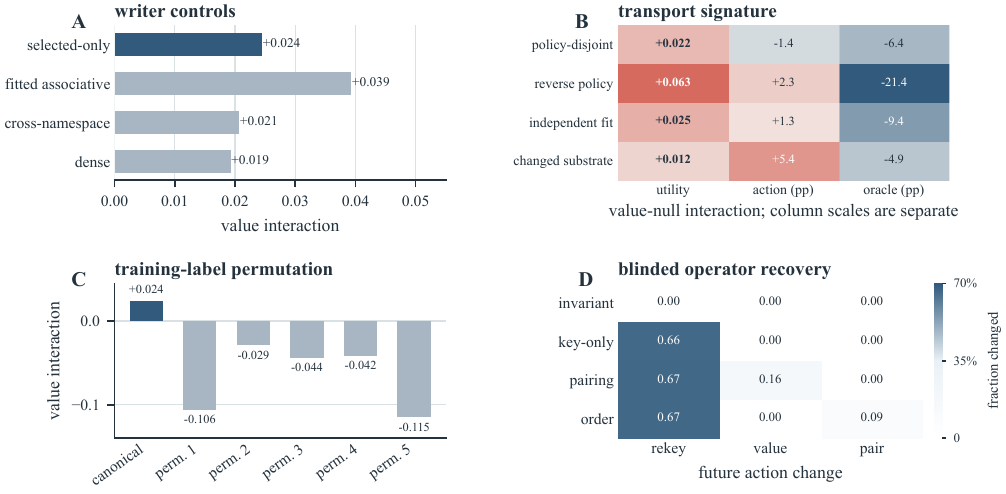}
\vspace{-4mm}
\caption{\textbf{Writer calibration and blinded operator recovery.} Panel A summarizes value-null interactions for the calibrated writer controls; panel B reports the transport signature across held-out fits; panel C checks training-label permutations; and panel D shows the action signature used to recover the hidden operator family. These are fixed-draw calibration summaries, with operator counts distinct from inferential replicate streams.}
\label{fig:writer-reconstruction}
\end{figure}

For an opaque writer $W$, the decoder uses the direct action signature
\begin{equation}
\mathbf{s}(W)=\bigl(\bar A_{\mathrm{rekey}},\bar D_{A,\mathrm{value}},
\bar D_{A,\mathrm{pair}}\bigr),
\label{eq:hidden-signature}
\end{equation}
where the first component uses the within-cell response $A$ and the latter two use the same-derangement cross-cell contrast $D_A$ defined in the main text. With threshold $10^{-6}$, the four predeclared classes have distinct observable signatures: zero rekey response is the invariant-class signature; rekey response alone is the key-only signature; value response with quiet pair transport is the commutative-pairing signature; and pair response is the order-sensitive signature. The decoder is applied before opaque identities are opened. Table~\ref{tab:hidden-operator-signatures} reports the resulting calibration outputs; Figure~\ref{fig:writer-reconstruction} visualizes the same calibration and holdout signature families. $n$ counts operator parameterizations, not inference clusters.

\begin{table}[ht]
\vspace{-4mm}
\centering
\small
\caption{Blinded operator-family signatures. Entries are direct action proportions averaged over three parameterizations, each evaluated on 48 replicate streams and nine outer arm derangements under the realized slot draw. $n$ counts operator parameterizations, not inference units. Correct is the number of parameterizations recovered by the predeclared rule; these are calibration outputs rather than population accuracy estimates.}
\label{tab:hidden-operator-signatures}
\begin{tabular}{lrrrrr}
\toprule
Family & $n$ & $\bar A_{\mathrm{rekey}}$ & $\bar D_{A,\mathrm{value}}$ & $\bar D_{A,\mathrm{pair}}$ & Correct \\
\midrule
Replay-invariant & 3 & 0.000 & 0.000 & 0.000 & 3 \\
Key-only & 3 & 0.664 & 0.000 & 0.000 & 3 \\
Commutative-pairing & 3 & 0.666 & 0.163 & 0.000 & 3 \\
Order-sensitive & 3 & 0.667 & 0.000 & 0.090 & 3 \\
\bottomrule
\end{tabular}
\vspace{-2mm}
\end{table}


All twelve operators are assigned to their generating class. The invariant class marks a resolution boundary: the future query may not probe information in the state, so a zero replay signature establishes equivalence under this protocol rather than an empty state.

\subsection{Held-out composite operators}
\label{app:hidden-operator-holdout}

The four-cell signature is tested on eight prefix-causal composites whose identities are also hidden during decoding. The held-out families combine value transport with temporal decay, key exposure with value accumulation, or key exposure with temporal decay. They use the same 48 streams, six strata, 48 tasks per stratum, nine derangements, and four cells as the canonical study. Let $e_\ell$ and $c(u)$ be as above, and let $v$ be either a scalar value direction or $(0.7,0.2,-0.3,-0.8)^{\mathsf T}$. The update laws are
\begin{equation}
\begin{aligned}
U_{\mathrm{vo}}(s,\ell,u)&=d s+\gamma c(u)v,\\
U_{\mathrm{kv}}(s,\ell,u)&=s+\bigl(\kappa+\gamma c(u)\bigr)e_\ell,\\
U_{\mathrm{ko}}(s,\ell,u)&=d s+\bigl(\kappa+\gamma c(u)\bigr)e_\ell.
\end{aligned}
\label{eq:holdout-composite-operators}
\end{equation}
The value--order family uses four parameterizations with $d\in\{0.10,0.65\}$; the key--value family uses $(\kappa,\gamma)\in\{(0.50,0.75),(1.00,1.25)\}$; and the key--order family uses $(d,\kappa,\gamma)\in\{(0.15,0.25,1.00),(0.60,0.25,1.00)\}$. The value--order readout uses the stated scalar or vector direction, while the key--value and key--order readouts select the largest state coordinate.

Table~\ref{tab:hidden-operator-holdout} reports direct action signatures conditional on $\sigma^\star$. Value--order operators respond to both value and pair transport, key--value operators retain rekey and value responses with quiet pair transport, and key--order operators respond to all three direct components. Here $n$ is the number of opaque operators in a family.

\begin{table}[ht]
\vspace{-4mm}
\centering
\small
\caption{Direct action signatures of eight held-out composite state operators. Rows pool opaque operators assigned to each update family; $n$ is the number of operators, not an inference sample size. Each operator is evaluated on 48 replicate streams and nine outer arm derangements under the realized slot draw. Entries are proportions of future actions changed and are calibration outputs for the declared composite families.}
\label{tab:hidden-operator-holdout}
\begin{tabular}{lrrrr}
\toprule
Family & $n$ & $\bar A_{\mathrm{rekey}}$ & $\bar D_{A,\mathrm{value}}$ & $\bar D_{A,\mathrm{pair}}$ \\
\midrule
Value--order & 4 & 0.000 & 0.162 & 0.162 \\
Key--value & 2 & 0.667 & 0.021 & 0.000 \\
Key--order & 2 & 0.665 & 0.160 & 0.142 \\
\bottomrule
\end{tabular}
\vspace{-2mm}
\end{table}


\subsection{Coefficient-randomized operators}
\label{app:random-operator-stress}

We randomize the state coefficients before assigning opaque identifiers. Twenty-four four-dimensional writers are drawn from two laws: twelve position-free commutative sums and twelve affine recurrent updates. For label $\ell$, draw independent vectors $b_\ell$ and $v_\ell$; the commutative law is $s_{t+1}=s_t+b_\ell+c(u_t)v_\ell$, and the recurrent law is $s_{t+1}=A s_t+b_\ell+c(u_t)v_\ell$ with spectral radius $0.62$. All writers use the canonical argmax readout and start at zero. The generating law is withheld during~signature~calculation.

Table~\ref{tab:random-operator-stress} reports the mean direct signatures and the terminal-state distance between aligned and pair-preserving cells. The commutative law has zero pair displacement and zero pair-action response for all twelve operators; the recurrent law has both quantities. These are fixed-draw values, and $n$ counts coefficient realizations rather than future tasks or inference clusters.

\begin{table}[ht]
\vspace{-4mm}
\centering
\small
\caption{Coefficient-randomized online operator stress test. Entries are mean future-action changes across 12 opaque operators and 48 replicate streams per operator under the realized slot draw. $n$ counts operators and does not define the inferential sample size. Pair-state is the terminal-state distance between aligned and pair-preserving cells, averaged over contexts. Action entries are proportions; Pair-state is a continuous distance.}
\label{tab:random-operator-stress}
\begin{tabular}{lrrrrr}
\toprule
Generating law & $n$ & $\bar A_{\mathrm{rekey}}$ & $\bar D_{A,\mathrm{value}}$ & $\bar D_{A,\mathrm{pair}}$ & Pair-state \\
\midrule
Commutative sum & 12 & 0.487 & 0.034 & 0.000 & 0.000 \\
Affine recurrent & 12 & 0.498 & 0.187 & 0.226 & 0.450 \\
\bottomrule
\end{tabular}
\vspace{-2mm}
\end{table}


\subsection{Readout and state-coordinate patterns}
\label{app:readout-invariance}

We test the pair-preserving terminal-state equality under deterministic changes of coordinates and future readout with 24 additional opaque online writers in state dimensions 4, 6, and 8. Twelve use position-free commutative sums and twelve use affine recurrent updates; each is evaluated with coordinate, orthogonal, and dense mixed readout bases. The coefficient and readout assignments are fixed before replay, while the future tasks, executor, 48 streams, nine derangements, and four cells~remain~shared.

Table~\ref{tab:readout-invariance} records the separation. Commutative writers have identical aligned and pair-preserving terminal states under all readouts, whereas recurrent writers show pair displacement and pair-action response. The result tests the zero law within the declared synthetic interface; it does not characterize unobserved state coordinates.

\begin{table*}[ht]
\vspace{-4mm}
\centering
\small
\caption{Pair-preserving responses under independently generated state dimensions and future readouts. Each entry pools twelve operators across 48 replicate streams under the realized slot draw; the operator count is a calibration count, not an inference sample size. Pair action is the fraction of future choices changed, Pair state is the terminal-state Euclidean distance, and $n_A$ and $n_S$ count operators with nonzero action and state responses, respectively.}
\label{tab:readout-invariance}
\begin{tabular*}{\textwidth}{@{\extracolsep{\fill}}lrrrrrrrr@{}}
\toprule
 & \multicolumn{4}{c}{\textbf{Commutative}} & \multicolumn{4}{c}{\textbf{Recurrent}} \\
\cmidrule(lr){2-5}\cmidrule(lr){6-9}
Readout & $D_A^{\rm pair}$ & State & $n_A$ & $n_S$ & $D_A^{\rm pair}$ & State & $n_A$ & $n_S$ \\
\midrule
Coordinate & 0.000 & 0.000 & 0 & 0 & 0.189 & 0.518 & 12 & 12 \\
Orthogonal & 0.000 & 0.000 & 0 & 0 & 0.215 & 0.518 & 12 & 12 \\
Mixed & 0.000 & 0.000 & 0 & 0 & 0.211 & 0.518 & 12 & 12 \\
\bottomrule
\end{tabular*}
\vspace{-2mm}
\end{table*}


\subsection{Independent nonlinear state-coordinate check}
\label{app:nonlinear-holdout}

We test the commutative update law under a nonlinear state representation and a readout chosen independently of the operator family. This holdout lies outside the canonical affine state representation. Twenty-four opaque prefix-causal writers are generated before their signatures are opened: twelve use a nonlinear commutative accumulator and twelve use a nonlinear recurrent update. Each writer is assigned a base dimension $d\in\{6,8,10\}$, randomized label and utility features, and a context-specific future readout. Coefficients are fixed before replay, and the same selected trace, future tasks, executor, outer derangements, and four replay cells are used for every writer.

For the commutative family, let $z_t\in\mathbb R^d$ be an accumulator and write the state as $s_t=(z_t,\tanh z_t)$. With an increment $\phi(\ell_t,u_t)$, the update is
\begin{equation}
z_{t+1}=z_t+\phi(\ell_t,u_t),
\quad
s_{t+1}=\bigl(z_{t+1},\tanh z_{t+1}\bigr).
\label{eq:nonlinear-commutative-holdout}
\end{equation}
The nonlinearity changes the state coordinates without changing the order-free sum. Complete-pair transport therefore leaves the terminal state equal to the aligned state, up to floating-point tolerance. For the recurrent family, with $s_t\in\mathbb R^d$, a randomized stable matrix $A$, and a normalized matrix $B$, the update is
\begin{equation}
s_{t+1}=\tanh\!\left(A s_t+\phi(\ell_t,u_t)
       +0.08\,s_t\odot(Bs_t)\right).
\label{eq:nonlinear-recurrent-holdout}
\end{equation}
The state-dependent interaction makes the order of supplied pairs part of the writer relation. The future readout is held fixed within each context and is the only route from the terminal state to a future action; no future outcome or terminal batch is supplied to the update.

The increment has the form $\phi(\ell,u)=b_\ell+(u-1/2)v_\ell$, where the label features $b_\ell$ and value features $v_\ell$ are independently drawn before the operator is assigned an opaque identifier. The recurrent transition matrix is scaled to spectral radius $0.42$, and $B$ is normalized by its spectral norm before the displayed cross-state term is applied. For each future context $c$, a matrix $Q_c$ is drawn with one row per arm and row-normalized before replay; the action is $\operatorname{TieBreak}\arg\max\nolimits_a(Q_cs)_a$, using the canonical arm order for exact ties. The matrices $Q_c$ are fixed across all four cells and are chosen independently for each context. Thus the readout has access only to the terminal state and the held-fixed~future~context.

For aligned and pair cells, the reported pair-state quantity is the average Euclidean distance
\begin{equation}
\bar d_{\mathrm{pair}}=
\frac{1}{|\mathcal C|}\sum\nolimits_{c\in\mathcal C}
\left\|s^{\mathrm{aligned}}_c-s^{\mathrm{pair}}_c\right\|_2,
\label{eq:nonlinear-pair-state-distance}
\end{equation}
subsequently averaged over the paired outer derangements and replicate streams. A writer is counted as pair-zero when this distance and its pair-action response are at most $10^{-9}$ for every evaluated replicate. The family identity and all coefficient assignments are opened only after these direct signatures have been computed. For the recurrent rows, the pair response is a joint pairing--order--exposure contrast; the pairing-specific zero-law interpretation is reserved for the commutative rows.

At the direct action level, the commutative state equality implies $D_{A,\mathrm{pair}}=0$ under the same outer derangement, because the pair and aligned readouts receive the same terminal state. This is the action-level quantity reported in the pair column of Table~\ref{tab:nonlinear-operator-holdout}.

Table~\ref{tab:nonlinear-operator-holdout} reports the direct signatures. The commutative family has zero pair-action response for all twelve writers and a terminal-state distance at numerical precision, whereas the recurrent family has nonzero pair displacement and future-action response. The rekey and value columns show that both families receive the same outer intervention and value-null controls; the pair-preserving response differs between them. Under an independent within-block slot draw, the commutative pair response remains zero and the recurrent response remains nonzero (Table~\ref{tab:nonlinear-operator-holdout}). The two rows per draw reuse the same operators, replicate streams, future support, and executor, so the second draw is a sensitivity condition rather than an additional operator or inferential cluster. The result is conditional on the declared synthetic interface and future support.

\begin{table}[ht]
\vspace{-4mm}
\centering
\small
\caption{Nonlinear state-coordinate checks under complete-pair transport. Each writer law uses 12 operator instances, reused across draws with the same 48 streams, nine outer maps, and six future strata; Draw B uses an independently sampled within-block permutation. The State entry is mean terminal-state distance; Pair, Value, and Rekey are changed-action proportions. State distances below $10^{-9}$ are displayed as zero.}
\label{tab:content-nonlinear}
\label{tab:nonlinear-operator-holdout}
\begin{tabular*}{\textwidth}{@{\extracolsep{\fill}}lrrrrrrrr@{}}
\toprule
 & \multicolumn{4}{c}{\textbf{Draw A}} & \multicolumn{4}{c}{\textbf{Draw B}} \\
\cmidrule(lr){2-5}\cmidrule(lr){6-9}
\textbf{Writer law} & \textbf{State} & \textbf{Pair} & \textbf{Value} & \textbf{Rekey} & \textbf{State} & \textbf{Pair} & \textbf{Value} & \textbf{Rekey} \\
\midrule
Nonlinear commutative & 0.000 & 0.000 & 0.030 & 0.465 & 0.000 & 0.000 & 0.031 & 0.465 \\
Nonlinear recurrent & 0.439 & 0.218 & 0.180 & 0.510 & 0.441 & 0.221 & 0.185 & 0.510 \\
\bottomrule
\end{tabular*}
\vspace{-2mm}
\end{table}

\subsection{Family-blind calibration of the pair law}
\label{app:blind-relation-challenge}

\begin{table}[ht]
\vspace{-4mm}
\centering
\small
\caption{Family-blind calibration of the complete-pair law. The decoder classifies a writer as pair-displacing when its mean pair-action response is at least $0.050$; this threshold is fixed before coefficients, readouts, and hidden assignments are generated. Each draw contains 16 pair-preserving and 16 pair-displacing writers. Total is the denominator of the Correct count; accuracy is a proportion.}
\label{tab:content-blind}
\begin{tabular*}{\columnwidth}{@{\extracolsep{\fill}}llrrr@{}}
\toprule
\textbf{Draw} & \textbf{Target property} & \textbf{Total} & \textbf{Correct} & \textbf{Accuracy} \\
\midrule
A & Pair-preserving & 16 & 16 & 1.000 \\
A & Pair-displacing & 16 & 15 & 0.938 \\
B & Pair-preserving & 16 & 16 & 1.000 \\
B & Pair-displacing & 16 & 16 & 1.000 \\
\midrule
A+B & All challenge writers & 64 & \textbf{63} & \textbf{0.984} \\
\bottomrule
\end{tabular*}
\vspace{-2mm}
\end{table}

The preceding holdouts test named operator combinations. This challenge tests the observable pair-law predicate without reusing those family names. The generating update defines the targets: a pair-preserving law leaves its complete-pair sum unchanged, whereas a pair-displacing law includes an order- or state-dependent term. Before the hidden assignments are opened, the decoder is fixed to classify an operator as pair-displacing when its mean direct pair-action signature is at least $0.05$; all smaller values are classified as pair-preserving. The threshold is predeclared before coefficient generation and hidden assignment, and is not fit to challenge outputs, operator scores, or family labels. The prediction uses only the future-action contrast between the aligned and pair cells, pooled over the nine outer derangements and 48 replicate streams. Terminal-state distance is recorded as a diagnostic and is not supplied to the decoder. The algebraic target and the resolution threshold are separate: the former comes from the update algebra, and the latter fixes what the declared future support~can~resolve.

The challenge contains four new prefix-causal laws, with eight opaque operators per law and base dimensions $d\in\{6,8,10\}$. The first law keeps a nonlinear feature sum, $g_{t+1}=g_t+\psi(\ell_t,u_t)$, and exposes $s=\tanh(Mg+b)$ only after the adaptation prefix. The second keeps a symmetric second-order statistic, $G_{t+1}=G_t+\psi(\ell_t,u_t)\psi(\ell_t,u_t)^\top$, followed by an independent nonlinear projection. Both are pair-preserving: complete-pair transport changes the order of summands but not the terminal statistic. The feature map is
\begin{equation}
\psi(\ell,u)=\sin\!\left(b_\ell+(u-1/2)v_\ell
             +0.15\bigl(b_\ell+(u-1/2)v_\ell\bigr)^2\right).
\label{eq:blind-challenge-feature-map}
\end{equation}

The third law is a bilinear recurrence, $h_{t+1}=\tanh(Ah_t+\psi_t+0.14\,\psi_t\odot(Bh_t))$; the fourth is a position-mixed accumulator, $h_{t+1}=h_t+W_{j\bmod 5}\psi_t$, where $j$ is the prefix position. These two laws are pair-displacing through a state-dependent interaction and a position-dependent linear operator, respectively. They are outside the canonical key/value/order update grammar and were not used to set the decoder threshold. Each operator receives independently generated coefficients and a context-specific row-normalized future readout. Coefficients, readouts, the selected trace, future tasks, executor, and outer derangements are fixed before replay; future outcomes and terminal batches are never supplied to the update.

Table~\ref{tab:blind-relation-challenge} reports Draw A, the held-out operator-level calibration draw. The pair-preserving laws are separated without a false positive. The pair-displacing laws produce 15 correct decisions out of 16; the remaining operator has a mean pair-action signature of $0.047$, just below the frozen threshold, while its terminal-state distance is $0.414$. The latent state changes, but the declared future support does not expose that change strongly enough for the observable decoder, so we count this case as a~false~negative.

We repeat the complete challenge with independently generated operator coefficients, context-specific readouts, and hidden assignments while retaining the same fixed trace, future support, executor, outer derangements, decoder, and threshold. Draw B separates all 32 writers; together the two draws give 63 correct decisions out of 64, with no false positive (Table~\ref{tab:content-blind}). These are finite calibration results for the declared interface, not a population-generalization rate.

\begin{table*}[ht]
\vspace{-4mm}
\centering
\small
\caption{Family-blind calibration of the construction-defined pair law on four held-out writer laws. The decoder uses the mean direct pair-action signature and a threshold fixed before the hidden assignment; pair state is a diagnostic, not a ranking criterion. Pair action and calibration accuracy are proportions; Pair-state distance is a continuous diagnostic; $n$ is the number of writers.}
\label{tab:blind-relation-challenge}
\begin{tabular*}{\textwidth}{@{\extracolsep{\fill}}lrrrr@{}}
\toprule
Target relation & $n$ & Pair action & Pair state & Calibration accuracy \\
\midrule
Pair-preserving & 16 & 0.000 & 0.000 & 1.000 \\
Pair-displacing & 16 & 0.161 & 0.470 & 0.938 \\
All challenge writers & 32 & 0.081 & 0.235 & 0.969 \\
\bottomrule
\end{tabular*}
\vspace{-2mm}
\end{table*}

The threshold is a resolution convention for the observable action signature, not an estimator of a population prevalence. Re-evaluating the same fixed signatures over a fixed descriptive grid shows that the zero false-positive result is unchanged, while sensitivity falls as the resolution requirement is raised (Table~\ref{tab:blind-threshold-sensitivity}). We therefore report $0.05$ as the fixed calibration threshold.

\begin{table}[ht]
\vspace{-4mm}
\centering
\small
\caption{Threshold sensitivity for the fixed blind challenge. Each row re-evaluates the same 32 operator-level signatures; no operator, law, or readout is refit. Specificity and sensitivity are computed against the construction-defined pair relation.}
\label{tab:blind-threshold-sensitivity}
\begin{tabular}{rrrr}
\toprule
Threshold & Accuracy & Specificity & Sensitivity \\
\midrule
0.025 & 1.000 & 1.000 & 1.000 \\
0.050 & 0.969 & 1.000 & 0.938 \\
0.075 & 0.906 & 1.000 & 0.813 \\
0.100 & 0.906 & 1.000 & 0.813 \\
0.125 & 0.844 & 1.000 & 0.688 \\
0.150 & 0.781 & 1.000 & 0.563 \\
\bottomrule
\end{tabular}
\vspace{-2mm}
\end{table}

\section{Continuous representation controls}
\label{app:representation-controls}

\subsection{Dense-state control details}
\label{app:dense-control}

The dense writer replaces the explicit family--context--arm ledger with one continuous vector per family--context pair. For every replay cell it receives the supplied pair $(\ell_t^q,u_t^q)$ and updates \(h_{t+1}=0.985h_t+(u_t^q-0.5)e_{\ell_t^q}\); its future choice maximizes \(h^{\mathsf T}e_a\) over the four future-action features. The feature vectors are fixed within each replicate and shared across all paired interventions. The writer has no arm-indexed count or utility table; this is the dense-state construction used for the comparison.

\begin{table}[ht]
\vspace{-4mm}
\centering
\small
\caption{Dense continuous-state interactions on held-out streams. Each comparison uses the same 48 replicate streams and nine paired derangements. Utility is identity minus replay; action and executor-relative oracle entries are normalized map-minus-identity differences. Entries are fixed-draw calibration point estimates.}
\label{tab:dense-associative-interactions}
\begin{tabular}{@{}lrrr@{}}
\toprule
Contrast & Utility & Action & Oracle \\
\midrule
OCRR $-$ value null & 0.019 & 0.027 & $-0.055$ \\
OCRR $-$ time placebo & 0.028 & 0.015 & $-0.063$ \\
Pair placebo $-$ value null & 0.021 & 0.027 & $-0.060$ \\
Value null $-$ time placebo & 0.008 & $-0.012$ & $-0.008$ \\
\bottomrule
\end{tabular}
\vspace{-2mm}
\end{table}

The OCRR--value-null interaction is $+0.019$ in future utility, with interval $[+0.010,+0.030]$ and cluster sign-flip $p=0.00010$. Its oracle interaction is $-5.5\,\text{pp}$, with interval $[-7.9,-3.4]$. The pair-preserving cell gives $+0.021$ utility with interval $[+0.012,+0.031]$ and action change $+2.7\,\text{pp}$ with interval $[+1.3,+4.3]$; its oracle change is $-6.0\,\text{pp}$ with interval $[-8.2,-4.0]$ relative to the value null. The OCRR--value-null action change is $+2.7\,\text{pp}$ with interval $[+1.3,+4.3]$. The value-null--time-placebo utility difference is $+0.008$, with interval $[-0.000,+0.017]$, and its action change is $-1.2\,\text{pp}$ with interval $[-2.6,+0.1]$; the OCRR--time-placebo action change is $+1.5\,\text{pp}$ with interval $[+0.0,+3.0]$. This comparison provides no evidence that the utility interactions of the two placebo~cells~differ.

\subsection{Cross-namespace control details}
\label{app:cross-namespace-control}

The cross-namespace writer uses separate producer and future-action identifiers. Let $\mathcal K_P$ and $\mathcal K_A$ be disjoint copies of the arm space, and let $\phi:\mathcal A\to\mathcal K_P$ and $\chi:\mathcal A\to\mathcal K_A$ be bijections. For each replicate, independent unit action features $q_a\in\mathbb R^{32}$ are transformed by an orthogonal binding $B$ into producer-side features $p_a=Bq_a$. The writer receives a producer key $\phi(a)$ and updates one continuous state vector with the feature associated with that key. The future policy scores action $a$ with the bound query $p_a$. The producer and action identifier strings are disjoint, the action-side key is used only by the future readout, and the binding is fixed before replay. The intervention therefore changes the stored producer key while leaving the future action interface unchanged. Equivalently, with $\psi$ denoting the producer-key lookup, the update and future readout are
\begin{equation}
\resizebox{0.96\linewidth}{!}{$\displaystyle
 h_{t+1}=0.985h_t+(u_t^q-0.5)\psi(\ell_t^q),\quad
 \pi(h)=\operatorname{TieBreak}\arg\max\nolimits_a h^{\mathsf T}p_a,\quad
 \psi(\phi(a))=p_a=Bq_a .
$}
\label{eq:cross-namespace-interface}
\end{equation}
The dense control is the special case with identity binding and shared producer and action keys; the cross-namespace control keeps the two key spaces disjoint while retaining the same vector binding. Since the decay is below one, this writer is an order-sensitive continuous control rather than an instance of the commutative proposition.

\begin{table*}[ht]
\vspace{-4mm}
\centering
\small
\caption{Uncertainty for continuous-writer pairing interactions. Estimate is the point estimate, while Lower and Upper are separate 95\% replicate-cluster bootstrap interval endpoints; utility, action, and oracle estimates and interval endpoints are normalized differences. The cluster sign-flip $p$ value is reported for the utility interaction; NR means not reported.}
\label{tab:cross-namespace-uncertainty}
\begin{tabular*}{\textwidth}{@{\extracolsep{\fill}}lrrrrrrrr@{}}
\toprule
 & \multicolumn{4}{c}{\textbf{Dense}} & \multicolumn{4}{c}{\textbf{Cross-namespace}} \\
\cmidrule(lr){2-5}\cmidrule(lr){6-9}
Readout & Estimate & Lower & Upper & $p$ & Estimate & Lower & Upper & $p$ \\
\midrule
$U$ & 0.019 & 0.010 & 0.030 & $<0.001$ & 0.021 & 0.012 & 0.030 & $<0.001$ \\
$A$ & 0.027 & 0.013 & 0.043 & NR & 0.019 & 0.009 & 0.030 & NR \\
$O$ & $-0.055$ & $-0.079$ & $-0.034$ & NR & $-0.069$ & $-0.091$ & $-0.048$ & NR \\
\bottomrule
\end{tabular*}
\vspace{-2mm}
\end{table*}

The cross-namespace writer gives a utility interaction of $+0.021$ with interval $[+0.012,+0.030]$ and sign-flip $p=0.00005$. Its action interaction is $+1.9\,\text{pp}$ with interval $[+0.9,+3.0]$, and its oracle interaction is $-6.9\,\text{pp}$ with interval $[-9.1,-4.8]$. Against the time placebo, the corresponding utility interaction is $+0.029$ with interval $[+0.020,+0.038]$ and the oracle interaction is $-7.7\,\text{pp}$ with interval $[-9.9,-5.5]$. The same ordering holds when the state update and future action use distinct identifier namespaces; it is conditional on the fixed continuous binding and executor. \noindent\textbf{Online commutative namespace control.}

To calibrate the pairing interpretation with an explicitly online commutative writer, we use disjoint producer and future-action key spaces and the update \(h=\sum\nolimits_t (u_t^q-0.5)p_{\ell_t^q}\) and a fixed linear future readout. The writer receives one supplied label--utility pair at each prefix step, the future is frozen before readout, and the same nine outer maps and blockwise pair transport are applied to all 48 replicate streams. Its value-null interaction is $+0.023$ (95\% cluster interval $[+0.013,+0.034]$), with an action interaction of $+1.7$ percentage points (interval $[+0.5,+3.0]$) and an oracle-match interaction of $-7.7$ percentage points (interval $[-10.0,-5.4]$). The cluster sign-flip values are $0.00015$, $0.0126$, and $0.00005$, respectively. Pair transport produces zero future-action mismatches for every outer map and replicate stream, as required by the commutative update.

\section{Fitted associative controls}
\label{app:fitted-controls}

We fit a low-rank associative writer on streams disjoint from the evaluation streams. During replay it maintains the continuous state \(h=\sum\nolimits_t (u_t^q-0.5)p_{\ell_t^q}\), where producer features \(p_a\) and future-action queries \(q_a\) are fitted jointly from selected adaptation outcomes and counterfactual future utilities. The future readout is a bounded score \(\operatorname{sigmoid}(q_a^{\mathsf T}h)\). The sum is commutative: reordering the same label--utility pairs leaves \(h\) unchanged. This fitted representation control is order-invariant, while its training target remains the synthetic executor's future utility.

The fitting streams provide selected outcomes and counterfactual future utility targets for the four future actions. The feature dimension is eight; feature and query arrays are optimized with squared-norm regularization and frozen before evaluation. No evaluation-stream future utility enters the fit, and uncertainty is quantified across evaluation replicate streams conditional on the fitted writer. Refit rows vary only the fitted representation; the replay law, future support, and evaluation streams~remain~fixed.

\begin{table}[ht]
\vspace{-4mm}
\centering
\small
\caption{Interactions for a privileged fitted low-rank associative state writer on held-out streams. Utility is identity minus replay; action and oracle entries are normalized map-minus-identity differences for the named contrast.}
\label{tab:learned-associative-interactions}
\begin{tabular}{@{}lrrr@{}}
\toprule
Contrast & Utility & Action & Oracle \\
\midrule
Fitted $-$ value null & 0.039 & 0.018 & $-0.094$ \\
Fitted $-$ time placebo & 0.039 & 0.019 & $-0.090$ \\
Pair placebo $-$ value null & 0.039 & 0.018 & $-0.094$ \\
\bottomrule
\end{tabular}
\vspace{-2mm}
\end{table}

For the fitted writer, the value-permutation interaction is $+0.039$ in future utility, with 95\% interval $[+0.028,+0.052]$ and cluster sign-flip $p=0.00005$. The corresponding action interaction is $+1.8$ percentage points with interval $[+0.7,+2.8]$, and the oracle interaction is $-9.4$ percentage points with interval $[-11.8,-7.2]$. Against the time placebo, the utility, action, and oracle interactions are $+0.039$, $+1.9$ percentage points, and $-9.0$ percentage points, with intervals $[+0.027,+0.051]$, $[+0.8,+3.0]$, and $[-11.3,-6.7]$, respectively. The pair-preserving cell is identical to the fitted aligned cell at the state level because the update is commutative; its corresponding interactions are $+0.039$, $+1.8$ percentage points, and $-9.4$ percentage points, with intervals $[+0.028,+0.052]$, $[+0.7,+2.8]$, and $[-11.7,-7.2]$. This invariance isolates the fitted writer's response to pairing from temporal placement within the fixed trace.

\subsection{Selected-outcome-only terminal-batch control}

The selected-outcome-only writer retains the state-update law while restricting the information boundary. For a training block $e\in\mathcal{E}_{\mathrm{sel}}$, let $(a_{e,i},y_{e,i})_{i=1}^{48}$ be its selected arms and scalar outcomes. Define the commutative terminal state and its leave-one-out state by
\begin{equation}
h_e=\sum\nolimits_{j=1}^{48}(y_{e,j}-0.5)p_{a_{e,j}},\quad
h_{e,-i}=h_e-(y_{e,i}-0.5)p_{a_{e,i}},\quad
\widehat y_{e,i}=\operatorname{sigmoid}\!\left(q_{a_{e,i}}^{\mathsf T}h_{e,-i}\right).
\label{eq:selected-only-fit}
\end{equation}
The current selected outcome is the target, while the predictor state contains the other selected outcomes through a commutative sum. Because the sum uses the entire block, $h_{e,-i}$ can include outcomes later than $i$ in the realized trace. This is a terminal-batch representation on a fixed trace; the sequential recurrent writer below supplies the online temporal branch. The fitted parameters minimize
\begin{equation}
\mathcal{L}_{\mathrm{sel}}(P,Q)=
\frac{1}{|\mathcal{E}_{\mathrm{sel}}|}
\sum\nolimits_{e\in\mathcal{E}_{\mathrm{sel}}}\sum\nolimits_{i=1}^{48}
\frac{1}{2}\left(\widehat y_{e,i}-y_{e,i}\right)^2
 +\frac{\lambda}{2}\left(\lVert P\rVert_F^2+\lVert Q\rVert_F^2\right),
\quad \lambda=10^{-4}.
\label{eq:selected-only-loss}
\end{equation}
Here $P=\{p_a\}$ and $Q=\{q_a\}$ are eight-dimensional producer and future-action arrays, and $\operatorname{sigmoid}$ is the logistic function. The objective uses disjoint training blocks and no counterfactual future utility, oracle label, or evaluation-stream row. The centered terminal state is evaluated only after the parameters are frozen. Five refits vary the representation while leaving the replay law and evaluation~streams~unchanged.

The selected-outcome-only writer gives a value-null interaction of $+0.024$ with interval $[+0.015,+0.035]$ and sign-flip $p<0.0001$. Its action interaction is $+0.3$ percentage points with interval $[-0.8,+1.5]$, and its executor-relative oracle interaction is $-8.9$ percentage points with interval $[-11.3,-6.5]$. The time-placebo interaction has utility $+0.024$ with interval $[+0.015,+0.033]$ and oracle interaction $-8.9$ percentage points with interval $[-11.2,-6.6]$. Because the update is commutative, the aligned and pair-preserving terminal states are equal. The paired-state check records zero future-action mismatches under every outer map, and the two cells share the same replicate-level estimate vector and resampling law.

\begin{table}[ht]
\vspace{-4mm}
\centering
\small
\caption{Order-invariant selected-outcome-only commutative state control on held-out streams. The writer is fitted with a leave-one-out commutative selected-outcome objective and receives no counterfactual future utility during fitting. Each row is a readout-specific normalized estimate; intervals are 95\% replicate-cluster bootstrap intervals with separate Lower and Upper endpoints, and $p$ is the two-sided cluster sign-flip value.}
\label{tab:selected-only-associative-interactions}
\begin{tabular}{@{}llrrrr@{}}
\toprule
Contrast & Readout & Estimate & Lower & Upper & $p$ \\
\midrule
Selected-only $-$ value null & Utility & 0.024 & 0.015 & 0.035 & $<0.001$ \\
Selected-only $-$ value null & Action & 0.003 & $-0.008$ & 0.015 & 0.596 \\
Selected-only $-$ value null & Oracle & $-0.089$ & $-0.113$ & $-0.065$ & $<0.001$ \\
Selected-only $-$ time placebo & Utility & 0.024 & 0.015 & 0.033 & $<0.001$ \\
Selected-only $-$ time placebo & Action & 0.005 & $-0.007$ & 0.016 & 0.467 \\
Selected-only $-$ time placebo & Oracle & $-0.089$ & $-0.112$ & $-0.066$ & $<0.001$ \\
Pair placebo $-$ value null & Utility & 0.024 & 0.015 & 0.035 & $<0.001$ \\
Pair placebo $-$ value null & Action & 0.003 & $-0.008$ & 0.015 & 0.596 \\
Pair placebo $-$ value null & Oracle & $-0.089$ & $-0.113$ & $-0.065$ & $<0.001$ \\
\bottomrule
\end{tabular}
\vspace{-2mm}
\end{table}

The terminal state is commutative, so a fitting objective that follows a chosen prefix order would introduce an avoidable training convention. We compare the prefix objective under arm-sorted and realized chronological orders with the leave-one-out objective used here. The prefix variant predicts each selected outcome from the state before that observation; its terminal evaluation still uses the same commutative sum. Its contrast changes sign under the two orders, whereas the leave-one-out contrast remains positive across the five fits (Table~\ref{tab:selected-only-fit-objective}). The prefix-causal operator family described earlier supplies the online counterpart. The replay law, evaluation streams, and inference units~remain~fixed.

\begin{table}[ht]
\vspace{-4mm}
\centering
\small
\caption{Sensitivity to the selected-only fitting objective. Each row summarizes five refits on the same disjoint training streams. $U$ is the normalized utility interaction; Minimum and Maximum are the lower and upper endpoints of the observed range across refits, not confidence limits. $O$ is the normalized oracle interaction; mismatch counts aligned-versus-pair-preserving action disagreements. The mismatch count holds for every refit and each of nine nonidentity maps separately, with $48\times6\times48=13{,}824$ comparisons per map; the identity map is excluded.}
\label{tab:selected-only-fit-objective}
\begin{tabular}{@{}llrrrrr@{}}
\toprule
Fit & Order & Utility & Minimum & Maximum & Oracle & Pair mismatch count \\
\midrule
Prefix & Arm-sorted & 0.021 & 0.016 & 0.027 & $-0.066$ & 0 \\
Prefix & Chronological & $-0.008$ & $-0.012$ & $-0.005$ & 0.015 & 0 \\
Leave-one-out & Order-invariant & 0.023 & 0.021 & 0.027 & $-0.078$ & 0 \\
\bottomrule
\end{tabular}
\vspace{-2mm}
\end{table}

\subsection{Initialization stability of the selected-only fit}
\label{app:selected-only-fit-stability}

The selected-outcome-only control was refit five times with the same disjoint training and held-out evaluation streams, state-update law, and intervention maps. Each parameter set was frozen before evaluation. The utility interaction remains positive across all five fits, the pair-preserving state remains action-invariant, and the spread is attributable to fitting variation rather than a change in the replay or future boundary. Table~\ref{tab:selected-only-fit-stability} reports these structural readouts; task-level uncertainty remains clustered by replicate stream.

\begin{table}[ht]
\vspace{-4mm}
\centering
\small
\caption{Refit stability of the order-invariant selected-outcome-only commutative writer. Every fit uses the leave-one-out commutative objective, selected adaptation outcomes only, and the same disjoint training and held-out evaluation streams. Utility intervals are 95\% replicate-cluster bootstrap intervals with separate Lower and Upper endpoints; action and oracle entries are normalized differences in future-action and executor-relative oracle response. Pair mismatch counts compare aligned and pair-preserving future actions. The mismatch count holds for each of nine nonidentity maps separately, with $48\times6\times48=13{,}824$ comparisons per map; the identity map is excluded.}
\label{tab:selected-only-fit-stability}
\begin{tabular}{@{}lrrrrrr@{}}
\toprule
Refit & Utility & Lower & Upper & Action & Oracle & Pair mismatch count \\
\midrule
R1 & 0.024 & 0.015 & 0.035 & 0.003 & $-0.089$ & 0 \\
R2 & 0.023 & 0.013 & 0.033 & 0.003 & $-0.080$ & 0 \\
R3 & 0.021 & 0.012 & 0.031 & 0.006 & $-0.075$ & 0 \\
R4 & 0.027 & 0.018 & 0.036 & $-0.002$ & $-0.071$ & 0 \\
R5 & 0.021 & 0.012 & 0.030 & 0.004 & $-0.075$ & 0 \\
\bottomrule
\end{tabular}
\vspace{-2mm}
\end{table}

\subsection{Permutation-draw sensitivity}
\label{app:permutation-draw}

The structural estimand is conditional on the realized within-block draw, so we evaluate four fixed realizations of the same blockwise law while keeping the selected-only fit, training and evaluation streams, future tasks, and outer maps fixed. The value-null and time-placebo interactions remain positive across these realizations for the order-invariant selected-only writer, and the pair-preserving action remains invariant. Table~\ref{tab:permutation-draw-sensitivity} reports the estimates and replicate-cluster intervals. The draw IDs use the predeclared time/value--pair seed pairs $(20260822,20260824)$, $(20260901,20260901)$, $(20260902,20260902)$, and $(20260903,20260903)$, respectively; the fit, trace, future support, outer maps, and evaluator are shared across all four draws. This four-draw check complements the five-draw primary operator-robustness check in Appendix~\ref{app:structural-results}.

\begin{table*}[ht]
\vspace{-4mm}
\centering
\small
\caption{Sensitivity of the order-invariant selected-outcome-only commutative contrast to the within-block permutation draw. Four fixed realizations of the same blockwise law use the same fitted writer and evaluation streams. Value and time are the value-null and time-placebo interactions, respectively, reported as normalized estimates with separate 95\% replicate-cluster Lower and Upper endpoints. Pair-preserving action mismatch counts are shown in the last column. The mismatch count holds for each of nine nonidentity maps separately, with $48\times6\times48=13{,}824$ comparisons per map; the identity map is excluded.}
\label{tab:permutation-draw-sensitivity}
\begin{tabular*}{\textwidth}{@{\extracolsep{\fill}}rrrrrrrr@{}}
\toprule
 & \multicolumn{3}{c}{\textbf{Value}} & \multicolumn{3}{c}{\textbf{Time}} \\
\cmidrule(lr){2-4}\cmidrule(lr){5-7}
Draw & Estimate & Lower & Upper & Estimate & Lower & Upper & Pair mismatch \\
\midrule
0 & 0.024 & 0.015 & 0.035 & 0.024 & 0.015 & 0.033 & 0 \\
1 & 0.028 & 0.018 & 0.039 & 0.026 & 0.016 & 0.036 & 0 \\
2 & 0.024 & 0.014 & 0.033 & 0.025 & 0.016 & 0.035 & 0 \\
3 & 0.029 & 0.020 & 0.039 & 0.021 & 0.014 & 0.030 & 0 \\
\bottomrule
\end{tabular*}
\vspace{-2mm}
\end{table*}

\subsection{Learned recurrent state-writer check}
\label{app:learned-recurrent-check}

To test whether the protocol's interpretation depends on an imposed commutative update law, we fit a nonlinear recurrent writer on the same selected-outcome information boundary and evaluate it on disjoint streams. Its state update is
\begin{equation}
 h_{i+1}=\tanh(Ah_i+p_{a_i}+(y_i-0.5)v_{a_i}).
\label{eq:learned-recurrent-update}
\end{equation}
with learned transition matrix $A$, label inputs $p_a$, utility inputs $v_a$, and future-action queries. The fitting objective predicts the current selected outcome from the pre-update state, and the frozen evaluation writer updates its state sequentially in the realized adaptation order while the evaluation future remains fixed. The transition is order-sensitive and no pair-preserving equality is assumed; the state has dimension eight and is refit under the same regularized objective.

Table~\ref{tab:learned-recurrent-check} reports the initial conditional interactions, and Table~\ref{tab:learned-recurrent-fit-stability} reports refit sensitivity. The initial value-null interaction is inconclusive, whereas the time-placebo response is positive. Across the five fits, this check covers the order-sensitive branch of the state-boundary interface, with uncertainty reported for the fitted representation.

\begin{table}[ht]
\vspace{-4mm}
\centering
\small
\caption{Learned recurrent state-writer boundary check for a reference fit. The writer is fit only to selected adaptation outcomes on disjoint training blocks and is frozen before evaluation on disjoint streams. Entries are normalized future-utility interaction estimates with 95\% replicate-cluster Lower and Upper endpoints; Action is the corresponding interaction in the normalized future-action fraction. The nonlinear recurrent transition imposes no commutativity.}
\label{tab:learned-recurrent-check}
\begin{tabular}{@{}lrrrr@{}}
\toprule
Contrast & Utility & Lower & Upper & Action \\
\midrule
Value null & 0.001 & $-0.001$ & 0.003 & 0.001 \\
Time placebo & 0.005 & 0.001 & 0.009 & $-0.012$ \\
Pair minus value & 0.000 & $-0.003$ & 0.002 & $-0.005$ \\
\bottomrule
\end{tabular}
\vspace{-2mm}
\end{table}

\begin{table}[ht]
\vspace{-4mm}
\centering
\small
\caption{Refit sensitivity of the learned recurrent boundary check. Each row contains one refit, with separate Value and Time estimates and their 95\% replicate-cluster Lower and Upper endpoints; Value and Time denote the corresponding normalized utility interactions.}
\label{tab:learned-recurrent-fit-stability}
\begin{tabular*}{\textwidth}{@{\extracolsep{\fill}}lrrrrrr@{}}
\toprule
 & \multicolumn{3}{c}{\textbf{Value}} & \multicolumn{3}{c}{\textbf{Time}} \\
\cmidrule(lr){2-4}\cmidrule(lr){5-7}
Refit & Estimate & Lower & Upper & Estimate & Lower & Upper \\
\midrule
R1 & 0.001 & $-0.001$ & 0.003 & 0.005 & 0.001 & 0.009 \\
R2 & 0.000 & $-0.004$ & 0.004 & 0.001 & $-0.003$ & 0.005 \\
R3 & 0.001 & $-0.002$ & 0.003 & 0.000 & $-0.001$ & 0.002 \\
R4 & 0.003 & $-0.001$ & 0.007 & 0.002 & $-0.001$ & 0.006 \\
R5 & $-0.003$ & $-0.006$ & 0.001 & $-0.003$ & $-0.007$ & 0.001 \\
\bottomrule
\end{tabular*}
\vspace{-2mm}
\end{table}

\subsection{Independent learned-interface check}
\label{app:independent-learned-interface}

The preceding recurrent check uses the primary substrate. We evaluate the same state-boundary question on an independently generated sequence substrate whose future utility is produced by an order-sensitive latent recurrence. An adaptation episode contains 32 selected key--utility pairs followed by 48 fixed future queries. A generic tanh recurrent encoder learns separate key and utility inputs and a future action--query readout from 256 training episodes. The regularized objective is optimized on those episodes, and the 16-dimensional parameters are frozen before evaluation on 48 disjoint episodes. Neither evaluation sequence, future query, nor evaluation target is used during the fit, and the update is not constrained to commute or to preserve complete pairs.

For each evaluation episode, the future query set and executor utilities are held fixed. Rekeying moves producer keys across their original slots, value transport permutes utilities across slots, and pair transport moves complete key--utility pairs. Nine fixed-point-free key maps are applied to each frozen fit. The table reports the direct aligned-minus-pair response in future utility, future-action change, and executor-relative oracle agreement. The utility response is positive and the oracle response negative for every fit, with the action response positive throughout. The table below reports each fit and their~arithmetic~mean.

\begin{table}[ht]
\vspace{-4mm}
\centering
\small
\caption{Direct pair-transport response of the independently learned writer. Each row is one frozen fit, evaluated on the same 48 disjoint episodes and nine derangements. $U$, $A$, and $O$ are aligned-minus-pair future utility, future-action response, and executor-relative oracle response, respectively, all reported as normalized differences. The final row is the arithmetic mean across fits.}
\label{tab:independent-learned-interface}
\begin{tabular}{@{}lrrr@{}}
\toprule
Fit & Utility & Action & Oracle \\
\midrule
Fit 1 & 0.123 & 0.534 & $-0.508$ \\
Fit 2 & 0.125 & 0.552 & $-0.521$ \\
Fit 3 & 0.124 & 0.552 & $-0.519$ \\
Fit 4 & 0.124 & 0.547 & $-0.514$ \\
Fit 5 & 0.125 & 0.561 & $-0.516$ \\
\midrule
Mean & 0.124 & 0.549 & $-0.516$ \\
\bottomrule
\end{tabular}
\vspace{-2mm}
\end{table}

\subsection{Additive commutative nuisance control}

The associative writer couples a producer feature to a selected utility. To test whether a value-permutation response could arise from separate marginal sensitivities, we fit an additive nuisance writer. For each supplied family--context block, all sums below run over its selected slots; its state is
\begin{equation}
h_{\mathrm{sep}}=\left(\sum\nolimits_t p_{\ell_t},\;\sum\nolimits_t(u_t-0.5)\right)\in\mathbb R^9,
\quad
s_a=\operatorname{sigmoid}((q_a^{\mathrm{sep}})^{\mathsf T}h_{\mathrm{sep}}),\quad
q_a^{\mathrm{sep}}\in\mathbb R^9.
\label{eq:additive-nuisance-state}
\end{equation}
The label features $p_{\ell}\in\mathbb R^8$ and future-action queries $q_a^{\mathrm{sep}}\in\mathbb R^9$ are learned from selected outcomes only on the same disjoint fitting streams, with the regularized objective used by the selected-only associative writer. Labels and utilities can each affect the future readout; their terminal summaries contain no cross-term. A within-block value permutation therefore preserves $h_{\mathrm{sep}}$, and moving complete pairs or labels across slots also preserves it. The predicted structural interaction is exactly zero for all three cells, independently of the learned marginal parameters.

The writer is fit once across the 192 fitting blocks and frozen before the 48-stream evaluation. The reported cluster summaries condition on this frozen fit and quantify evaluation-replicate variation. Under the additive state law, the value-null, time-placebo, and pair-placebo contrasts for future utility, future action, and executor-relative oracle agreement are exactly zero for every replicate and derangement. This zero response follows from the absence of a label--utility cross-term in the state.

\subsection{Policy-disjoint selected-outcome control}

The selected-only control can be tested against selection-history dependence by fitting the same eight-dimensional commutative writer on a disjoint training trace generated by \memory{}, then freezing its parameters before evaluation on the OCRR trace. The training streams use 48 distinct replicate indices, contain 9,216 selected adaptation outcomes, and supply no counterfactual future utility. The evaluation retains the OCRR future, outer maps, and blockwise null law used above. This changes the adaptation policy that supplies the training sequence while preserving the information boundary and the state operator.

The policy-disjoint writer gives a value-null utility interaction of $+0.022$ with interval $[+0.013,+0.031]$ and sign-flip $p<0.0001$. Its action interaction is $-1.4$ percentage points with interval $[-2.9,+0.1]$; the interval contains zero. The same direction is recovered against the time placebo, while the pair-preserving cell has the exact aligned state and action for every map. The utility interaction remains positive after this training-policy change, whereas the action readout remains inconclusive; interpretation remains conditional on the executor-relative synthetic state interface. The fit is global over the 192 training blocks and frozen before the 48-stream evaluation, so the reported uncertainty varies the evaluation replicate clusters conditional on that fit.

\begin{table}[ht]
\vspace{-4mm}
\centering
\small
\caption{Policy-disjoint order-invariant selected-outcome-only commutative control. The writer is fitted on selected outcomes generated by \memory{} and evaluated on the fixed OCRR trace; no counterfactual future utility enters the fit. Each row is one normalized interaction estimate with separate 95\% replicate-cluster Lower and Upper endpoints, and $p$ is the two-sided cluster sign-flip value.}
\label{tab:policy-disjoint-associative-interactions}
\begin{tabular}{@{}llrrrr@{}}
\toprule
Contrast & Metric & Estimate & Lower & Upper & $p$ \\
\midrule
Policy-disjoint $-$ value null & Utility & 0.022 & 0.013 & 0.031 & $<0.001$ \\
Policy-disjoint $-$ value null & Action & $-0.014$ & $-0.029$ & 0.001 & 0.082 \\
Policy-disjoint $-$ value null & Oracle & $-0.064$ & $-0.085$ & $-0.044$ & $<0.001$ \\
Policy-disjoint $-$ time placebo & Utility & 0.018 & 0.011 & 0.026 & $<0.001$ \\
Policy-disjoint $-$ time placebo & Action & $-0.023$ & $-0.037$ & $-0.009$ & 0.002 \\
Policy-disjoint $-$ time placebo & Oracle & $-0.065$ & $-0.082$ & $-0.048$ & $<0.001$ \\
Pair placebo $-$ value null & Utility & 0.022 & 0.013 & 0.031 & $<0.001$ \\
Pair placebo $-$ value null & Action & $-0.014$ & $-0.029$ & 0.001 & 0.082 \\
Pair placebo $-$ value null & Oracle & $-0.064$ & $-0.085$ & $-0.044$ & $<0.001$ \\
\bottomrule
\end{tabular}
\vspace{-2mm}
\end{table}

\subsection{Reverse cross-policy fit}

The policy-disjoint control changes the policy that generated the fitting trace. We reverse that direction as a second cross-policy check, reusing the same two replicate pools with their training and evaluation roles exchanged: the same selected-outcome-only commutative writer is fit on OCRR traces and evaluated on disjoint \memory{} traces. The future utility table is never used in the fit, and the evaluation writer is frozen before the \memory{} future is read. This direction changes both the training policy and the evaluation policy while retaining the state operator, fixed-trace intervention, nine outer maps, and blockwise null law. The two directions are reciprocal cross-fitting checks rather than independent replications.

The reverse fit gives the utility, action, and oracle interactions shown in Table~\ref{tab:policy-reverse-associative-interactions}. Its positive utility interaction has the same sign as the \memory{}-to-OCRR fit; the action and oracle readouts are operator consequences within the synthetic executor. The two directions form reciprocal cross-fitting checks conditional on the shared substrate and commutative writer.

\begin{table}[ht]
\vspace{-4mm}
\centering
\small
\caption{Reverse cross-policy order-invariant selected-outcome-only commutative control. The writer is fitted on selected outcomes generated by OCRR and evaluated on the fixed \memory{} trace; no counterfactual future utility enters the fit. Each row is one normalized interaction estimate with separate 95\% replicate-cluster Lower and Upper endpoints, and $p$ is the two-sided cluster sign-flip value.}
\label{tab:policy-reverse-associative-interactions}
\begin{tabular}{@{}llrrrr@{}}
\toprule
Contrast & Metric & Estimate & Lower & Upper & $p$ \\
\midrule
Reverse $-$ value null & Utility & 0.063 & 0.053 & 0.073 & $<0.001$ \\
Reverse $-$ value null & Action & 0.023 & 0.012 & 0.034 & $<0.001$ \\
Reverse $-$ value null & Oracle & $-0.214$ & $-0.231$ & $-0.196$ & $<0.001$ \\
Reverse $-$ time placebo & Utility & 0.054 & 0.044 & 0.064 & $<0.001$ \\
Reverse $-$ time placebo & Action & 0.030 & 0.019 & 0.044 & $<0.001$ \\
Reverse $-$ time placebo & Oracle & $-0.200$ & $-0.217$ & $-0.182$ & $<0.001$ \\
Pair placebo $-$ value null & Utility & 0.063 & 0.053 & 0.073 & $<0.001$ \\
Pair placebo $-$ value null & Action & 0.023 & 0.012 & 0.034 & $<0.001$ \\
Pair placebo $-$ value null & Oracle & $-0.214$ & $-0.231$ & $-0.196$ & $<0.001$ \\
\bottomrule
\end{tabular}
\vspace{-2mm}
\end{table}

\subsection{Independent replicate structural check}
\label{app:independent-replicate-check}

We fit the selected-outcome-only commutative writer on one set of 48 replicate streams and evaluate it on a disjoint set of 48 streams. The fitting target uses selected adaptation outcomes only; the evaluation keeps the same state-boundary intervention, nine outer maps, held-fixed future, and executor readout. The evaluation streams are independent of the streams used for the primary structural suite.

\begin{table}[ht]
\vspace{-4mm}
\centering
\small
\caption{Independent replicate structural check for the order-invariant selected-only writer. Utility, action, and oracle entries are normalized differences. Both rows use the same 48 evaluation replicate streams and nine paired derangements; the pair-preserving row equals the aligned row by the declared commutative update law, so the repeated numbers are the invariance result.}
\label{tab:independent-gate-interactions}
\begin{tabular}{@{}lrrr@{}}
\toprule
State cell & Utility & Action & Oracle \\
\midrule
Selected-only fit & 0.025 & 0.013 & $-0.094$ \\
Pair-preserving cell & 0.025 & 0.013 & $-0.094$ \\
\bottomrule
\end{tabular}
\vspace{-2mm}
\end{table}

The selected-outcome-only writer gives a utility interaction of $+0.025$ with 95\% replicate-cluster interval $[+0.015,+0.035]$ and sign-flip $p<0.0001$. Its action interaction is $+1.3$ percentage points with interval $[+0.4,+2.2]$ and $p=0.0091$; the oracle interaction is $-9.4$ percentage points with interval $[-11.9,-7.0]$ and $p<0.0001$. The pair-preserving cell reproduces the aligned state on every future action under every outer map, giving $0$ mismatches in the paired-state check. This result is conditional on the declared synthetic executor and the disjoint evaluation set.

\subsection{Training-label permutation control}
\label{app:training-label-null}

The selected-only control tests sensitivity to the training label--utility pairing while leaving the held-out evaluation protocol unchanged. Within each of the 192 training blocks, we independently permute the selected labels, preserve the selected-utility sequence and block label multiset, fit the same leave-one-out commutative objective, and freeze the writer before reading the canonical held-out labels. The five fixed control fits reverse the utility and oracle signs of the canonical result, while the pair-preserving state remains exactly invariant (Table~\ref{tab:training-label-null}); the action interaction varies around zero. This permutation changes the alignment between fitting labels and canonical held-out labels. The separate global-renaming null also changes the stored-key and candidate coordinates and therefore tests equivariance under a different intervention.

\begin{table}[ht]
\vspace{-4mm}
\centering
\small
\caption{Training-label permutation calibration for the order-invariant selected-outcome-only writer. The canonical row uses the producer labels in the fitting trace. Rows S1--S5 independently permute those labels within each of 192 training blocks, preserve the selected-utility sequence and block label multiset, and evaluate the frozen fit on the canonical held-out labels. $U$, $A$, and $O$ are normalized utility, action, and executor-relative oracle interactions, respectively; mismatch is the pair-preserving action mismatch count. The five shuffled rows are fit-variation summaries and do not add inference units. The mismatch count holds for each of nine nonidentity maps separately, with $48\times6\times48=13{,}824$ comparisons per map; the identity map is excluded.}
\label{tab:training-label-null}
\begin{tabular}{@{}lrrrr@{}}
\toprule
Fit & Utility & Action & Oracle & Pair mismatch count \\
\midrule
Canonical & 0.025 & 0.013 & $-0.094$ & 0 \\
S1 & $-0.106$ & 0.002 & 0.107 & 0 \\
S2 & $-0.029$ & $-0.042$ & 0.026 & 0 \\
S3 & $-0.044$ & 0.022 & 0.062 & 0 \\
S4 & $-0.042$ & $-0.010$ & 0.054 & 0 \\
S5 & $-0.115$ & $-0.047$ & 0.103 & 0 \\
\bottomrule
\end{tabular}
\vspace{-2mm}
\end{table}

The aligned and deranged label-invariant states are identical on all 13,824 future tasks, with zero action change and zero utility change. The component variants show that the effect is distributed across the carried global prior and local arm-indexed ledger; neither component alone provides an operator-independent interpretation. These estimates are descriptive structural controls and do not add independent replicates to the primary test.

\section{Inference and condition diagnostics}
\label{app:inference-diagnostics}

\subsection{Statistical definitions}
\label{app:statistical-definitions}
\label{app:statistical-analysis}

For each condition and split, the analysis gives mean selected-task utility, oracle regret, effective cost, utility per cost, calibration mean absolute error (MAE), failure rate, and oracle gap. Each comparison is computed on tasks shared by the treatment and baseline conditions, and the point estimand is the mean of the 48 replicate-level differences. Bootstrap resampling samples complete replicate blocks; the primary percentile interval uses 20,000 draws. The confirmatory sign-flip test flips each replicate-level mean difference with probability one half under the stated cluster-level sign-symmetry null, using 20,000 Monte Carlo draws and the $(k+1)/(B+1)$ correction. The resulting $p$-value tests sign symmetry of the replicate-level contrast, while the point estimate remains their mean; this does not posit a sign-symmetric law for an arbitrary superpopulation mean. Fixed-label maps are the declared coupled interventions, so their structural role remains separate from the replicate-level~sampling~distribution.

The original controller study designated OCRR versus \noupdate{} future utility as its confirmatory efficacy test. Here these controller comparisons provide supporting context for the assignment audit. Secondary efficacy comparisons, including online controls, disjoint-stream reconstructions, changed-substrate transfers, and the secondary condition matrix, are contextual contrasts. Representation and operator-interaction tables, including Table~\ref{tab:selected-only-associative-interactions}, are diagnostics of the state-boundary estimand; the value-permutation table is its matched-count mechanism control. All reported $p$-values are two-sided replicate-cluster sign-flip summaries for their stated contrast; none is treated as a familywise decision test. Arms, prompts, and controller calls are not independent~scientific~samples.

\subsection{Primary and secondary inference}
\label{app:primary-inference}

The primary OCRR--\noupdate{} future-utility comparison has difference +0.008, 95\% replicate-cluster interval [-0.001, +0.016], and sign-flip $p=0.0795$. The interval includes zero, so the pooled comparison is inconclusive at the $\alpha=0.05$ threshold. OCRR's secondary efficiency comparison with \extrasampling{} is positive on utility per effective cost ($+0.036$). Against the primary online \shuffle{} control, the difference is +0.143 with interval [+0.116, +0.171]; this is a composite misalignment contrast. In the secondary full matrix, OCRR versus \staticselected{} is +0.006 with interval [-0.006, +0.017], OCRR versus \contextucb{} is +0.136 with interval [+0.122, +0.150], and OCRR versus \replay{} is +0.029 with interval [+0.021, +0.037]. The stratified table retains the opposite temporal-regime signs and the zero-utility primary rows. Across all nine fixed derangements, pooled OCRR--replay estimates range from $+0.024$ to $+0.330$; the sign is stratum- and regime-dependent.

A secondary structural interaction analysis averages the OCRR--\noupdate{} effect within each replicate before contrasting temporal low-noise and high-noise contexts. The paired low-minus-high estimate is +0.161 with interval [+0.114, +0.208] and cluster sign-flip $p<0.0001$. It is a secondary analysis of the noise-dependent transfer pattern.

\noindent\textbf{Disjoint-stream reconstruction.}

The disjoint-stream reconstruction uses 48 replicate streams that do not overlap with the primary streams, under the declared selection--outcome--update--held-fixed-future order. Across its seven conditions, the design contains 29,952 task specifications: each stream has 144 development, 192 adaptation, and 288 future tasks. It contains 161,280 selected non-development condition records, namely 23,040 per condition; each condition contributes 13,824 future-task decisions. There are no future-state updates. The pooled OCRR--\noupdate{} utility difference is $+0.015$ with a 95\% replicate-cluster interval $[+0.008,+0.021]$ and sign-flip $p<0.0001$. Source-localization and survival-risk contrasts remain zero; the temporal low- and high-reliability strata are $+0.120$ and $-0.030$. The disjoint-stream reconstruction recovers this conditional pattern under the same state-writing rules.

The current fixed-trace control suite preserves the same future keys and placebo information boundaries. Its pooled OCRR--time-placebo interactions are $+0.323$ in future utility, $-4.9\,\text{pp}$ in action-change fraction, and $-42.7\,\text{pp}$ in oracle-match change. The corresponding aligned-minus-time utility interactions for UCB1, local-only, and global-only are $+0.129$, $+0.241$, and $+0.298$. These values provide secondary structural evidence for the operator-conditional reading.

\subsection{Additional condition diagnostics}
\label{app:condition-diagnostics}

The primary contrast is interpreted with the complete condition matrix. Table~\ref{tab:condition-summary} reports condition-level means for adaptation and future splits, effective cost, utility per cost, calibration, and failure rate; its oracle gap is the utility regret reported in the outcome block.

\begin{table*}[ht]
\vspace{-4mm}
\centering
\small
\caption{Condition-level outcomes and secondary diagnostics from the primary matrix. The upper panel reports means over selected tasks for adaptation and future splits. Tasks ($n$) is the number of selected or future tasks; Utility, Oracle gap, Cost, and Utility per cost are reported on their stated scales, with Cost denoting effective cost and Utility per cost dividing Utility by Cost. The lower panel reports calibration mean absolute error (MAE) and Failure rate. Oracle gap, MAE, and Failure rate are lower-is-better diagnostics; Utility and Utility per cost are future-outcome summaries.}
\label{tab:condition-summary}
\begin{tabular*}{\textwidth}{@{\extracolsep{\fill}}llrrrrr@{}}
\toprule
\multicolumn{7}{l}{\textit{Condition-level outcomes}}\\
Condition & Split & Tasks ($n$) & Utility & Oracle gap & Cost & Utility per cost \\
\midrule
\noupdate{} & adaptation & 9216 & 0.544 & 0.222 & 10.000 & 0.054 \\
\noupdate{} & future & 13824 & 0.461 & 0.158 & 10.000 & 0.046 \\
\staticpred{} & adaptation & 9216 & 0.437 & 0.329 & 10.000 & 0.044 \\
\staticpred{} & future & 13824 & 0.467 & 0.153 & 10.000 & 0.047 \\
\memory{} & adaptation & 9216 & 0.555 & 0.211 & 10.000 & 0.056 \\
\memory{} & future & 13824 & 0.421 & 0.199 & 10.000 & 0.042 \\
\skill{} & adaptation & 9216 & 0.613 & 0.153 & 10.000 & 0.061 \\
\skill{} & future & 13824 & 0.365 & 0.255 & 10.000 & 0.036 \\
\extrasampling{} & adaptation & 9216 & 0.347 & 0.419 & 30.000 & 0.012 \\
\extrasampling{} & future & 13824 & 0.333 & 0.287 & 30.000 & 0.011 \\
\shuffle{} & adaptation & 9216 & 0.363 & 0.403 & 10.000 & 0.036 \\
\shuffle{} & future & 13824 & 0.326 & 0.294 & 10.000 & 0.033 \\
OCRR & adaptation & 9216 & 0.627 & 0.139 & 10.000 & 0.063 \\
OCRR & future & 13824 & 0.469 & 0.151 & 10.000 & 0.047 \\
\bottomrule
\end{tabular*}
\par\medskip
\begin{tabular*}{\textwidth}{@{\extracolsep{\fill}}llrr@{}}
\toprule
\multicolumn{4}{l}{\textit{Secondary condition diagnostics}}\\
Condition & Split & Calibration MAE & Failure rate \\
\midrule
\noupdate{} & adaptation & 0.325 & 0.000 \\
\noupdate{} & future & 0.222 & 0.000 \\
\staticpred{} & adaptation & 0.230 & 0.000 \\
\staticpred{} & future & 0.171 & 0.000 \\
\memory{} & adaptation & 0.277 & 0.000 \\
\memory{} & future & 0.293 & 0.000 \\
\skill{} & adaptation & 0.206 & 0.000 \\
\skill{} & future & 0.303 & 0.000 \\
\extrasampling{} & adaptation & 0.348 & 0.000 \\
\extrasampling{} & future & 0.282 & 0.000 \\
\shuffle{} & adaptation & 0.313 & 0.000 \\
\shuffle{} & future & 0.269 & 0.000 \\
OCRR & adaptation & 0.224 & 0.000 \\
OCRR & future & 0.294 & 0.000 \\
\bottomrule
\end{tabular*}
\vspace{-2mm}
\end{table*}

Calibration mean absolute error (MAE) and failure rate are diagnostic fields. The oracle utility gap is reported once in the outcome block under the unified name \emph{Oracle gap}. A lower oracle gap describes alignment with the synthetic executor's counterfactual criterion; it does not establish semantic correctness of a strategy label.

The secondary full matrix adds selected-arm static calibration, a contextual UCB controller, and fixed-trace permutation replay. Their paired contrasts use the same replicate-cluster procedure as the primary matrix and distinguish state representation, selected-outcome calibration, and~replay~response.

\begin{table*}[ht]
\vspace{-4mm}
\centering
\small
\caption{Secondary full-matrix future-utility contrasts. Each estimate is the left condition minus the right condition on the normalized future-utility scale. The 95\% interval is split into Lower and Upper endpoints; task and replicate counts are reported separately. The row shorthands \emph{Selected}, \emph{UCB1}, \emph{Replay}, and \emph{Static} denote \staticselected{}, \contextucb{}, \replay{}, and \staticpred{}, respectively. These are secondary condition and mechanism contrasts.}
\label{tab:secondary-comparison}
\begin{tabular*}{\textwidth}{@{\extracolsep{\fill}}lrrrrrr@{}}
\toprule
Secondary comparison & Difference & Lower & Upper & Sign-flip $p$ & Tasks & Reps. \\
\midrule
OCRR--\emph{Selected} & 0.006 & $-0.006$ & 0.017 & 0.324 & 13824 & 48 \\
OCRR--\emph{UCB1} & 0.136 & 0.122 & 0.150 & $<0.001$ & 13824 & 48 \\
OCRR--\emph{Replay} & 0.029 & 0.021 & 0.037 & $<0.001$ & 13824 & 48 \\
\emph{Selected}--\emph{Static} & $-0.004$ & $-0.011$ & 0.003 & 0.245 & 13824 & 48 \\
\bottomrule
\end{tabular*}
\vspace{-2mm}
\end{table*}

\noindent\textbf{Overview of fitted and cross-interface writers.} The complete fitted-state and changed-substrate summary, including fitting and evaluation-stream counts, is given in Table~\ref{tab:fitted-transport-summary}. The following sections give the task-specific constructions and interval estimates.
\section{Changed-substrate reconstruction}
\label{app:changed-substrate}

An independent matrix tests dependence on the primary executor with a fresh task generator, executor, probe allocator, and policy ledger. The two decision families are hypothesis screening and model validation. Each family has low- and high-noise contexts, 24 development, 48 adaptation, and 48 future tasks per context in each of 48 replicate streams. The controller sees only family, context, and its pre-decision state; the selected outcome is written after the decision, while the future task set and evaluator remain fixed after adaptation. The outcome functions are separate from those of the primary substrate, so the reconstruction changes the executor as well as the task names.

The changed-substrate task construction uses $r$ for replicate streams, $j$ for the global task index, and $\ell$ for the within-split task index. A deterministic indexed draw fixes each world and its arm-specific observations before the policy acts. Each family--context cell contains 24 development, 48 adaptation, and 48 future tasks. In hypothesis screening, twelve candidates have latent values $z_i\sim\mathcal N(0,\tau_\phi^2)$, with phase scales $\tau_\phi\in\{0.90,1.10,0.78\}$ for development, adaptation, and future; future values additionally receive a linear offset from $-0.18$ to $0.18$. The $k$th probe observes $b_{ik}=z_i+\epsilon_{ik}$, where $\epsilon_{ik}\sim\mathcal N(0,\eta_c^2)$ and $\eta_c\in\{0.16,0.62\}$ for the two contexts. Twelve probes are allocated by the arm-specific ledger, and the selected three candidates receive utility
\begin{equation}
 U_{\mathrm{HS}}=\operatorname{clip}_{[0,1]}\!\left(
 0.50+0.22\,\frac{\bar z_{\mathrm{sel}}-\bar z}{\operatorname{sd}(z)+10^{-6}}\right).
 \label{eq:changed-hs-utility}
\end{equation}
The oracle replaces $\bar z_{\mathrm{sel}}$ by the mean of the three largest latent values. Arm-specific probe draws are fixed by the same indexed law, with separate streams for hypothesis screening and model validation.

In model validation, ten latent response probabilities are generated as $p_i=\operatorname{sigmoid}(\ell_i)$, where $\ell_i\sim\mathcal N(0,0.85^2)+\operatorname{linspace}(-0.75,0.75,10)+\kappa_\phi$ and $\kappa_\phi\in\{-0.20,0.15,0.38\}$ by phase. A probe observes the clipped response $b_i=\operatorname{clip}_{[0,1]}(p_i+\epsilon_i)$ with standard deviation $0.035$ or $0.18$ according to context. The controller's utility is
\begin{equation}
U_{\mathrm{MV}}=\exp\!\left(-\frac{1}{0.045\!\times\!10}
 \sum\nolimits_{i=1}^{10}(\hat p_i-p_i)^2\right),
 \label{eq:changed-mv-utility}
\end{equation}
with unprobed estimates fixed at $0.5$; the executor-only oracle has utility one. These equations define a new outcome-generating family, while the adaptation--future information order and the arm-indexed state interface are held comparable to the primary substrate.

The changed-substrate probe allocator and arm ledger complete the decision rule. For either family, \textsc{coverage} probes candidate $i_t=t\bmod n$, \textsc{replicate} repeats the preceding candidate (and starts with $i_0=0$), and \textsc{mixed} uses coverage for the initial $\lfloor n/2\rfloor$ probes before using the adaptive rule. The adaptive rule uses the lowest-index unvisited candidate in the absence of history; otherwise it scores a visited candidate $i$ by
\begin{equation}
 s_i^{\mathrm{probe}}=\bar b_i+\frac{0.30}{\sqrt{n_i}},
 \label{eq:changed-adaptive-score}
\end{equation}
with empirical mean $\bar b_i$ and count $n_i$. Unvisited candidates receive $0.50+0.08U_i$ with a fixed uniform draw, and the largest score is selected with the deterministic lowest-index tie rule. The probe budget is twelve in both families.

The independent ledger stores counts and utility sums by family, context, and arm. \noupdate{} selects \textsc{mixed}; \staticpred{} selects the largest development mean; \memory{} repeats the last arm after a value at least $0.56$ and otherwise follows round-robin; \skill{} selects the largest family-level mean; and \extrasampling{} samples an arm uniformly. \shuffle{} replaces the selected utility by a uniformly sampled earlier utility within the same family--context key before updating its ledger. \textsc{OCRR} uses a separate ledger score
\begin{equation}
\begin{aligned}
 s_i^{\mathrm{ledger}}
   &=\bar u_i+\frac{0.30}{\sqrt{1+n_i}}+0.08\max(0,L_i),\\
 L_i
   &=\begin{cases}
       \bar u_i-1.55\sqrt{\dfrac{\max(0.02,\bar u_i(1-\bar u_i))}{n_i}},
          & n_i\ge 4,\\
       0, & n_i<4.
      \end{cases}
\end{aligned}
\label{eq:changed-ledger-score}
\end{equation}
over arm means, and quarantines an arm for six steps after at least six observations when the current value is below $0.35$ and its running mean is below $0.47$. Every condition receives only the selected adaptation outcome after its decision; the future state is frozen before the future split. All ledger draw and tie conventions are fixed before evaluation; probe ties use the lowest candidate index, whereas ledger ties select the lexicographically largest arm in the fixed arm order.

The pooled OCRR--\noupdate{} difference is +0.175 with 95\% replicate-cluster interval $[+0.171,+0.180]$ and sign-flip $p<0.0001$. The family--context estimates are $+0.094$ and $+0.069$ for hypothesis screening at low and high noise, and $+0.436$ and $+0.103$ for model validation. This secondary result is conditional on the controlled substrate and does not address external language-agent performance.

The same changed-substrate comparisons also provide a fixed-label robustness check. The nine non-identity derangements produce pooled OCRR--replay contrasts from $+0.171$ to $+0.268$, with all four family--context strata positive for every map. This remains conditional on the changed-substrate arm-indexed ledger and the fixed adaptation trace; the derangements reuse the same replicate streams. Table~\ref{tab:cross-substrate-replay} gives the complete map-level estimates.

\begin{table}[ht]
\vspace{-4mm}
\centering
\small
\caption{Changed-substrate fixed-label replay. The displayed values are point estimates for the changed substrate; map strings are permutations of source labels C, A, R, and M, listed in that order. OCRR--\replay{} is the normalized pooled future-utility difference; positive strata is an integer count of the four family--context strata with a positive estimate. The maps reuse the same replicate streams and are fixed-label robustness variants, not independent samples.}
\label{tab:cross-substrate-replay}
\begin{tabular}{rlrr}
\toprule
\textbf{ID} & \textbf{Map} & \textbf{OCRR--\replay{}} & \textbf{Positive strata} \\
\midrule
1 & ACMR & $0.223$ & 4 \\
2 & ARMC & $0.218$ & 4 \\
3 & AMCR & $0.223$ & 4 \\
4 & RCMA & $0.268$ & 4 \\
5 & RMCA & $0.268$ & 4 \\
6 & RMAC & $0.263$ & 4 \\
7 & MCAR & $0.177$ & 4 \\
8 & MRCA & $0.176$ & 4 \\
9 & MRAC & $0.171$ & 4 \\
\bottomrule
\end{tabular}
\vspace{-2mm}
\end{table}

The four nonpooled strata of this selected-only commutative calibration are reported in Table~\ref{tab:cross-substrate-selected-only}. They use the same held-out streams throughout; the strata are descriptive decompositions of the pooled result, not additional inferential units.

\noindent\textbf{Selected-only control on the independent substrate.} We fit the same order-invariant selected-outcome-only writer on 24 streams of the independent substrate and froze it before evaluating the other 24 streams. The fit uses selected adaptation outcomes only; it receives no future utility, and the four state-boundary cells use the same nine outer maps, held-fixed future, and executor within this substrate. Inference is over the 24 held-out replicate streams, with maps treated as paired interventions within each stream. The pooled utility, action, and oracle interactions are $+0.012$ with interval $[+0.005,+0.020]$ and $p=0.0040$, $+5.4$ percentage points with interval $[+2.2,+8.4]$ and $p=0.0027$, and $-4.9$ percentage points with interval $[-7.7,-2.1]$ and $p=0.0028$, respectively. All four utility interactions have positive point estimates, although the stratum-level statement is descriptive: the hypothesis--high interval is $[0.000,+0.002]$ with $p=0.498$, and the model--validation--high interval is $[-0.010,+0.025]$ with $p=0.334$. The high-noise strata therefore do not independently establish a positive interaction. The pair-preserving cell has zero action mismatches across all maps. Table~\ref{tab:cross-substrate-selected-only} reports the point estimates for this secondary held-out control on a distinct synthetic substrate; replicate streams remain the inferential units.

\begin{table}[ht]
\vspace{-4mm}
\centering
\small
\caption{Changed-substrate transport by task family and noise level. Entries are aligned-minus-value interactions from the fitted selected-outcome writer. Utility is normalized; action and oracle interactions are differences of proportions. The Pair changes column counts action mismatches under complete-pair transport across all nine maps. Each stratum has 24 evaluation streams and 48 future tasks per stream, giving 10,368 compared actions.}
\label{tab:content-substrate-strata}
\label{tab:cross-substrate-selected-only}
\begin{tabular*}{\textwidth}{@{\extracolsep{\fill}}lrrrrrrrr@{}}
\toprule
 & \multicolumn{4}{c}{\textbf{Low}} & \multicolumn{4}{c}{\textbf{High}} \\
\cmidrule(lr){2-5}\cmidrule(lr){6-9}
\textbf{Task family} & $\widehat{\Gamma}_U$ & $\widehat{\Gamma}_A$ & $\widehat{\Gamma}_O$ & \textbf{Pair changes} & $\widehat{\Gamma}_U$ & $\widehat{\Gamma}_A$ & $\widehat{\Gamma}_O$ & \textbf{Pair changes} \\
\midrule
Hypothesis & 0.009 & 0.028 & $-0.087$ & 0 & 0.001 & 0.009 & $-0.005$ & 0 \\
Validation & 0.030 & 0.069 & $-0.069$ & 0 & 0.010 & 0.111 & $-0.036$ & 0 \\
\bottomrule
\end{tabular*}
\vspace{-2mm}
\end{table}

\begin{figure}[ht]
\centering
\includegraphics[width=\linewidth]{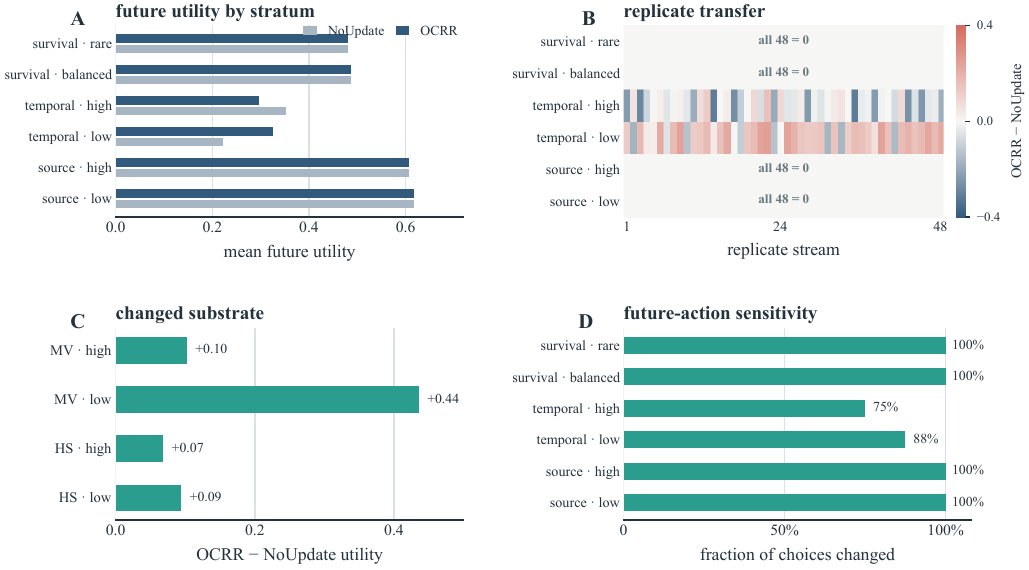}
\vspace{-6mm}
\caption{\textbf{Transfer and action sensitivity across controlled strata.} Panel A compares the primary OCRR and no-update future-utility levels; panel B keeps the replicate stream visible for the same transfer contrast; panel C shows the independent changed-substrate utility checks; and panel D reports the fraction of future choices changed by replay. Replicate streams remain the sampling units throughout.}
\label{fig:transfer-summary}
\vspace{-4mm}
\end{figure}

The changed substrate also receives the time-preserving placebo under the same nine maps. The placebo uses the original OCRR adaptation outcomes and slots, preserves every block label multiset, and changes 2,802 of 9,216 stored labels across 192 blocks; no block remains identical, and the selected-outcome and future-task conditions are unchanged. The future task set and evaluator remain fixed after adaptation, so the placebo interaction is a within-trace operator-family check. The direct aligned--placebo contrast is +0.246 in future utility with 97.4\% of future actions changed. Averaged over the nine maps, the utility interaction is +0.327 and the oracle-match interaction is -0.980 across 48 replicate streams. Figure~\ref{fig:transfer-summary} collects the stratum-level utility, transfer, changed-substrate, and action-sensitivity views.

\section{Public-data endpoint check}
\label{app:public-data-replication}

The public-data endpoint check evaluates the controller with measured scientific observations. It uses the red wine quality and Wisconsin diagnostic breast cancer data sets from the UCI Machine Learning Repository~\citep{cortez2009ucidata,wolberg1993ucidata}. For each task, the four arms define sampling designs over a candidate pool. The evaluator is a fixed ridge-regularized logistic model fitted only on the selected design and scored by held-out log loss on an evaluation pool. Utility is the exponential transform of this loss. The evaluator is independent of the controller and exposes no~unselected~outcome.

Each data family has an iid context and a shifted context. The iid context draws a candidate pool of 96 observations and a disjoint evaluation pool of 96 observations without replacement; the shifted context draws the two pools from opposite sides of the median of the first standardized feature. The wine label is one for quality at least six, and the breast-cancer label is one for a malignant diagnosis. Feature standardization uses the mean and standard deviation of the complete unlabeled data family before task construction. This label-free transductive normalization is fixed before the task split; no outcome label enters it. Every context has 24 development, 24 adaptation, and 24 future tasks in each of 48 replicate streams. A task presents only its family and context to the controller. After the sampling arm is selected, the selected design's held-out utility is revealed and may update the state. All arm counterfactuals, future pools, and evaluator labels remain outside the controller.

The evaluator uses 80 ridge-gradient steps with step size $0.12$, penalty $0.08$ on non-intercept coefficients, and an unregularized intercept. Coverage selects observations in increasing order of the first standardized feature; adaptive uses a four-example coverage warm start and then selects the unvisited observation closest to the boundary between the current positive and negative centroids; replicate selects the nearest unvisited observation to the latest selected point; mixed uses coverage for the first half and adaptive selection for the second half. Exact-index ties use the original pool order. Each arm selects 24 observations. The effective-cost multiplier is one for every condition except \extrasampling{}, for which it is three; utility per cost divides the held-out utility by this effective cost. Together, these rules define an executor whose outcomes are fixed by measured labels rather than a latent task generator.

The secondary analysis uses replicate stream as the inference cluster and task as the decision unit. Its external-executor endpoint is OCRR--\noupdate{} future utility; OCRR--\shuffle{} tests dependence on outcome identity, and OCRR--\extrasampling{} compares utility per effective cost. These are descriptive secondary comparisons. The four stratum results and condition-level summaries are reported in Table~\ref{tab:public-data-strata}, with cluster uncertainty in the appendix tables. The endpoint and structural branches use the same measured substrate for different targets: measured-outcome transfer and typed state-writer responses.

\begin{table}[ht]
\vspace{-4mm}
\centering
\small
\caption{Public-data endpoint-check stratum estimates. OCRR and \noupdate{} are normalized utility ratios; $\Delta U_{\mathrm{O-N}}$ is the O-minus-\noupdate{} utility difference in $10^{-3}$ normalized utility units. The estimate, lower, and upper columns give the estimate and 95\% cluster-bootstrap interval over the 48 replicate streams; each stratum contains 24 future tasks per stream.}
\label{tab:public-data-strata}
\begin{tabular}{llrrrrr}
\toprule
\textbf{Data} & \textbf{Context} & \textbf{OCRR} & \textbf{\noupdate{}} & \multicolumn{3}{c}{$\boldsymbol{\Delta U_{\mathrm{O-N}}}$ ($10^{-3}$)} \\
\cmidrule(lr){5-7}
&  &  &  & \textbf{Estimate} & \textbf{Lower} & \textbf{Upper} \\
\midrule
Breast cancer & IID & $0.834$ & $0.834$ & $-0.240$ & $-0.660$ & $0.084$ \\
Breast cancer & Shift & $0.755$ & $0.755$ & $-0.007$ & $-0.021$ & $0.000$ \\
Wine quality & IID & $0.541$ & $0.544$ & $-2.780$ & $-5.243$ & $-0.487$ \\
Wine quality & Shift & $0.552$ & $0.557$ & $-4.705$ & $-7.793$ & $-1.927$ \\
\bottomrule
\end{tabular}
\vspace{-2mm}
\end{table}

\begin{table}[ht]
\vspace{-4mm}
\centering
\small
\caption{Pooled public-data executor contrasts. Metrics are future utility $U$ and utility per effective cost $U/c$. Contrasts use $10^{-3}$ units for the stated metric. Estimate is the mean paired contrast; Lower and Upper delimit its 95\% cluster-bootstrap interval over 48 replicate streams and 4,608 future tasks~per~condition.}
\label{tab:public-data-pooled}
\begin{tabular*}{\textwidth}{@{\extracolsep{\fill}}llrrrr@{}}
\toprule
\textbf{Comparison} & \textbf{Metric} & \textbf{Estimate} & \textbf{Lower} & \textbf{Upper} & $n_{\rm streams}$ \\
\midrule
OCRR $-$ \noupdate{} & $U$ & $-1.933$ & $-2.774$ & $-1.165$ & 48 \\
OCRR $-$ \shuffle{} & $U$ & $5.666$ & $3.416$ & $8.063$ & 48 \\
OCRR $-$ \extrasampling{} & $U/c$ & $18.702$ & $18.660$ & $18.743$ & 48 \\
\bottomrule
\end{tabular*}
\vspace{-2mm}
\end{table}

\begin{table}[ht]
\vspace{-4mm}
\centering
\small
\caption{Public-data executor condition means. Utility, regret, and utility/cost are normalized continuous quantities; utility/cost divides the held-out utility by the realized effective sampling cost.}
\label{tab:public-data-conditions}
\begin{tabular}{lrrr}
\toprule
\textbf{Condition} & \textbf{Utility} & \textbf{Regret} & \textbf{Utility/cost} \\
\midrule
\extrasampling{} & $0.666$ & $0.024$ & $0.009$ \\
\memory{} & $0.667$ & $0.023$ & $0.028$ \\
\skill{} & $0.655$ & $0.035$ & $0.027$ \\
\noupdate{} & $0.673$ & $0.017$ & $0.028$ \\
OCRR & $0.671$ & $0.019$ & $0.028$ \\
\shuffle{} & $0.665$ & $0.025$ & $0.028$ \\
\staticpred{} & $0.673$ & $0.016$ & $0.028$ \\
\bottomrule
\end{tabular}
\vspace{-2mm}
\end{table}

\subsection{Cross-fitted persistent agent on measured outcomes}
\label{app:public-persistent-agent-replay}

We apply the same fixed-trace cells to a persistent scientific decision agent whose writer is learned from public measured outcomes. The agent receives the selected arm, its held-out utility, and the family--context key. It maintains a context-specific recurrent state and uses a frozen action readout after the adaptation boundary:
\begin{equation}
h_t=\tanh\!\left(W_hh_{t-1}+W_u[\operatorname{onehot}(a_t);\operatorname{onehot}(f,c);y_t]+b_h\right),\quad
\widehat{u}_a=\operatorname{sigmoid}(w_a^\top h_T+b_a).
\label{eq:public-persistent-agent}
\end{equation}
The state dimension is 16. The adaptation policy uses an eight-task arm-balanced warm-up and then the frozen readout; the warm-up fixes a common initial exposure without consulting any future outcome. The recurrent parameters are fit by predicting the four-arm mean utility on the future tasks of each training stream. This target is formed only inside the training fold. Each fit uses 500 full-batch epochs, learning rate $0.012$, $\ell_2$ penalty $10^{-4}$, and gradient-norm clipping at 8; the three parameter seeds are 20260861--20260863 and no evaluation-fold criterion selects among them.

The two cross-fitting folds train on replicates 0--23 and evaluate on 24--47, then swap the roles. Each fit uses 24 adaptation tasks per context and three parameter initializations. On an evaluation fold, the fitted parameters are frozen before the agent receives any evaluation-fold outcome. The aligned adaptation action trace is then held fixed while replay changes only the typed state sequence. Rekey maps producer labels at their original slots; value reassigns utilities; pair transports complete label--utility pairs; and time reassigns labels. Future task keys, the public measured-outcome evaluator, and the terminal readout are common to all cells. No future task updates the state.

Table~\ref{tab:public-persistent-agent-folds} reports the two held-out folds. Replicate stream is the inferential unit; the three fits are averaged within a replicate before the cluster bootstrap. Value reassignment changes almost no actions, whereas time and pair transport change actions in both folds; utility and oracle effects differ by fold. The evidence concerns a fold-specific, order-sensitive action response on this~measured-outcome~executor.

\begin{table*}[ht]
\vspace{-4mm}
\centering
\small
\caption{Cross-fitted persistent-agent replay on public measured outcomes. $A$ is the changed-action proportion; $\Delta U_{\mathrm{C-A}}$ and $\Delta O_{\mathrm{C-A}}$ are cell-minus-aligned contrasts in $10^{-3}$ normalized utility units and proportion units, respectively. Estimate is the point estimate; Lower and Upper delimit its 95\% replicate-cluster bootstrap interval. Fold I trains on replicates 0--23 and evaluates 24--47; Fold II swaps these sets.}
\label{tab:public-persistent-agent-folds}
\begin{tabular*}{\textwidth}{@{\extracolsep{\fill}}llrrrrrr@{}}
\toprule
 &  & \multicolumn{3}{c}{\textbf{Fold I}} & \multicolumn{3}{c}{\textbf{Fold II}} \\
\cmidrule(lr){3-5}\cmidrule(lr){6-8}
\textbf{Cell} & \textbf{Metric} & \textbf{Estimate} & \textbf{Lower} & \textbf{Upper} & \textbf{Estimate} & \textbf{Lower} & \textbf{Upper} \\
\midrule
Rekey & $A$ & $0.753$ & $0.750$ & $0.757$ & $0.620$ & $0.607$ & $0.631$ \\
Rekey & $\Delta U_{\mathrm{C-A}}$ & $-0.602$ & $-1.194$ & $-0.023$ & $0.870$ & $-0.016$ & $1.714$ \\
Rekey & $\Delta O_{\mathrm{C-A}}$ & $-0.082$ & $-0.099$ & $-0.066$ & $0.027$ & $0.005$ & $0.048$ \\
Key-slot & $A$ & $0.243$ & $0.191$ & $0.295$ & $0.240$ & $0.188$ & $0.292$ \\
Key-slot & $\Delta U_{\mathrm{C-A}}$ & $-0.476$ & $-0.854$ & $-0.123$ & $0.285$ & $-0.303$ & $0.819$ \\
Key-slot & $\Delta O_{\mathrm{C-A}}$ & $-0.009$ & $-0.027$ & $0.007$ & $0.003$ & $-0.016$ & $0.021$ \\
Value & $A$ & $0.000$ & $0.000$ & $0.000$ & $0.003$ & $0.000$ & $0.010$ \\
Value & $\Delta U_{\mathrm{C-A}}$ & $0.000$ & $0.000$ & $0.000$ & $0.012$ & $0.000$ & $0.036$ \\
Value & $\Delta O_{\mathrm{C-A}}$ & $0.000$ & $0.000$ & $0.000$ & $0.000$ & $0.000$ & $0.000$ \\
Pair & $A$ & $0.306$ & $0.253$ & $0.358$ & $0.257$ & $0.215$ & $0.299$ \\
Pair & $\Delta U_{\mathrm{C-A}}$ & $-0.786$ & $-1.219$ & $-0.397$ & $0.105$ & $-0.321$ & $0.555$ \\
Pair & $\Delta O_{\mathrm{C-A}}$ & $-0.027$ & $-0.046$ & $-0.010$ & $0.008$ & $-0.007$ & $0.024$ \\
\bottomrule
\end{tabular*}
\vspace{-2mm}
\end{table*}

\subsection{Frozen language-model bridge on measured outcomes}
\label{app:llm-bridge}

\noindent\textbf{Archived time-labelled cells.}
\label{app:textual-time-audit}
The retained textual-replay implementation applies its time permutation to both keys and utilities, producing another complete-pair transport rather than the key-only time intervention in Table~\ref{tab:typed-cells}. Rebuilding the inputs for both available Sol/Arm runs reproduces all 144 stored request digests per run for this cell with the complete-pair implementation; the key-only alternative matches none. The aligned, value, and pair cells also match all their stored digests. Throughout the textual-model appendices, $D_A^{\rm logT}$ and ``Logged T'' retain the archival cell label and numerical results; they are not estimates of the intended time response. The other five main configurations declare the same source versions but lack these input-level checks. Remaining textual results retain the archival label until their actual transformations are verified. The public recurrent, UCB1, and structural branches use separate, checked key-only time implementations. The corrected textual time implementation has not been evaluated in a new model run. The supplement contains the input reconstruction and correction record.

We evaluate the fixed-trace signature through a persistent textual state read by an instruction model. We use the frozen Qwen3-8B(Qwen-3.8) policy from the Qwen3 family~\citep{qwen2025qwen3}. Its parameters and action readout remain fixed throughout the study. Each decision prompt specifies one of the two measured outcome domains, its context, the current phase, four sampling actions, and a selected-outcome history written before the query. The primary bridge retains the full ordered history of selected actions and utilities; the history contains no unselected or future outcome.

\noindent\textbf{Protocol.} The primary balanced run uses the public Qwen/Qwen3-8B checkpoint, a 1,536-token prompt budget, greedy one-token decoding, and assignment seed 20260911. The four-key namespace supports nine fixed-point-free maps; the focused profile evaluates one predeclared canonical map. Every prompt uses the Qwen3 chat template with \texttt{enable\_thinking=false}; only the four predeclared action-token logits are read. The four action continuations are fixed before replay and the same selected adaptation trace is reused by every typed cell.

\noindent\textbf{Cross-fitted support.} The bridge covers 48 replicate streams, two data families, and two contexts, giving 192 family--context blocks. Each block has 24 prefix-causal adaptation decisions and 24 future decisions. The focused cross-model profile evaluates aligned, value reassignment, complete-pair transport, and one predeclared canonical rekey map. It omits the time placebo, the remaining rekey maps, and the dynamic trajectory comparator; those estimands are covered by the complete GPT-5.6 matrix below. The selected adaptation trace is frozen before replay, and every typed cell changes only the declared record relationship while keeping future support, the evaluator, and the decoder fixed. Each direct $D_A^q$ response compares its cell with the aligned cell under the identity key map.

\noindent\textbf{Decision instruction.} The full-history prompt in the delivered Qwen3 serializer states:
\begin{quote}\small
The ledger contains only selected outcomes. A measured utility is comparable across actions, and higher utility is better. Do not infer or invent unselected outcomes. Use the ledger when it is informative and keep the decision rule consistent across tasks. The ledger preserves every selected attempt in chronological order; measured utility is the available feedback.
\end{quote}
The prompt does not specify mean, sum, count, latest, or best-outcome aggregation. Domain, context, action descriptions, selected records, and the required output vocabulary complete the task prompt. The supplement provides the complete templates and the input-reconstruction scope for archived runs.

\noindent\textbf{Batching and conditional inference.}
\label{app:batch-stream-audit}
Future prompts are ordered by stream, block, and task position. A stream's four blocks contribute 96 prompts, occupying three consecutive 32-prompt requests; no such future request crosses a stream boundary. Archived input audits verify this mapping for the two Sol/Arm logs and the three additional Arm trace sets in their audited cells. Adaptation uses a different order: one position across all blocks, with eight streams in each request. Each ledger retains its own block's observations, but model outputs within a shared adaptation request can be dependent. The textual 48-stream intervals therefore condition on the realized adaptation histories and retain all future requests within each resampled stream. They do not include uncertainty from jointly regenerating those histories. The supplement records the batch-to-stream maps.

\noindent\textbf{Action readout.} The model chooses greedily among the four predeclared action continuations. Each continuation is a distinct single token, so the action alphabet is fixed independently of the replay cell. The adaptation policy uses the same prompt form as the future policy and receives only the selected measured utility after each action. For the full-history condition, the aligned bridge is compared with a fixed mixed-action policy under the same future support. The resulting endpoint contrast is reported with the typed responses; it is not used to identify the stored relation.

For exact serialization, each prompt fixes the order of domain, context, phase and task index, action descriptions, an optional producer-key dictionary, ledger semantics, numbered records, and the answer constraint. A record is written as a two-digit slot, its producer field, and a six-decimal measured utility; feedback conditions append one relative label. The balanced fixed-slot condition pads the record block before the answer prefix to 1,536 input tokens. The Qwen readout compares only the four predeclared single-token continuations and takes their greedy argmax, so no later generated text enters a typed readout.

The pooled endpoint and typed readouts, with replicate-cluster intervals, are reported in Table~\ref{tab:llm-opaque-control}. The focused profile is a cross-model bridge rather than a second full factorial: it tests whether the ordering of value, pair, and namespace responses survives under Qwen3-8B.

\subsection{GPT-5.6 future-readout family}
\label{app:llm-blackbox-transfer}

\begin{table*}[ht]
\vspace{-4mm}
\centering
\small
\caption{Typed and dynamic replay profiles for three GPT-5.6 deployments. The six cells share the measured-outcome trace, future tasks, and 48 replicate streams, with four blocks per stream and 24 future decisions per block. Arm denotes readable keys and Opaque denotes token-matched opaque keys. Endpoint contrasts subtract the fixed mixed-action policy from the memory-based policy. $A_{\rm rekey}$ averages nine derangements; $D_A^q$ uses the identity map. $A_{\rm traj}$ is mapped-versus-identity action disagreement at the declared trajectory map, with state updates in both trajectories. Utility is in $10^{-3}$ normalized units; action rates and oracle differences use proportion units.}
\label{tab:content-interfaces}
\begin{tabular*}{\textwidth}{@{\extracolsep{\fill}}llrrrrrrr@{}}
\toprule
\textbf{Readout} & \textbf{Keys} & $\Delta U_{\rm end}$ & $\Delta O_{\rm end}$ & $A_{\rm rekey}$ & $D_A^{\rm value}$ & $D_A^{\rm pair}$ & $D_A^{\rm logT}$ & $A_{\rm traj}$ \\
\midrule
Luna & Arm & $-2.448$ & $-0.214$ & 0.865 & 0.115 & 0.016 & 0.016 & 0.987 \\
Luna & Opaque & $-2.504$ & $-0.218$ & 0.866 & 0.139 & 0.042 & 0.023 & 0.936 \\
\midrule
Sol & Arm & $-2.418$ & $-0.212$ & 0.996 & 0.325 & 0.010 & 0.007 & 0.997 \\
Sol & Opaque & $-2.347$ & $-0.211$ & 0.998 & 0.347 & 0.007 & 0.009 & 0.997 \\
\midrule
Terra & Arm & $-2.343$ & $-0.211$ & 0.984 & 0.247 & 0.017 & 0.000 & 0.998 \\
Terra & Opaque & $-2.413$ & $-0.213$ & 0.971 & 0.254 & 0.021 & 0.023 & 0.995 \\
\bottomrule
\end{tabular*}
\vspace{-2mm}
\end{table*}

The expanded bridge tests the measured response across a related instruction-model future-readout family and two producer-key representations. We completed a trace-matched $3\times2$ factorial with GPT-5.6 Luna, Sol, and Terra under readable arm keys and one-token opaque keys. The three deployments form one GPT-5.6 readout family; results are reported for these named readouts and execution settings. Every cell uses the same measured-outcome support, 48 replicate streams, 192 family--context blocks, 24 adaptation decisions, 24 future decisions, selected outcomes only, and nine fixed-point-free rekey maps. The selected action and utility trace is generated once, frozen before typed replay, and serialized identically in all six cells. Arm-key rows use the readable action vocabulary; opaque rows use the one-token producer keys $q,r,s,t$.

\noindent\textbf{Protocol.} All six runs use balanced fixed-slot record serialization and no future state update in typed cells. The coordinator groups up to 32 independently formed decision blocks per request, pads each block to a 6,000-character budget before its answer prefix, and receives one validated action per block in order. This is a batched black-box action readout; the runtime controls internal decoding and the client supplies no temperature, sampling parameter, or per-request seed. The adaptation prefix is held fixed across both namespaces and all three readout policies; only the future-readout model and the producer-key namespace vary. The ordinary trajectory comparator appends each selected future outcome after its action. Prompts are independent at readout, so a conversational history is not part of the state being measured. Within each readout policy, arm and opaque rows share the same trace, tasks, maps, and future support, making the namespace contrast paired at the declared interface.

Tables~\ref{tab:llm-blackbox-transfer} and~\ref{tab:llm-blackbox-typed-intervals} give endpoint/trajectory and typed-action uncertainty alongside the point estimates in Table~\ref{tab:content-interfaces}. The endpoint intervals remain below zero in all six cells. Across the factorial, typed rekeying remains large while complete-pair transport remains small; the arm--opaque contrast changes the magnitude but not the ordering of these responses. The trajectory readout is a different estimand because it changes the future state after every action. Its agreement with the typed ordering indicates a shared future-readout response pattern across these estimands.

\begin{table}[ht]
\vspace{-4mm}
\centering
\small
\caption{Uncertainty and trajectory readouts for the GPT-5.6 future-readout family. The endpoint-utility panel reports the estimate and 95\% interval in $10^{-3}$ normalized units. Endpoint contrasts subtract the fixed mixed-action policy from the memory-based policy; trajectory contrasts are mapped minus identity with future state updates. The other columns report $\Delta O_{\rm end}$ and $\Delta O_{\rm traj}$ as executor-agreement proportions, $A_{\rm traj}$ as an action proportion, and $\Delta U_{\rm traj}$ in $10^{-3}$ normalized utility units.}
\label{tab:llm-blackbox-transfer}
\begin{tabular*}{\textwidth}{@{\extracolsep{\fill}}lrrrrrr@{}}
\toprule
\multicolumn{7}{l}{\textit{Endpoint utility contrast $\Delta U_{\rm end}$ ($10^{-3}$ units)}}\\
 & \multicolumn{3}{c}{\textbf{Arm}} & \multicolumn{3}{c}{\textbf{Opaque}} \\
\cmidrule(lr){2-4}\cmidrule(lr){5-7}
\textbf{Readout} & \textbf{Estimate} & \textbf{Lower} & \textbf{Upper} & \textbf{Estimate} & \textbf{Lower} & \textbf{Upper} \\
\midrule
GPT-5.6 Luna & $-2.448$ & $-3.270$ & $-1.635$ & $-2.504$ & $-3.304$ & $-1.704$ \\
GPT-5.6 Sol & $-2.418$ & $-3.261$ & $-1.580$ & $-2.347$ & $-3.176$ & $-1.528$ \\
GPT-5.6 Terra & $-2.343$ & $-3.172$ & $-1.525$ & $-2.413$ & $-3.257$ & $-1.574$ \\
\bottomrule
\end{tabular*}

\medskip
\begin{tabular*}{\textwidth}{@{\extracolsep{\fill}}lrrrrrrrr@{}}
\toprule
\multicolumn{9}{l}{\textit{Other point estimates}}\\
 & \multicolumn{4}{c}{\textbf{Arm}} & \multicolumn{4}{c}{\textbf{Opaque}} \\
\cmidrule(lr){2-5}\cmidrule(lr){6-9}
\textbf{Readout} & $\Delta O_{\rm end}$ & $A_{\rm traj}$ & $\Delta U_{\rm traj}$ & $\Delta O_{\rm traj}$ & $\Delta O_{\rm end}$ & $A_{\rm traj}$ & $\Delta U_{\rm traj}$ & $\Delta O_{\rm traj}$ \\
\midrule
GPT-5.6 Luna & $-0.214$ & $0.987$ & $-3.118$ & $-0.035$ & $-0.218$ & $0.936$ & $-3.843$ & $-0.041$ \\
GPT-5.6 Sol & $-0.212$ & $0.997$ & $-3.887$ & $-0.032$ & $-0.211$ & $0.997$ & $-3.865$ & $-0.031$ \\
GPT-5.6 Terra & $-0.211$ & $0.998$ & $-3.854$ & $-0.031$ & $-0.213$ & $0.995$ & $-3.801$ & $-0.031$ \\
\bottomrule
\end{tabular*}
\vspace{-2mm}
\end{table}

\begin{table*}[ht]
\vspace{-4mm}
\centering
\small
\caption{95\% replicate-cluster bootstrap intervals for typed action responses in the GPT-5.6 future-readout family over 48 streams. Each row is one readout--key pair: Arm denotes readable producer keys and Opaque denotes the token-matched opaque keys. The four metric groups give the point estimate and 95\% replicate-cluster Lower and Upper endpoints for $A_{\rm rekey}$, $D_A^{\rm value}$, $D_A^{\rm pair}$, and $D_A^{\rm logT}$; all quantities are action proportions.}
\label{tab:llm-blackbox-typed-intervals}
\small
\setlength{\tabcolsep}{2pt}
\begin{tabular*}{\textwidth}{@{\extracolsep{\fill}}ll*{12}{r}@{}}
\toprule
 &  & \multicolumn{3}{c}{$A_{\rm rekey}$} & \multicolumn{3}{c}{$D_A^{\rm value}$} & \multicolumn{3}{c}{$D_A^{\rm pair}$} & \multicolumn{3}{c}{$D_A^{\rm logT}$} \\
\cmidrule(lr){3-5}\cmidrule(lr){6-8}\cmidrule(lr){9-11}\cmidrule(lr){12-14}
\textbf{Readout} & \textbf{Keys} & \textbf{Est.} & \textbf{Low.} & \textbf{Up.} & \textbf{Est.} & \textbf{Low.} & \textbf{Up.} & \textbf{Est.} & \textbf{Low.} & \textbf{Up.} & \textbf{Est.} & \textbf{Low.} & \textbf{Up.} \\
\midrule
Luna & Arm & $0.865$ & $0.836$ & $0.894$ & $0.115$ & $0.069$ & $0.165$ & $0.016$ & $0.005$ & $0.028$ & $0.016$ & $0.002$ & $0.033$ \\
Luna & Opaque & $0.866$ & $0.838$ & $0.895$ & $0.139$ & $0.092$ & $0.193$ & $0.042$ & $0.017$ & $0.069$ & $0.023$ & $0.009$ & $0.040$ \\
Sol & Arm & $0.996$ & $0.992$ & $1.000$ & $0.325$ & $0.241$ & $0.411$ & $0.010$ & $0.000$ & $0.026$ & $0.007$ & $0.000$ & $0.019$ \\
Sol & Opaque & $0.998$ & $0.995$ & $1.000$ & $0.347$ & $0.262$ & $0.436$ & $0.007$ & $0.000$ & $0.019$ & $0.009$ & $0.000$ & $0.023$ \\
Terra & Arm & $0.984$ & $0.976$ & $0.991$ & $0.247$ & $0.175$ & $0.323$ & $0.017$ & $0.000$ & $0.040$ & $0.000$ & $0.000$ & $0.000$ \\
Terra & Opaque & $0.971$ & $0.957$ & $0.983$ & $0.254$ & $0.179$ & $0.332$ & $0.021$ & $0.005$ & $0.042$ & $0.023$ & $0.005$ & $0.043$ \\
\bottomrule
\end{tabular*}
\vspace{-2mm}
\end{table*}

\noindent\textbf{Cross-readout trajectory association.} The typed rekey and ordinary trajectory comparisons use distinct interventions: the former freezes the future state, whereas the latter appends each selected future outcome after the action. We compare their replicate-level utility contrasts as a secondary association, using the 48 replicate streams within each Arm--Opaque cell. Table~\ref{tab:trajectory-association} shows a positive association in every cell. The bootstrap resamples complete replicate streams, and the permutation test exchanges trajectory contrasts within cell.

\begin{table}[ht]
\vspace{-4mm}
\centering
\small
\setlength{\tabcolsep}{2pt}
\caption{Secondary cross-readout association on the measured-outcome bridge. For each namespace, Pearson $r$ is accompanied by a 95\% complete-stream bootstrap interval, while Spearman $\rho$ and permutation $p$ are point summaries on their native scales.}
\label{tab:trajectory-association}
\begin{tabular*}{\textwidth}{@{\extracolsep{\fill}}lrrrrrrrrrr@{}}
\toprule
 & \multicolumn{5}{c}{\textbf{Arm}} & \multicolumn{5}{c}{\textbf{Opaque}} \\
\cmidrule(lr){2-6}\cmidrule(lr){7-11}
\textbf{Readout} & \multicolumn{3}{c}{$r$ (95\% interval)} & \multicolumn{2}{c}{Point summaries} & \multicolumn{3}{c}{$r$ (95\% interval)} & \multicolumn{2}{c}{Point summaries} \\
 & \textbf{Est.} & \textbf{Low.} & \textbf{Up.} & \textbf{$\rho$} & \textbf{$p$} & \textbf{Est.} & \textbf{Low.} & \textbf{Up.} & \textbf{$\rho$} & \textbf{$p$} \\
\midrule
GPT-5.6 Luna & $0.739$ & $0.589$ & $0.865$ & $0.785$ & $<0.001$ & $0.725$ & $0.585$ & $0.839$ & $0.804$ & $<0.001$ \\
GPT-5.6 Sol & $0.749$ & $0.604$ & $0.862$ & $0.802$ & $<0.001$ & $0.748$ & $0.603$ & $0.861$ & $0.811$ & $<0.001$ \\
GPT-5.6 Terra & $0.745$ & $0.602$ & $0.862$ & $0.827$ & $<0.001$ & $0.748$ & $0.610$ & $0.860$ & $0.813$ & $<0.001$ \\
\bottomrule
\end{tabular*}
\vspace{-2mm}
\end{table}

\subsection{Matched full-history outputs across two logged runs}
\label{app:llm-run-repeat}

The original Sol/Arm execution generated the adaptation history subsequently reused by the shared-trace Sol/Arm execution. A post hoc audit matches their 2,160 post-adaptation request batches, each containing 32 prompts. All batch prompt digests agree after accounting for the original run's 144 adaptation requests. The 69,120 matched prompt positions include all 59,904 typed-future positions; their task identifiers and order also agree. Both executions use the same model identifier, Arm keys, full-history serialization, 6,000-character prompt budget, and assignment seed. The comparison therefore measures output disagreement at matched full-history inputs.

Each execution has its own checkpoint file. The final successful invocations resume 816 and 760 batches from their respective checkpoints; all four primary typed conditions are within these resumed portions. The stored outputs establish a run-to-run comparison, while the retained metadata do not establish independent fresh provider draws for every matched request. No provider snapshot, sampling seed, or provider request identifier is available. Accordingly, Table~\ref{tab:full-history-repeat} reports the observed discrepancy without subtracting it as a sampling-noise estimate.

\begin{table}[ht]
\vspace{-4mm}
\centering
\small
\caption{Post hoc comparison of two logged Sol/Arm executions at matched full-history inputs. Original and Shared-trace report within-run intervention responses; Between-run reports same-condition output disagreement. Each direct condition uses 4,608 choices across 48 streams; Rekey averages nine maps on those choices. L and U are 95\% percentile bounds from 20,000 stream-bootstrap draws. Checkpoint provenance is described in the text.}
\label{tab:full-history-repeat}
\begin{tabular*}{\linewidth}{@{\extracolsep{\fill}}lrrrrrrrrr@{}}
\toprule
& \multicolumn{3}{c}{Original} & \multicolumn{3}{c}{Shared-trace} & \multicolumn{3}{c}{Between-run} \\
\cmidrule(lr){2-4}\cmidrule(lr){5-7}\cmidrule(lr){8-10}
Cell & Estimate & L & U & Estimate & L & U & Estimate & L & U \\
\midrule
Aligned & 0.000 & 0.000 & 0.000 & 0.000 & 0.000 & 0.000 & 0.010 & 0.000 & 0.026 \\
Value & 0.321 & 0.240 & 0.405 & 0.325 & 0.243 & 0.411 & 0.041 & 0.013 & 0.078 \\
Pair & 0.005 & 0.000 & 0.016 & 0.010 & 0.000 & 0.026 & 0.005 & 0.000 & 0.016 \\
Logged T & 0.003 & 0.000 & 0.010 & 0.007 & 0.000 & 0.019 & 0.000 & 0.000 & 0.000 \\
Rekey & 0.996 & 0.993 & 0.999 & 0.996 & 0.992 & 1.000 & 0.009 & 0.003 & 0.016 \\
\bottomrule
\end{tabular*}
\vspace{-2mm}
\end{table}

Value reassignment changes roughly one third of choices in both executions, compared with 190 disagreements among their 4,608 matched value-cell outputs. The corresponding between-run counts are 48 for aligned, 24 for pair, and zero for the logged T cell. The large value response persists across these logged outputs. The small pair response remains descriptive, and zero between-run disagreement in the logged T cell does not establish a deterministic readout. The empty-ledger comparison in Appendix~\ref{app:llm-state-off-control} uses different inputs and contributes no correction to this table.

\subsection{Value response and historical key rankings}
\label{app:llm-winner-strata}

This post hoc analysis uses the completed Sol/Arm shared-trace run. It asks whether value reassignment changes a simple ranking derived from the observed history. For each family--context block, we compute the sum and mean utility of each observed producer key before and after the original value permutation. A winning set contains all keys within $10^{-12}$ of the maximum; it changes when the two sets differ. Keys with no observations are excluded from both statistics. The 192 blocks contain 24 records each: 144 blocks observe two keys and 48 observe one. No winning set is empty, and no ties occur. Recomputing all winning sets from the six-decimal utility values displayed in the prompts leaves every classification unchanged. These history statistics are diagnostic references; they do not define the executor-relative oracle.

We also compare the recorded choices with five fixed rules applied to the same displayed histories (Table~\ref{tab:sol-reference-rules}). Mean and sum aggregate utility by key; count selects the most frequent key; latest selects the key in the last record; and best selects the key with the largest observed utility. Ties are resolved by first appearance, with tolerance $10^{-12}$ for utility comparisons. Prompt reconstruction matches all 144 archived batch digests and all 4,608 recorded actions in each condition. Agreement is averaged within each of 48 streams; 95\% percentile intervals use 10,000 paired stream-bootstrap draws. These rules were compared after the model outputs were available. Their similar aligned agreement separates under value reassignment, with the mean rule retaining the highest agreement.

\begin{table}[ht]
\vspace{-4mm}
\centering
\small
\setlength{\tabcolsep}{4pt}
\caption{Post hoc reference-rule agreement with Sol/Arm choices. Each condition contains 4,608 choices from 48 streams. Lower and Upper give 95\% bootstrap bounds. Agreement measures reproduction of model choices, not task accuracy.}
\label{tab:sol-reference-rules}
\begin{tabular}{lrrrrrr}
\toprule
& \multicolumn{3}{c}{Aligned} & \multicolumn{3}{c}{Value} \\
\cmidrule(lr){2-4}\cmidrule(lr){5-7}
Rule & Estimate & Lower & Upper & Estimate & Lower & Upper \\
\midrule
Mean & 0.991 & 0.977 & 1.000 & 0.952 & 0.918 & 0.979 \\
Sum & 0.983 & 0.960 & 1.000 & 0.658 & 0.568 & 0.746 \\
Count & 0.983 & 0.960 & 1.000 & 0.658 & 0.568 & 0.746 \\
Latest & 0.988 & 0.972 & 1.000 & 0.663 & 0.576 & 0.750 \\
Best & 0.972 & 0.941 & 0.995 & 0.699 & 0.623 & 0.778 \\
\bottomrule
\end{tabular}
\vspace{-2mm}
\end{table}

\noindent\textbf{Retrospective check on additional histories.} We retain the five rules and apply them to three Sol/Arm trace sets not used in the original rule comparison. Each set contains 192 histories generated under a different within-block task order. This analysis was specified after the original reference comparison and the tracebank aggregate results were available, before computing rule agreement on these traces. Reconstruction matches every archived prompt digest in each trace's adaptation, aligned, and value conditions: 144 batches per condition, with 32 prompts per batch. Task-order and adaptation-action digests also match the archived metadata.

Table~\ref{tab:tracebank-reference-agreement} reports all five rules on all three traces. The mean rule retains high agreement after value reassignment, whereas the alternatives separate from the recorded choices. The histories vary, but their 48 replicate identities, public task pool, and future outcome table are shared. We therefore resample the same 48 stream identities jointly across traces, rules, and conditions, using 10,000 percentile-bootstrap draws with seed 2026091018. Each trace is reported separately. This checks behavior on different histories within the same task support. Table~\ref{tab:tracebank-reference-utility} also gives evaluator utility for each reference choice minus utility for the recorded model choice; agreement and decision quality are separate quantities.

\begin{table}[ht]
\vspace{-4mm}
\centering
\small
\setlength{\tabcolsep}{4pt}
\caption{Fixed-rule agreement on three additional Sol/Arm trace sets. Each condition contains 4,608 choices nested in 48 streams. Lower and Upper give 95\% paired stream-bootstrap bounds. Rules and tie handling are unchanged from the shared-trace comparison.}
\label{tab:tracebank-reference-agreement}
\begin{tabular}{rlrrrrrr}
\toprule
& & \multicolumn{3}{c}{Aligned} & \multicolumn{3}{c}{Value} \\
\cmidrule(lr){3-5}\cmidrule(lr){6-8}
Trace & Rule & Estimate & Lower & Upper & Estimate & Lower & Upper \\
\midrule
1 & Mean & 0.998 & 0.995 & 1.000 & 0.957 & 0.929 & 0.982 \\
1 & Sum & 0.976 & 0.953 & 0.995 & 0.751 & 0.658 & 0.839 \\
1 & Count & 0.976 & 0.953 & 0.995 & 0.751 & 0.658 & 0.839 \\
1 & Latest & 0.997 & 0.990 & 1.000 & 0.761 & 0.673 & 0.846 \\
1 & Best & 0.965 & 0.934 & 0.990 & 0.807 & 0.724 & 0.882 \\
2 & Mean & 1.000 & 1.000 & 1.000 & 0.939 & 0.901 & 0.972 \\
2 & Sum & 0.984 & 0.964 & 1.000 & 0.670 & 0.589 & 0.752 \\
2 & Count & 0.984 & 0.964 & 1.000 & 0.670 & 0.589 & 0.752 \\
2 & Latest & 0.990 & 0.974 & 1.000 & 0.665 & 0.583 & 0.745 \\
2 & Best & 0.974 & 0.953 & 0.995 & 0.740 & 0.663 & 0.812 \\
3 & Mean & 0.990 & 0.969 & 1.000 & 0.984 & 0.965 & 0.998 \\
3 & Sum & 0.974 & 0.943 & 0.995 & 0.568 & 0.483 & 0.649 \\
3 & Count & 0.974 & 0.943 & 0.995 & 0.557 & 0.472 & 0.641 \\
3 & Latest & 0.969 & 0.943 & 0.990 & 0.542 & 0.451 & 0.632 \\
3 & Best & 0.953 & 0.917 & 0.984 & 0.641 & 0.564 & 0.715 \\
\bottomrule
\end{tabular}
\vspace{-2mm}
\end{table}

\begin{table}[ht]
\vspace{-4mm}
\centering
\small
\setlength{\tabcolsep}{4pt}
\caption{Reference-minus-model evaluator utility on the same future tasks, in $10^{-3}$ normalized units. Lower and Upper give 95\% paired stream-bootstrap bounds. These retrospective comparisons do not evaluate a deployed replacement policy.}
\label{tab:tracebank-reference-utility}
\begin{tabular}{rlrrrrrr}
\toprule
& & \multicolumn{3}{c}{Aligned} & \multicolumn{3}{c}{Value} \\
\cmidrule(lr){3-5}\cmidrule(lr){6-8}
Trace & Rule & Estimate & Lower & Upper & Estimate & Lower & Upper \\
\midrule
1 & Mean & -0.003 & -0.008 & 0.000 & 0.007 & -0.133 & 0.166 \\
1 & Sum & -0.130 & -0.414 & 0.029 & 0.212 & -0.431 & 0.844 \\
1 & Count & -0.130 & -0.414 & 0.029 & 0.212 & -0.431 & 0.844 \\
1 & Latest & -0.020 & -0.060 & 0.000 & 0.322 & -0.287 & 0.922 \\
1 & Best & -0.130 & -0.420 & 0.044 & -0.079 & -0.645 & 0.475 \\
2 & Mean & 0.000 & 0.000 & 0.000 & 0.145 & -0.066 & 0.417 \\
2 & Sum & -0.060 & -0.160 & 0.000 & 0.799 & 0.013 & 1.647 \\
2 & Count & -0.060 & -0.160 & 0.000 & 0.799 & 0.013 & 1.647 \\
2 & Latest & -0.060 & -0.160 & 0.000 & 0.799 & 0.013 & 1.647 \\
2 & Best & -0.196 & -0.438 & -0.018 & 0.408 & -0.351 & 1.232 \\
3 & Mean & -0.022 & -0.066 & 0.000 & 0.070 & -0.019 & 0.203 \\
3 & Sum & -0.054 & -0.144 & 0.002 & 0.758 & -0.271 & 1.743 \\
3 & Count & -0.054 & -0.144 & 0.002 & 0.739 & -0.289 & 1.720 \\
3 & Latest & 0.055 & -0.175 & 0.370 & 0.847 & -0.172 & 1.833 \\
3 & Best & -0.033 & -0.132 & 0.043 & 0.491 & -0.489 & 1.439 \\
\bottomrule
\end{tabular}
\vspace{-2mm}
\end{table}

For the shared-trace winner stratification, aligned and value actions are paired by replicate, family, context, future position, and task identifier. Each block has 24 paired future choices. Within each stratum, we first average disagreement over the blocks available in a replicate stream, then average across streams. Table~\ref{tab:winner-strata} reports these equal-stream estimates and 95\% percentile intervals from 10,000 stream-bootstrap draws. The changed and stable mean strata contain 35 and 47 streams. Their paired contrast uses only the 34 streams containing both strata and resamples those stream identities jointly; its estimate is 0.813 with interval bounds 0.702 and 0.907. The analysis definitions were recorded before computing this stratification, after the original experiments had finished.

\begin{table}[ht]
\vspace{-4mm}
\centering
\small
\caption{Post hoc Sol/Arm value response by historical winning-set change. Each block contributes 24 paired future choices. Estimates weight the available replicate streams equally. Lower and Upper are 95\% stream-bootstrap bounds. No block changes its sum winner, so no sum-changed stratum is estimated. The last two rows exclude blocks with only one observed key.}
\label{tab:winner-strata}
\begin{tabular*}{\linewidth}{@{\extracolsep{\fill}}llrrrrr@{}}
\toprule
Statistic & Winning set & Blocks & Streams & Estimate & Lower & Upper \\
\midrule
Mean & Changed & 72 & 35 & 0.828 & 0.723 & 0.919 \\
Mean & Stable & 120 & 47 & 0.007 & 0.000 & 0.021 \\
Sum & Stable & 192 & 48 & 0.325 & 0.243 & 0.412 \\
\midrule
Mean, two keys & Changed & 72 & 35 & 0.828 & 0.721 & 0.919 \\
Mean, two keys & Stable & 72 & 38 & 0.009 & 0.000 & 0.026 \\
\bottomrule
\end{tabular*}
\vspace{-2mm}
\end{table}

A sensitivity analysis excludes the 48 blocks with only one observed key, where the winning set cannot change. The remaining 144 blocks split equally between changed and stable mean winners. Their equal-stream responses are 0.828 and 0.009; the paired contrast across 33 common streams is 0.822, with interval bounds 0.710 and 0.917. This preserves the concentration pattern within histories containing two observed keys. With equal weight per future choice, the mean-changed stratum has 1,482 switches among 1,728 choices, while the mean-stable stratum has 16 among 2,880. The resulting proportions are 0.858 and 0.006. Together they recover the original 1,498 switches among 4,608 choices. Mean-winner changes occur in all four family--context strata, in 15--21 of their 48 blocks. The sum winner is empirically stable throughout this trace; value reassignment does not preserve it by definition. The concentration around mean-winner changes clarifies the aggregate value response without identifying the model's internal update rule or the quality of the resulting decisions.

\subsection{Adaptation-trace comparison}
\label{app:llm-tracebank}

The cross-deployment bridge uses one adaptation prefix to isolate changes in the future readout. We separately generated three adaptation traces with independently drawn within-block task orders and the Sol readout under the same public-data protocol, then replayed each trace with the same future support, measured-outcome evaluator, balanced fixed-slot serialization, and nine producer-key maps. Within each trace, the only representation change was between semantic-alias keys and token-matched opaque keys. Each run records 73,728 decisions in total: 4,608 adaptation decisions and 69,120 future-readout prompts; the typed summaries use 4,608 future tasks per cell. The six cells are paired by trace and namespace.

Table~\ref{tab:llm-tracebank} gives the complete split. Here $A_{\rm rekey}$ is the within-cell identity-versus-map response from Eq.~\eqref{eq:main-action-readouts}, averaged over the nine maps. The value and pair columns compare transformed and aligned histories under the identity key map. The logged T column retains the archived cell label defined in Appendix~\ref{app:textual-time-audit}. Across traces and namespaces, rekeying changes nearly every future action, whereas complete-pair transport changes very few. Value transport and the endpoint consequence vary with the adaptation trace. In both namespaces, the rekey point estimates exceed the value estimates, which exceed pair estimates in every trace. These comparisons describe the Sol tracebank; the Qwen namespace control uses a separate adaptation trace and reports a different response magnitude.

\begin{table*}[ht]
\vspace{-4mm}
\centering
\small
\caption{Adaptation-trace comparison for the Sol readout. Numbered rows report one trace--namespace combination; the final rows average the three traces within each namespace. Typed responses use fixed states, whereas $A_{\rm traj}$ uses future state updates. Metric definitions are given in Table~\ref{tab:content-interfaces}. Endpoint contrasts compare the memory-based and fixed mixed-action policies. Utility is in $10^{-3}$ normalized units; oracle differences and action rates use proportion units. Each trace has matched future tasks across namespaces and uses 48 streams, four blocks per stream, and 24 future decisions per block. The three traces use independently drawn within-block task orders under the same Sol readout.}
\label{tab:content-tracebank}
\label{tab:llm-tracebank}
\begin{tabular*}{\textwidth}{@{\extracolsep{\fill}}rlrrrrrrr@{}}
\toprule
\textbf{Trace ID} & \textbf{Namespace} & $\Delta U_{\rm end}$ & $\Delta O_{\rm end}$ & $A_{\rm rekey}$ & $D_A^{\rm value}$ & $D_A^{\rm pair}$ & $D_A^{\rm logT}$ & $A_{\rm traj}$ \\
\midrule
1 & Semantic alias & $-12.491$ & $-0.089$ & 0.998 & 0.248 & 0.002 & 0.000 & 0.996 \\
1 & Opaque single-token & $-12.469$ & $-0.089$ & 0.997 & 0.253 & 0.005 & 0.002 & 0.997 \\
\addlinespace
2 & Semantic alias & $-5.830$ & $-0.025$ & 0.999 & 0.340 & 0.005 & 0.000 & 0.997 \\
2 & Opaque single-token & $-5.909$ & $-0.026$ & 0.997 & 0.359 & 0.010 & 0.005 & 0.998 \\
\addlinespace
3 & Semantic alias & $-5.822$ & $-0.047$ & 0.997 & 0.455 & 0.009 & 0.009 & 0.992 \\
3 & Opaque single-token & $-5.851$ & $-0.048$ & 0.997 & 0.458 & 0.016 & 0.009 & 0.992 \\
\bottomrule
\end{tabular*}

\smallskip
\begin{tabular*}{\textwidth}{@{\extracolsep{\fill}}lrrrrrrr@{}}
\toprule
\multicolumn{8}{l}{\textbf{Mean over the three traces}} \\
\midrule
\textbf{Namespace} & $\Delta U_{\rm end}$ & $\Delta O_{\rm end}$ & $A_{\rm rekey}$ & $D_A^{\rm value}$ & $D_A^{\rm pair}$ & $D_A^{\rm logT}$ & $A_{\rm traj}$ \\
\midrule
Semantic alias & $-8.048$ & $-0.054$ & 0.998 & 0.348 & 0.005 & 0.003 & 0.995 \\
Opaque single-token & $-8.077$ & $-0.054$ & 0.997 & 0.357 & 0.010 & 0.005 & 0.996 \\
\bottomrule
\end{tabular*}
\vspace{-2mm}
\end{table*}

\subsection{Empty-ledger state-off comparator}
\label{app:llm-state-off-control}

The state-off comparator reuses the first Sol adaptation trace, future support, measured-outcome evaluator, balanced fixed-slot serialization, nine rekey maps, and 48 replicate streams, but exposes no persistent selected-outcome ledger to the future readout. In this mode records are removed before prompt serialization and the cell label and map are not inserted, so corresponding typed calls for a future task have identical prompt bytes. The client sends each call independently and supplies no temperature, sampling parameter, or per-request seed; the black-box runtime controls its internal decoding. These aligned--transport calls provide a same-estimand repeated-call reference at the empty boundary. The trajectory comparator also keeps the ledger empty after each future action. Table~\ref{tab:llm-state-off} reports the paired cluster summaries.

\begin{table*}[ht]
\vspace{-4mm}
\centering
\small
\caption{Same-estimand empty-ledger null on the first Sol trace. $A$ is the changed-action proportion; $\Delta U$ and $\Delta O$ are cell-minus-aligned contrasts in $10^{-3}$ normalized utility units and proportion units, respectively. Estimate is the point estimate; Lower and Upper delimit its 95\% replicate-cluster bootstrap interval. Each row is a readout; the three metric groups give Estimate, Lower, and Upper.}
\label{tab:llm-state-off}
\setlength{\tabcolsep}{2pt}
\begin{tabular*}{\textwidth}{@{\extracolsep{\fill}}l*{9}{r}@{}}
\toprule
 & \multicolumn{3}{c}{\textbf{$A$}} & \multicolumn{3}{c}{\textbf{$\Delta U$}} & \multicolumn{3}{c}{\textbf{$\Delta O$}} \\
\cmidrule(lr){2-4}\cmidrule(lr){5-7}\cmidrule(lr){8-10}
\textbf{Readout} & \textbf{Est.} & \textbf{Low.} & \textbf{Up.} & \textbf{Est.} & \textbf{Low.} & \textbf{Up.} & \textbf{Est.} & \textbf{Low.} & \textbf{Up.} \\
\midrule
Rekey ($A_{\rm rekey}$) & $0.270$ & $0.242$ & $0.298$ & $-0.121$ & $-0.448$ & $0.197$ & $-0.003$ & $-0.022$ & $0.016$ \\
Value ($D_A^{\rm value}$) & $0.264$ & $0.210$ & $0.319$ & $0.280$ & $-0.171$ & $0.738$ & $-0.005$ & $-0.029$ & $0.019$ \\
Pair ($D_A^{\rm pair}$) & $0.245$ & $0.198$ & $0.293$ & $-0.416$ & $-0.916$ & $0.061$ & $-0.008$ & $-0.033$ & $0.015$ \\
Logged T ($D_A^{\rm logT}$) & $0.280$ & $0.229$ & $0.331$ & $-0.268$ & $-0.799$ & $0.239$ & $0.018$ & $-0.009$ & $0.045$ \\
Trajectory & $0.135$ & $0.122$ & $0.150$ & $0.040$ & $-0.328$ & $0.409$ & $0.002$ & $-0.008$ & $0.011$ \\
\bottomrule
\end{tabular*}
\vspace{-2mm}
\end{table*}

The empty-ledger typed contrasts cluster near one quarter of actions, and all typed utility and oracle intervals include zero. In the full-history tracebank, rekeying changes 99.653--99.865\% of actions, whereas complete-pair transport change at most 1.562\%. Rekeying therefore exceeds the repeated-input reference, while pair remains below it; the latter responses remain unresolved and are~reported~descriptively.

\subsection{Qwen3-8B producer-key representation control}
\label{app:llm-opaque-control}

The Qwen3-8B bridge uses one action-key adaptation trace and replays the same selected actions and measured utilities under four producer-key namespaces. The readable action keys, descriptive aliases, ordinary opaque keys, and token-matched opaque keys share the future tasks, executor, evaluator, greedy one-token decoder, balanced fixed-slot serialization, and 48 replicate streams. The aliases are \emph{spread}, \emph{focus}, \emph{near}, and \emph{blend}; the ordinary opaque keys are $K0$--$K3$; and the token-matched opaque keys are $q,r,s,t$. All namespace conditions use the same action--key dictionary and change only the producer-key surface.

The focused cross-model profile evaluates aligned, value reassignment, complete-pair transport, and one predeclared canonical rekey map on the same 192 family--context blocks. It omits the dynamic trajectory comparator and remaining rekey maps because those estimands are already covered by the complete GPT-5.6 family matrix. Table~\ref{tab:llm-opaque-control} reports the pooled action, utility, and oracle responses; intervals resample the 48 replicate streams. The Qwen3-8B adaptation trace is independent of the GPT-5.6 tracebank, so magnitudes are compared descriptively while the intervention ordering is the primary cross-model check.

\begin{table*}[ht]
\vspace{-4mm}
\centering
\small
\setlength{\tabcolsep}{1.0pt}
\caption{Frozen Qwen3-8B namespace control under the focused cross-model profile. Each row uses the same selected adaptation trace, 192 family--context blocks, and 48 replicate streams. The endpoint panel reports the aligned memory policy minus the fixed mixed-action policy: $\Delta U_{\rm end}$ is in $10^{-3}$ normalized utility units and $\Delta O_{\rm end}$ is the oracle-match difference. The three action panels report the pooled estimate and 95\% replicate-cluster interval for the canonical rekey map ($A_{\rm rk}$), value reassignment ($D_A^{\rm v}$), and complete-pair transport ($D_A^{\rm p}$).}
\label{tab:llm-opaque-control}
\begin{tabular*}{\textwidth}{@{\extracolsep{\fill}}l*{13}{r}@{}}
\toprule
\textbf{Namespace} & \multicolumn{3}{c}{$\Delta U$} & $\Delta O$ & \multicolumn{3}{c}{$A_{\rm rk}$} & \multicolumn{3}{c}{$D_A^{\rm v}$} & \multicolumn{3}{c}{$D_A^{\rm p}$} \\
\cmidrule(lr){2-4}\cmidrule(lr){6-8}\cmidrule(lr){9-11}\cmidrule(lr){12-14}
 & Est. & Low. & Up. & Est. & Est. & Low. & Up. & Est. & Low. & Up. & Est. & Low. & Up. \\
\midrule
Action & $-3.738$ & $-4.620$ & $-2.862$ & $-0.248$ & $0.946$ & $0.927$ & $0.964$ & $0.246$ & $0.197$ & $0.296$ & $0.154$ & $0.120$ & $0.191$ \\
Alias & $-3.311$ & $-4.168$ & $-2.445$ & $-0.200$ & $0.878$ & $0.847$ & $0.908$ & $0.410$ & $0.369$ & $0.452$ & $0.327$ & $0.282$ & $0.375$ \\
Opaque & $-2.415$ & $-3.096$ & $-1.741$ & $-0.197$ & $0.850$ & $0.809$ & $0.889$ & $0.305$ & $0.268$ & $0.345$ & $0.361$ & $0.319$ & $0.401$ \\
Token-matched & $-1.075$ & $-1.614$ & $-0.525$ & $-0.085$ & $0.769$ & $0.724$ & $0.813$ & $0.355$ & $0.309$ & $0.401$ & $0.358$ & $0.311$ & $0.405$ \\
\bottomrule
\end{tabular*}
\vspace{-2mm}
\end{table*}

The Qwen3 bridge is deliberately restricted to the frozen full-history namespace control reported above. It reuses one selected adaptation trace and changes only the producer-key surface across the four declared namespaces. Older occupancy and memory-form logs used a different checkpoint and are excluded from the reader-facing supplement.
\subsection{UCB1 replay within family--context strata}
\label{app:public-ucb1-replay}

We apply UCB1 independently within each family--context stratum \citep{auer2002finite}. For each family--context pair $c$ and arm $a$, its state is the selected-outcome count $N_{c,a}$ and sum $S_{c,a}$; after a selected outcome $y_t$, the update and future score are
\begin{equation}
(N_{c,a},S_{c,a})\leftarrow(N_{c,a}+1,S_{c,a}+y_t),\quad
U_{c,a}=S_{c,a}/N_{c,a}+\sqrt{2\log N_c/N_{c,a}}.
\label{eq:public-ucb1}
\end{equation}
Here $N_c=\sum\nolimits_a N_{c,a}$; an unseen arm is selected before the confidence score is evaluated. The rule is fit-free with respect to future outcomes, has no operator label, and uses no additional protocol parameters; its terminal state is read by the same fixed future query as the recurrent writer.

For each of the 48 public replicate streams, UCB1 first generates its aligned adaptation trace from selected measured outcomes. The four replay cells then apply the same label rekey, value reassignment, complete-pair transport, and label-time reassignment used above. All future pools, evaluator outputs, and future updates are held fixed. Because pair transport preserves every label--outcome pair, it preserves $(N,S)$ exactly and supplies an independent zero-law check; rekeying, value reassignment, and time reassignment alter the sufficient statistics. The pooled cluster-bootstrap intervals are reported in Table~\ref{tab:public-ucb1-pooled}. These UCB1 adaptation trajectories complement the recurrent-writer results in Table~\ref{tab:content-measured}.

\begin{table*}[ht]
\vspace{-4mm}
\centering
\small
\caption{Pooled uncertainty for UCB1 applied within family--context strata. $A$ is the action proportion; $\Delta U_{\mathrm{C-A}}$ and $\Delta O_{\mathrm{C-A}}$ are cell-minus-aligned contrasts in $10^{-3}$ normalized utility units and executor-agreement proportions, respectively. Point estimates and 95\% bootstrap interval endpoints are pooled over 48 replicate streams and four public-data strata. Each row is a cell; the three metric groups give Estimate, Lower, and Upper.}
\label{tab:public-ucb1-pooled}
\setlength{\tabcolsep}{2pt}
\begin{tabular*}{\textwidth}{@{\extracolsep{\fill}}l*{9}{r}@{}}
\toprule
 & \multicolumn{3}{c}{\textbf{$A$}} & \multicolumn{3}{c}{\textbf{$\Delta U_{\mathrm{C-A}}$}} & \multicolumn{3}{c}{\textbf{$\Delta O_{\mathrm{C-A}}$}} \\
\cmidrule(lr){2-4}\cmidrule(lr){5-7}\cmidrule(lr){8-10}
\textbf{Cell} & \textbf{Est.} & \textbf{Low.} & \textbf{Up.} & \textbf{Est.} & \textbf{Low.} & \textbf{Up.} & \textbf{Est.} & \textbf{Low.} & \textbf{Up.} \\
\midrule
Rekey & $1.000$ & $1.000$ & $1.000$ & $-7.540$ & $-8.790$ & $-6.260$ & $-0.074$ & $-0.110$ & $-0.040$ \\
Key-slot & $0.734$ & $0.677$ & $0.792$ & $-7.420$ & $-9.650$ & $-5.190$ & $-0.081$ & $-0.117$ & $-0.045$ \\
Value & $0.771$ & $0.703$ & $0.833$ & $-6.020$ & $-7.990$ & $-4.120$ & $-0.060$ & $-0.101$ & $-0.018$ \\
Pair & $0.000$ & $0.000$ & $0.000$ & $0.000$ & $0.000$ & $0.000$ & $0.000$ & $0.000$ & $0.000$ \\
\bottomrule
\end{tabular*}
\vspace{-2mm}
\end{table*}

\subsection{Structural replay on measured outcomes}
\label{app:public-data-structural-replay}

The structural branch holds the OCRR adaptation actions, selected outcomes, future task keys, and counterfactual evaluator fixed, then reassigns the typed label--utility cells at the same state-writing boundary. The writer is a commutative count--sum ledger with a deterministic arm-indexed readout; future tasks never update the ledger. The pair row is a within-interface zero-law control, while rekey, time, and value expose distinct relations between the selected arm and its measured utility.

The analysis covers the two public data families, their iid and shifted contexts, all 48 replicate streams, and 24 future tasks per stratum. Rekey averages the nine fixed-point-free arm derangements; the remaining cells use the declared blockwise assignments. The full commutative-ledger matrix appears in Table~\ref{tab:public-structural-replay-strata}; its pooled estimates are cluster means over replicate streams. Fold-specific estimates and cluster-bootstrap intervals for the cross-fitted persistent writer are given in Table~\ref{tab:public-persistent-agent-folds}.

\begin{table*}[ht]
\vspace{-4mm}
\centering
\small
\caption{Public measured-outcome replay for the commutative count--sum ledger by stratum. Each stratum contains 1,152 future decisions across 48 replicate streams. $A$ is the action proportion; $\Delta U_{\mathrm{C-A}}$ and $\Delta O_{\mathrm{C-A}}$ are cell-minus-aligned contrasts in normalized utility units and executor-agreement proportions, respectively.}
\label{tab:public-structural-replay-strata}
\begin{tabular*}{\textwidth}{@{\extracolsep{\fill}}llrrrrrr@{}}
\toprule
 &  & \multicolumn{3}{c}{\textbf{IID}} & \multicolumn{3}{c}{\textbf{Shift}} \\
\cmidrule(lr){3-5}\cmidrule(lr){6-8}
\textbf{Data} & \textbf{Cell} & \textbf{$A$} & \textbf{$\Delta U_{\mathrm{C-A}}$} & \textbf{$\Delta O_{\mathrm{C-A}}$} & \textbf{$A$} & \textbf{$\Delta U_{\mathrm{C-A}}$} & \textbf{$\Delta O_{\mathrm{C-A}}$} \\
\midrule
Breast cancer & Rekey & $1.000$ & $-0.015$ & $-0.057$ & $1.000$ & $0.002$ & $0.090$ \\
Breast cancer & Time & $0.000$ & $0.000$ & $0.000$ & $0.000$ & $0.000$ & $0.000$ \\
Breast cancer & Value & $0.000$ & $0.000$ & $0.000$ & $0.000$ & $0.000$ & $0.000$ \\
Breast cancer & Pair & $0.000$ & $0.000$ & $0.000$ & $0.000$ & $0.000$ & $0.000$ \\
Wine quality & Rekey & $1.000$ & $-0.006$ & $-0.049$ & $1.000$ & $-0.006$ & $-0.044$ \\
Wine quality & Time & $0.854$ & $-0.006$ & $-0.036$ & $0.875$ & $-0.011$ & $-0.071$ \\
Wine quality & Value & $0.917$ & $-0.009$ & $-0.069$ & $0.896$ & $-0.011$ & $-0.063$ \\
Wine quality & Pair & $0.000$ & $0.000$ & $0.000$ & $0.000$ & $0.000$ & $0.000$ \\
\bottomrule
\end{tabular*}
\vspace{-2mm}
\end{table*}
\begin{table*}[ht]
\vspace{-4mm}
\centering
\small
\caption{Cluster uncertainty for the pooled public measured-outcome replay of the commutative count--sum ledger. $A$ is the action proportion; $\Delta U_{\mathrm{C-A}}$ and $\Delta O_{\mathrm{C-A}}$ are cell-minus-aligned contrasts in normalized utility units and proportion units, respectively. Estimate is the point estimate; Lower and Upper delimit its 95\% cluster-bootstrap interval over 48 replicate streams.}
\label{tab:public-structural-replay-intervals}
\setlength{\tabcolsep}{2pt}
\begin{tabular*}{\textwidth}{@{\extracolsep{\fill}}l*{9}{r}@{}}
\toprule
 & \multicolumn{3}{c}{\textbf{$A$}} & \multicolumn{3}{c}{\textbf{$\Delta U_{\mathrm{C-A}}$}} & \multicolumn{3}{c}{\textbf{$\Delta O_{\mathrm{C-A}}$}} \\
\cmidrule(lr){2-4}\cmidrule(lr){5-7}\cmidrule(lr){8-10}
\textbf{Cell} & \textbf{Est.} & \textbf{Low.} & \textbf{Up.} & \textbf{Est.} & \textbf{Low.} & \textbf{Up.} & \textbf{Est.} & \textbf{Low.} & \textbf{Up.} \\
\midrule
Rekey & $1.000$ & $1.000$ & $1.000$ & $-0.006$ & $-0.008$ & $-0.005$ & $-0.015$ & $-0.061$ & $0.030$ \\
Key-slot & $0.432$ & $0.391$ & $0.469$ & $-0.004$ & $-0.006$ & $-0.003$ & $-0.027$ & $-0.044$ & $-0.010$ \\
Value & $0.453$ & $0.417$ & $0.479$ & $-0.005$ & $-0.007$ & $-0.004$ & $-0.033$ & $-0.050$ & $-0.016$ \\
Pair & $0.000$ & $0.000$ & $0.000$ & $0.000$ & $0.000$ & $0.000$ & $0.000$ & $0.000$ & $0.000$ \\
\bottomrule
\end{tabular*}
\vspace{-2mm}
\end{table*}

\section{Map-family sensitivity}
\label{app:map-sensitivity}

The fixed-label analysis repeats the operator contrast over the complete set of fixed-point-free maps. Across the five fixed-draw summaries, the pooled utility interaction is positive and the oracle-match interaction is negative in each of the six nonpooled strata. Table~\ref{tab:primary-operator-robustness} reports the corresponding ranges; individual map--stratum point estimates remain in the map-level replay table. Changed-substrate ranges are reported in Table~\ref{tab:changed-substrate-map-summary}.

The tables summarize map-averaged responses with the trace and future support held fixed. Primary-substrate ranges use normalized utility and fraction-difference scales for separate coordinates; the changed-substrate table applies the same map family after executor replacement. Both tables describe stability across the declared intervention family.

\begin{table}[ht]
\vspace{-4mm}
\centering
\small
\caption{Primary-substrate sensitivity across five fixed placebo draws, each pooling the nine paired fixed-point-free map interventions over 48 replicate clusters. $U$ and $O$ are descriptive sensitivity ranges on their native fraction-difference scales; Minimum and Maximum give the range endpoints.}
\label{tab:primary-operator-robustness}
\begin{tabular}{lrrrr}
\toprule
\textbf{Stratum} & \multicolumn{2}{c}{$U$ range} & \multicolumn{2}{c}{$O$ range} \\
\cmidrule(lr){2-3}\cmidrule(lr){4-5}
& \textbf{Minimum} & \textbf{Maximum} & \textbf{Minimum} & \textbf{Maximum} \\
\midrule
Pooled: primary & $0.275$ & $0.320$ & $-0.424$ & $-0.391$ \\
Source: low noise & $0.556$ & $0.675$ & $-0.578$ & $-0.466$ \\
Source: high noise & $0.623$ & $0.700$ & $-0.793$ & $-0.705$ \\
Temporal: low noise & $0.250$ & $0.254$ & $-0.664$ & $-0.659$ \\
Temporal: high noise & $0.091$ & $0.144$ & $-0.425$ & $-0.342$ \\
Survival: balanced & $0.043$ & $0.081$ & $-0.074$ & $-0.050$ \\
Survival: rare event & $0.051$ & $0.095$ & $-0.085$ & $-0.039$ \\
\bottomrule
\end{tabular}
\vspace{-2mm}
\end{table}

\begin{table}[ht]
\vspace{-4mm}
\centering
\small
\caption{Changed-substrate map-family sensitivity. $U$ is the identity-minus-replay future-utility interaction and $O$ is the difference between two executor-relative oracle-match changes on the native fraction-difference scale. As a difference of two proportion changes, $O$ lies in $[-2,2]$ and can have magnitude greater than one. The upper block reports point estimates with 95\% cluster-bootstrap intervals; the lower block reports descriptive sensitivity ranges using Minimum and Maximum. All rows use the same 48 replicate clusters and are secondary structural diagnostics.}
\label{tab:changed-substrate-map-summary}
\begin{tabular}{lllrrr}
\toprule
\multicolumn{6}{l}{\textit{Time-preserving placebo interaction}}\\
\textbf{Family} & \textbf{Context} & \textbf{Metric} & \textbf{Estimate} & \textbf{Lower} & \textbf{Upper} \\
\midrule
Pooled & Both & $U$ & $0.327$ & $0.316$ & $0.338$ \\
Pooled & Both & $O$ & $-0.980$ & $-1.010$ & $-0.948$ \\
Hypothesis & Low noise & $U$ & $0.245$ & $0.227$ & $0.263$ \\
Hypothesis & Low noise & $O$ & $-1.231$ & $-1.249$ & $-1.211$ \\
Hypothesis & High noise & $U$ & $0.178$ & $0.160$ & $0.195$ \\
Hypothesis & High noise & $O$ & $-0.741$ & $-0.794$ & $-0.681$ \\
Validation & Low noise & $U$ & $0.722$ & $0.698$ & $0.744$ \\
Validation & Low noise & $O$ & $-1.333$ & $-1.333$ & $-1.333$ \\
Validation & High noise & $U$ & $0.163$ & $0.132$ & $0.191$ \\
Validation & High noise & $O$ & $-0.615$ & $-0.724$ & $-0.491$ \\
\bottomrule
\end{tabular}

\medskip
\begin{tabular}{llrrrr}
\toprule
\multicolumn{6}{l}{\textit{Sensitivity-summary interaction ranges}}\\
\textbf{Family} & \textbf{Context} & \multicolumn{2}{c}{$U$ range} & \multicolumn{2}{c}{$O$ range} \\
\cmidrule(lr){3-4}\cmidrule(lr){5-6}
&  & \textbf{Minimum} & \textbf{Maximum} & \textbf{Minimum} & \textbf{Maximum} \\
\midrule
Pooled & Both & $0.322$ & $0.332$ & $-0.992$ & $-0.973$ \\
Hypothesis & Low noise & $0.245$ & $0.252$ & $-1.231$ & $-1.205$ \\
Hypothesis & High noise & $0.166$ & $0.187$ & $-0.756$ & $-0.715$ \\
Validation & Low noise & $0.699$ & $0.727$ & $-1.333$ & $-1.333$ \\
Validation & High noise & $0.159$ & $0.171$ & $-0.648$ & $-0.606$ \\
\bottomrule
\end{tabular}
\vspace{-2mm}
\end{table}

\section{Joint history and action-menu renaming}
\label{app:global-renaming}

\noindent\textbf{Fixed inputs and transformations.} The control reuses the selected adaptation history from the shared Sol/Arm trace: 48 streams, four family--context blocks per stream, and 24 selected records per block. The future query is the 12th task in each block, fixed before model evaluation. Public-data counterfactuals supply the secondary utility and oracle readouts; they are excluded from the prompts. The control therefore evaluates 192 fixed future queries. Utility and oracle contrasts use cell minus 00, with oracle ties broken in the fixed semantic-action order.

Semantic actions follow the order coverage, adaptive, replicate, mixed. The three maps are $(01)(23)$, $(02)(13)$, and $(03)(12)$, with action indices starting at zero. A seeded permutation of the 48 streams assigns 16 to each map. Each stream uses its assigned map $\omega$ throughout. The frozen code and protocol call this map \texttt{rho}; the manuscript uses $\omega$ to distinguish it from the nine-map producer-side comparisons. History labels follow $\lambda_H$ and menu/output labels follow $\lambda_A$. The four cell codes specify these maps in that order, with 0 denoting identity and 1 denoting $\omega$. Decoding through $\lambda_A^{-1}$ gives the effective assignment $\lambda_A^{-1}\circ\lambda_H$. The display namespaces are A, B, C, D and W, X, Y, Z. Within each namespace, menu descriptions and semantic row order remain fixed. Every displayed menu token and the output-schema enumeration follow $\lambda_A$ without subsequent alphabetical sorting. Utilities retain six decimal places, records retain their slots, and each prompt retains the 6,000-character serialization budget. Neither cell codes nor the intervention hypothesis appear in the prompts.

\noindent\textbf{Calls and inference unit.} The matrix contains 48 streams $\times$ two namespaces $\times$ four cells $\times$ two fresh calls, giving 768 requests. A request holds the four family--context prompts from one stream, namespace, cell, and repeat. Different representations of the same history never share a request. Each request starts in a fresh session with GPT-6 Astra, max reasoning effort, and the ordinary service tier. The output schema requires four action tokens. Concurrency is four and the per-attempt timeout is 180 seconds; no client token cap is supplied. The four decisions within a request may be dependent; all remain in their stream cluster. The estimand concerns their marginal action distributions under this four-item~input~interface.

\noindent\textbf{Repeat-corrected contrast.} At a fixed stream, block, and namespace, let $p_c(a)$ be the probability of semantic action $a$ in cell $c$ under the declared response policy. For distinct cells $c,e$, independent calls from stable cell distributions satisfy
\begin{equation}
\begin{aligned}
\mathbb{E}[\mathbf{1}(Y_{cr}\ne Y_{es})]
 &= 1-\sum\nolimits_a p_c(a)p_e(a),\\
\mathbb{E}[\mathbf{1}(Y_{j1}\ne Y_{j2})]
 &= 1-\sum\nolimits_a p_j(a)^2,\quad
\mathbb{E}[T_{\mathrm{gr}}(c,e)]
 = \tfrac12\sum\nolimits_a\bigl(p_c(a)-p_e(a)\bigr)^2.
\end{aligned}
\end{equation}
Finite-sample estimates may be negative and are retained. Representation contrasts use 00--11 and 10--01; assignment contrasts use 00--10 and 11--01. For each stream, the primary contrast averages assignment minus representation distance equally over the two namespaces and four blocks. Its within-cell correction terms cancel algebraically. Streams receive equal weight in the final mean. Percentile bootstrap intervals use 10,000 resamples, drawing 16 streams with replacement within each map stratum. The primary interval is the prespecified inferential result; namespace, map-stratum, and individual-distance intervals are descriptive. No equivalence margin is selected, so an interval covering zero does not establish invariance.

\noindent\textbf{Validation and missingness.} One fresh format-only retry is allowed after a schema-invalid final response. Both attempts remain in the record, and the first valid response under this policy enters the analysis. Valid scientific outputs, tool attempts, and transport/model failures are not retried by the experiment scheduler. Provider-internal reconnect notifications are retained separately. Requests require observed model, effort, and tier metadata matching the frozen settings. The full primary matrix requires all 48 complete streams. Missing decisions are never imputed. A separate descriptive sensitivity excludes every stream containing a format retry and carries no confirmatory interval. The artifact retains earlier model-access failures and constant-response connectivity checks separately; none contributes a scientific observation.

\noindent\textbf{Observed results.} The full matrix contains 768 valid first-attempt requests and 3,072 decoded block decisions, with no missing requests, format retries, or tool calls. The recorded runtime settings match GPT-6 Astra, max reasoning effort, and the default service tier throughout. Captured stderr includes two internal stream-retry warnings, without additional client attempts; all diagnostics are retained. Reported usage totals 15,278,650 input tokens (including 8,742,144 cached tokens) and 55,789 output tokens (including 37,999 reasoning tokens). Each of the 48 stream-level primary contrasts equals 1.000. Both namespaces and all three map strata have the same contrast. The no-retry sensitivity therefore uses all 48 streams and also equals 1.000. The primary bootstrap interval has coincident endpoints because all resampled stream values agree; it quantifies variation in this observed sample under the frozen resampling scheme.

\begin{table}[ht]
\vspace{-4mm}
\centering\small
\caption{Primary assignment-minus-renaming contrast. The percentile interval uses 10,000 complete-stream resamples within the three fixed 16-stream map strata. Every observed stream has the same contrast, producing the degenerate conditional bootstrap interval.}
\label{tab:global-renaming-ci}
\begin{tabular}{lrrr}
\toprule
Contrast & Estimate & Lower & Upper \\
\midrule
Pooled primary & 1.000 & 1.000 & 1.000 \\
\bottomrule
\end{tabular}
\vspace{-2mm}
\end{table}

\begin{table}[ht]
\vspace{-4mm}
\centering\small
\caption{Raw cross-call and within-cell disagreement for the joint history/menu control. Cross-call values average all four comparisons of two independent repeats; within-cell values compare the two repeats. Each entry averages 192 tasks in 48 streams.}
\label{tab:global-renaming-raw}
\begin{tabular}{lrr}
\toprule
Comparison & A--D & W--Z \\
\midrule
Consistent renaming (00--11) & 0.000 & 0.000 \\
Reassigned renaming (10--01) & 0.000 & 0.000 \\
Assignment (00--10) & 1.000 & 1.000 \\
Assignment (11--01) & 1.000 & 1.000 \\
\midrule
Within cell 00 & 0.000 & 0.000 \\
Within cell 01 & 0.000 & 0.000 \\
Within cell 10 & 0.000 & 0.000 \\
Within cell 11 & 0.000 & 0.000 \\
\bottomrule
\end{tabular}
\vspace{-2mm}
\end{table}

\noindent\textbf{Historical-mean reference.} A post hoc comparison asks whether a simple history statistic reproduces the observed choices. We round each displayed utility to six decimal places, compute the mean for each observed key, and select the unique maximum. Unobserved keys are excluded; no mean ties occur at tolerance $10^{-12}$. The prediction follows the history-label map and is decoded through the inverse menu-label map, exactly as for the model output. It agrees with all 3,072 recorded outputs, including every cell, namespace, and repeat. These outputs reuse 192 histories in 48 streams; they are not independent histories. Of the histories, 144 contain two observed keys and 48 contain one. The comparison explains the observed four-cell pattern with a simple behavioral reference. It does not distinguish this rule from other rules that agree on these histories. The supplement supplies the offline comparison script.

\begin{table}[ht]
\vspace{-4mm}
\centering\small
\caption{Utility and oracle readouts for the joint history/menu control. Each namespace uses the same 192 frozen future tasks. $\Delta U$ and $\Delta O$ subtract cell 00; the evaluator and its tie rule are fixed.}
\label{tab:global-renaming-utility}
\begin{tabular*}{\linewidth}{@{\extracolsep{\fill}}lrrrrrrrr@{}}
\toprule
 & \multicolumn{4}{c}{A--D} & \multicolumn{4}{c}{W--Z} \\
\cmidrule(lr){2-5}\cmidrule(lr){6-9}
Cell & $U$ & $O$ & $\Delta U$ & $\Delta O$ & $U$ & $O$ & $\Delta U$ & $\Delta O$ \\
\midrule
00 & 0.669 & 0.219 & 0.000 & 0.000 & 0.669 & 0.219 & 0.000 & 0.000 \\
01 & 0.664 & 0.281 & $-0.005$ & 0.063 & 0.664 & 0.281 & $-0.005$ & 0.063 \\
10 & 0.664 & 0.281 & $-0.005$ & 0.063 & 0.664 & 0.281 & $-0.005$ & 0.063 \\
11 & 0.669 & 0.219 & 0.000 & 0.000 & 0.669 & 0.219 & 0.000 & 0.000 \\
\bottomrule
\end{tabular*}
\vspace{-2mm}
\end{table}

Consistent renaming preserves the observed semantic actions and consequently their evaluator scores. Assignment changes have mean utility shift $-0.005$ and oracle-agreement shift 0.063 under both namespaces. The two consequence measures weight changed choices differently: utility retains score magnitudes, whereas oracle agreement counts maximizers. Their opposite signs illustrate why a large action response does not determine its utility consequence. These secondary values use the original frozen future evaluator and are distinct from the primary assignment-minus-renaming contrast.

\section{Additional composite calibration}
\label{app:composite-main-details}

On the tested queries, the order-sensitive construction responds to pair transport but not to value reassignment. Among the tested composite families, order-dependent writers respond to pair transport, whereas key--value writers retain pair invariance (Table~\ref{tab:content-composite}). Their labels come from the specified update laws, independently of the observed action profiles.

\begin{table}[ht]
\vspace{-4mm}
\centering
\begin{minipage}[c]{0.56\textwidth}
\small
\begin{tabular*}{\linewidth}{@{\extracolsep{\fill}}lrrrr@{}}
\toprule
\textbf{Family} & $n_{\rm op}$ & $A_{\rm rekey}$ & $D_A^{\rm value}$ & $D_A^{\rm pair}$ \\
\midrule
Value--order & 4 & 0.000 & 0.162 & 0.162 \\
Key--value & 2 & 0.667 & 0.021 & 0.000 \\
Key--order & 2 & 0.665 & 0.160 & 0.142 \\
\bottomrule
\end{tabular*}
\end{minipage}\hfill
\begin{minipage}[c]{0.41\textwidth}
\setlength{\abovecaptionskip}{0pt}
\caption{Held-out composite families, excluded from construction of the canonical-family rule. Responses follow Table~\ref{tab:content-calibration}; $n_{\rm op}$ counts writers. Classification correctness was not evaluated.}
\label{tab:content-composite}
\end{minipage}
\vspace{-2mm}
\end{table}

\section{Measured decision consequences}
\label{sec:record-controls}

Pair transport and key-slot reassignment produce similar action responses across folds, while their utility and oracle point estimates have opposite signs (Table~\ref{tab:content-measured}). The Fold II utility intervals include zero for both interventions, so the positive point estimates do not establish an improvement. Value reassignment produces little or no response. Scoring each replayed action on the same held-out task reveals why sensitivity alone cannot indicate whether the stored arrangement is useful: a switch can improve or worsen the decision. Action disagreement weights all switches equally; utility retains the direction and magnitude of their consequences. A post hoc decomposition shows offsetting gains and losses alongside exact utility ties, so small net changes cannot be attributed uniformly to equivalent actions (Appendix~\ref{app:utility-resolution}).

\begin{table}[ht]
\vspace{-4mm}
\centering
\begin{minipage}[c]{0.53\textwidth}
\small
\setlength{\tabcolsep}{2.5pt}
\begin{tabular*}{\linewidth}{@{\extracolsep{\fill}}lrrrrrr@{}}
\toprule
& \multicolumn{3}{c}{\textbf{Fold I}} & \multicolumn{3}{c}{\textbf{Fold II}} \\
\cmidrule(lr){2-4}\cmidrule(l){5-7}
\textbf{Cell} & \textbf{Action} & \textbf{$\Delta U$} & \textbf{$\Delta O$} & \textbf{Action} & \textbf{$\Delta U$} & \textbf{$\Delta O$} \\
\midrule
Rekey & 0.753 & $-0.602$ & $-0.082$ & 0.620 & 0.870 & 0.027 \\
Key-slot & 0.243 & $-0.476$ & $-0.009$ & 0.240 & 0.285 & 0.003 \\
Value & 0.000 & 0.000 & 0.000 & 0.003 & 0.012 & 0.000 \\
Pair & 0.306 & $-0.786$ & $-0.027$ & 0.257 & 0.105 & 0.008 \\
\bottomrule
\end{tabular*}
\end{minipage}\hfill
\begin{minipage}[c]{0.44\textwidth}
\setlength{\abovecaptionskip}{0pt}
\caption{Measured-outcome replay. Fold I trains on streams 0--23 and evaluates on 24--47; Fold II reverses this split (24 evaluation streams each). Action is disagreement; $\Delta U$ and $\Delta O$ subtract aligned values. Normalized utility units are $10^{-3}$; action and oracle entries are proportions. Intervals: Table~\ref{tab:public-persistent-agent-folds}.}
\label{tab:content-measured}
\end{minipage}
\vspace{-2mm}
\end{table}

The synthetic matched-controller comparison reaches the same distinction through a different control. Outcome-Calibrated Recursive Research (OCRR) and its time-preserving placebo both change most decisions under replay, yet their identity-minus-replay utility gaps have opposite signs (Appendix~\ref{app:controller-details}). The utility contrast reports the direction of the decision-quality change on the evaluated tasks. On measured outcomes, UCB1 preserves its decisions under pair transport, as expected from its count--sum update. Because UCB1 and the recurrent writer follow different adaptation trajectories, this reference tests preservation within each trajectory.

\section{Utility resolution and action switches}
\label{app:utility-resolution}

We decompose the utility changes on the measured-outcome replay support (Table~\ref{tab:content-measured}) and the original Astra history/menu support (Table~\ref{tab:global-renaming-utility}). This post hoc analysis reports signed changes, absolute changes, and the consequences conditional on an action switch. The utility gaps between the best and second-best actions average 15.253 and 12.238, respectively, over 48 stream means in $10^{-3}$ utility units. Their stream-mean medians in the same units are 14.908 and 9.725. The delivered analysis also reports empirical gap distributions and near-optimal-choice rates at every fixed regret threshold in $\{0,0.001,0.005,0.010,0.020,0.050\}$.

\begin{table}[ht]
\vspace{-4mm}
\centering
\small
\caption{Retrospective utility decomposition. Utility columns use $10^{-3}$ units; $\Delta U$ subtracts aligned utility. The first two utility columns average within each stream, then equally over 48 streams. The conditional absolute change and sign proportions first average over switched pairs within each stream, then over the $N_S$ streams containing a switch. Measured rows pool both evaluation folds. The Astra row compares cells 10 and 00 with outcome values present.}
\setlength{\tabcolsep}{3pt}
\begin{tabular*}{\linewidth}{@{\extracolsep{\fill}}lrrrrrrr@{}}
\toprule
Condition & $N_S$ & $\Delta U$ & $|\Delta U|$ & $|\Delta U|\mid S$ & Harm & Tie & Gain \\
\midrule
Measured: rekey & 48 & 0.134 & 15.088 & 22.115 & 0.376 & 0.250 & 0.375 \\
Measured: key-slot & 45 & $-0.096$ & 6.100 & 25.812 & 0.433 & 0.152 & 0.416 \\
Measured: pair & 47 & $-0.341$ & 7.227 & 25.784 & 0.455 & 0.127 & 0.418 \\
Astra: reassignment & 48 & $-4.722$ & 21.477 & 21.477 & 0.375 & 0.260 & 0.365 \\
\bottomrule
\end{tabular*}
\label{tab:utility-resolution-compact}
\vspace{-2mm}
\end{table}

Absolute utility changes substantially exceed the corresponding net changes (Table~\ref{tab:utility-resolution-compact}). Improvements and harms offset each other; a separate fraction of switches preserves utility exactly. Consequently, a small net change does not imply that all changed actions are near-equivalent. Exact ties also distinguish optimal-set membership from agreement with a single tie-broken oracle label. For example, the Astra reassignment cell selects an optimal action on 0.375 of decisions, while its agreement with the archived oracle label is 0.281. The Astra analysis uses 768 archived requests and 3,072 decoded block actions. For the measured recurrent writer, the archived result lacks a per-task action ledger. We reconstruct its actions deterministically from the frozen replay implementation and reconcile the action and utility summaries with both stored folds. That aggregate check does not establish itemwise equality with the original execution. Its regret and switch decomposition therefore condition on the reconstructed trace; the best-versus-second-best gaps are read directly from the recorded counterfactual utilities.

\section{Archival black-box comparisons}
\label{app:archival-blackbox-results}

This appendix separates the fresh joint and memory evidence from archival black-box comparisons. The table summarizes the verification scope for each source; the following paragraphs give the reconstruction and reference-rule details.

\begin{table}[ht]
\vspace{-4mm}
\centering
\small
\caption{Verification scope of the joint, memory, and archival interface evidence. New runs check server-accepted inputs, output schemas, and runtime metadata. The joint matrix includes replacements for 14 capacity failures; original and replacement attempts are retained. Archival reconstruction has the scope stated in Appendix~\ref{app:textual-time-audit}.}
\setlength{\tabcolsep}{3pt}
\begin{tabular*}{\linewidth}{@{\extracolsep{\fill}}llll@{}}
\toprule
Study & Input verification & Repeat control & Future outcomes \\
\midrule
Joint ledger & All selected calls & Within condition & Archived tasks \\
Memory pipeline & All calls & Within condition & New task pools \\
Sol/Arm archive & Aligned, value, pair & Unavailable & Archived tasks \\
Other GPT-5.6 archives & Limited reconstruction & Unavailable & Archived tasks \\
\bottomrule
\end{tabular*}
\label{tab:verification-scope}
\vspace{-2mm}
\end{table}

The black-box comparison applies the record interventions where only textual memory and output actions are accessible. Each decision block is serialized into a fixed 6,000-character budget; the coordinator batches up to 32 blocks per request and extracts one action per block. Each request starts without preceding conversational turns; blocks within a request may remain dependent. Future requests stay within a stream, and textual intervals condition on the recorded adaptation histories (Appendix~\ref{app:batch-stream-audit}). Within each deployment, readable and opaque keys share the adaptation trace, future tasks, and intervention maps.

In the recorded outputs, rekeying produces the largest change, value reassignment an intermediate response, and pair transport a much smaller response across the tested deployments and namespaces (Table~\ref{tab:content-interfaces}). Value reassignment changes decisions with the key sequence fixed. For these readouts, pair transport moves intact records across fixed task slots. These cross-configuration responses are descriptive: matched fresh repeats are unavailable, and input reconstruction verifies the Sol/Arm aligned, value, and pair cells only; the remaining cells retain archival labels (Appendix~\ref{app:textual-time-audit}).

Several simple history rules reproduce most aligned Sol/Arm choices. After value reassignment, the historical-mean rule retains higher agreement than sum, count, latest-record, and best-utility rules (Table~\ref{tab:sol-reference-main}). This post hoc comparison separates the mean reference from competing predictors that largely agree with the original choices. Keeping the five rules fixed, a retrospective check on three additional trace sets retains the mean rule's advantage under value reassignment. These sets share the original future-task support. On the shared trace, responses concentrate where the mean-utility winner changes (Appendix~\ref{app:llm-winner-strata}).

\begin{table}[ht]
\vspace{-4mm}
\centering
\begin{minipage}[c]{0.65\textwidth}
\small
\setlength{\tabcolsep}{2pt}
\begin{tabular*}{\linewidth}{@{\extracolsep{\fill}}lrrrrrrrr@{}}
\toprule
& \multicolumn{2}{c}{Shared} & \multicolumn{2}{c}{Trace 1} & \multicolumn{2}{c}{Trace 2} & \multicolumn{2}{c}{Trace 3} \\
\cmidrule(lr){2-3}\cmidrule(lr){4-5}\cmidrule(lr){6-7}\cmidrule(lr){8-9}
Rule & A & V & A & V & A & V & A & V \\
\midrule
Mean & 0.991 & 0.952 & 0.998 & 0.957 & 1.000 & 0.939 & 0.990 & 0.984 \\
Sum & 0.983 & 0.658 & 0.976 & 0.751 & 0.984 & 0.670 & 0.974 & 0.568 \\
Count & 0.983 & 0.658 & 0.976 & 0.751 & 0.984 & 0.670 & 0.974 & 0.557 \\
Latest & 0.988 & 0.663 & 0.997 & 0.761 & 0.990 & 0.665 & 0.969 & 0.542 \\
Best & 0.972 & 0.699 & 0.965 & 0.807 & 0.974 & 0.740 & 0.953 & 0.641 \\
\bottomrule
\end{tabular*}
\end{minipage}\hfill
\begin{minipage}[c]{0.32\textwidth}
\setlength{\abovecaptionskip}{0pt}
\caption{Post hoc Sol/Arm reference agreement. A and V denote aligned and value conditions. Shared uses the original histories; traces 1--3 use independently ordered histories on the same future tasks. Each estimate averages 48 streams. Intervals: Appendix~\ref{app:llm-winner-strata}.}
\label{tab:sol-reference-main}
\end{minipage}
\vspace{-2mm}
\end{table}

\noindent\textbf{Separating assignment from displayed names.} The four-cell Astra control compares rekey responses with consistent renaming at the fixed 12th future task in each block (Table~\ref{tab:global-renaming}). Consistent renaming preserves every observed decoded choice in both namespaces, including under the reassigned history, whereas changing the assignment changes every observed choice under this four-item interface. Fresh calls within each cell agree, so the observed contrast is not accompanied by within-cell decoding disagreement. The same pattern holds across the prespecified map strata. A post hoc reference that selects the observed key with the highest mean utility matches every recorded action after the corresponding label mapping. A key-count selector with a unique winner can also produce this four-cell pattern without using utility values. This control establishes naming consistency; value sensitivity is tested separately on the Sol interface. Inference resamples complete streams, retaining dependence among the four decisions returned in each request. Appendix~\ref{app:global-renaming} reports the conditional bootstrap interval and utility readouts.

\begin{table}[ht]
\vspace{-4mm}
\centering
\begin{minipage}[c]{0.54\textwidth}
\small
\setlength{\tabcolsep}{3pt}
\begin{tabular*}{\linewidth}{@{\extracolsep{\fill}}lrrr@{}}
\toprule
Comparison & A--D & W--Z & Pooled \\
\midrule
Consistent renaming (00--11) & 0.000 & 0.000 & 0.000 \\
Reassigned renaming (10--01) & 0.000 & 0.000 & 0.000 \\
Assignment (00--10) & 1.000 & 1.000 & 1.000 \\
Assignment (11--01) & 1.000 & 1.000 & 1.000 \\
\midrule
Assignment minus renaming & 1.000 & 1.000 & 1.000 \\
\bottomrule
\end{tabular*}
\end{minipage}\hfill
\begin{minipage}[c]{0.43\textwidth}
\setlength{\abovecaptionskip}{0pt}
\caption{Joint history/menu control with GPT-6 Astra. Comparisons report $T_{\mathrm{gr}}$; the last row gives mean assignment minus mean renaming. Within each namespace, responses are averaged over four tasks in each of 48 streams, using two fresh calls per cell. Namespaces have equal weight in the pooled values.}
\label{tab:global-renaming}
\end{minipage}
\vspace{-2mm}
\end{table}

Independently ordered adaptation traces retain the rekey--value--pair ordering under the Sol readout in both namespaces (Table~\ref{tab:content-tracebank}). The magnitude of the value response varies with the resulting history. The Qwen representation control instead shows markedly lower rekey responses with aliases and opaque keys than with action names (Appendix~\ref{app:llm-opaque-control}), making namespace behavior interface-specific. Thus the ordering extends across these separately generated histories, while the strength of outcome-value sensitivity remains history-dependent. Logged executions with identical inputs show similar aggregate value responses despite disagreement on some individual outputs (Appendix~\ref{app:llm-run-repeat}).

\section{Additional experiments addressing diagnostic validity}
\label{app:new-diagnostic-experiments}

\subsection{Finite-memory repair selection on held-out public-data tasks}
\label{app:method-memory-repair}

We pre-specified a selection experiment testing whether replay diagnostics choose a finite-memory implementation for future public-data tasks. It uses the Breast Cancer Wisconsin (Diagnostic) and Wine Quality datasets, independent and shifted contexts, and a frozen ridge-logistic executor with utility $\exp(-\mathrm{logloss})$. Development has 16 streams and final test has 48, each with four family--context blocks. Each block has 32 historical tasks and one future query, with every action occurring eight times. The scientific unit is the stream; blocks and repeats are paired. Test query outcomes are neither computed nor revealed until all test decisions, action identifiers, and output schemas have been checked. The three public descriptors are the candidate pool mean and standard deviation of its first standardized feature and the mean of its second feature. Test histories are executed only after the development selection is frozen.

Each event stores descriptors, selected action, realized utility, logical event time, and an identifier. Sequential insertion gives each candidate at most 16 storage slots, while retrieval exposes only the nearest eight and evicted records cannot be recovered from a sidecar. The common internal schema stores action, count, mean utility, three descriptor means, event time, bin, and priority; bin and priority are hidden. Each displayed slot has a fixed 144-character budget with \texttt{EMPTY} padding. The candidates are first-in, first-out (FIFO) retention of the last 16 arrivals, retention of the 16 latest logical-time events, an identifier-priority reservoir of 16 events, and online action--descriptor-bin summaries with at most 16 slots. The aggregate uses four fixed first-descriptor bins at $-0.4$, $0$, and $0.4$ and stores counts and means. All candidates share storage and model-call budgets while using different summaries. Equal slots and character budgets do not imply equal tokens; actual input, output, reasoning, and retry usage are retained.

Retrieval ranks slots by squared Euclidean distance in the three descriptors, breaking ties by logical time, semantic action order, and priority, and exposes the nearest eight. \texttt{ingestion\_order} permutes complete events, including descriptors, actions, utilities, logical times, and identifiers, and changes only their ingestion order. It differs from the main-text pair intervention, which moves $(\text{key},\text{utility})$ pairs while keeping public input $x$ in its original slot. \texttt{value} permutes utilities and \texttt{rekey} cyclically relabels actions with the query menu fixed. Each transformation is fixed within a stream and block, and repeats reuse the same transformed input. The audit selector observes two responses for each of aligned, value, and complete-event-order conditions per candidate; the endpoint-only selector observes six aligned external-utility responses and no perturbation responses or retrieval state. Both receive no-memory and baseline-rekey controls, excluded from selection. Each selector has a logical development budget of 448 requests; shared aligned repeats and controls reduce the unique acquisition to 704. They use the same 64 development query tasks. Endpoint selection chooses the candidate with the highest six-repeat aligned utility. FIFO is retained when the mean baseline corrected ingestion-order response is at most $0.05$. Otherwise, eligibility requires corrected ingestion-order response at most $0.05$, corrected value response at least $0.05$, and development aligned utility at least FIFO utility minus $0.002$. The largest aligned-utility eligible candidate is selected, with frozen-order tie-breaking; FIFO is retained if none qualifies. The 768-request test matrix covers baseline, audit-selected, endpoint-selected, and no-memory conditions under aligned and complete-event-order histories. Co-primary audit--baseline and audit--endpoint contrasts use an equally weighted aligned/stress utility mixture, each with a 20,000-draw stream-paired 97.5\% percentile interval. Both lower endpoints must exceed zero; separate environments are secondary contrasts.

\noindent\textbf{Held-out results and acquisition checks.} The complete matrix contains 704 development and 768 test requests, with four item decisions per request. Every first attempt completed, with no format retry, provider retry notification, missing response, or executor failure. Server-accepted user messages match the frozen prompts byte for byte; final responses pass the schema, semantic-action decoding, and runtime checks. All test decisions were sealed before computing outcomes for the 192 new query tasks. Table~\ref{tab:memory-test-contrasts} reports the frozen comparisons. The audit and endpoint rules select the identical binned-mean implementation, so their test outputs are fresh-call replications of one choice. The first comparison supports higher utility than FIFO on the declared mixture; the required joint criterion for additional audit selection value is not met. The no-memory comparison is descriptive. Intervals condition on the one development selection, the four-item interface, and task generation from the two public datasets.

\begin{table}[ht]
\vspace{-4mm}
\centering
\small
\caption{Audit-selected minus comparator utility on the equal-mixture test, in $10^{-3}$ units. Lower and Upper are individual 97.5\% percentile-bootstrap limits from 20,000 paired draws of 48 streams. The first two comparisons form the prespecified family with a Bonferroni 95\% coverage target; bootstrap coverage is approximate. The no-memory comparison is descriptive.}
\label{tab:memory-test-contrasts}
\begin{tabular}{lrrr}
\toprule
Comparator & Estimate & Lower & Upper \\
\midrule
FIFO & 3.042 & 0.092 & 6.153 \\
Endpoint-selected binned mean & -0.292 & -1.079 & 0.209 \\
No memory & 4.580 & 0.970 & 8.219 \\
\bottomrule
\end{tabular}
\vspace{-2mm}
\end{table}

\begin{table}[ht]
\vspace{-4mm}
\centering
\small
\caption{Changed memory artifacts after complete-event reordering. Each count has denominator 192 test blocks. Stored summaries are compared as a sorted multiset of displayed fields at six-decimal precision; undisplayed state and slot order are excluded. Retrieved text is compared exactly. No-memory retains an unused FIFO sidecar but exposes no retrieved text to the policy.}
\label{tab:memory-test-artifacts}
\begin{tabular}{lrr}
\toprule
Memory & Stored summaries & Retrieved text \\
\midrule
FIFO & 192 & 192 \\
Binned mean (audit) & 0 & 0 \\
Binned mean (endpoint) & 0 & 0 \\
No memory & 192 & 0 \\
\bottomrule
\end{tabular}
\vspace{-2mm}
\end{table}

The selected compressor's retrieved summaries cover all 32 historical observations per block; FIFO retains 16 observations and retrieves eight. Both obey the same 16-slot storage and eight-slot retrieval ceilings. This comparison changes the representation and retained information together, so it does not isolate an aggregation mechanism at equal observation coverage. The two development selectors each receive a logical allowance of 448 requests; shared aligned calls yield 704 unique requests. Their actual total-token counts are 7,597,145 and 7,590,584, respectively. Across development and test, the provider reports 24,603,303 input tokens and 166,411 output tokens, for 24,769,714 total; 130,591 reasoning tokens are included in output rather than added again. The supplement retains usage by condition and attempt. Equal allowances do not imply equal realized token use. External scoring completes 8,192 selected-history arm fits and 1,024 query-arm fits; no scientific failure is retried.

\subsection{Retrospective joint history--menu reference challenge}
\label{app:method-joint-blackbox}

We construct a retrospective black-box challenge from the sealed public-data replication. It tests whether decisions under controlled ledger transformations can be reproduced by simple historical reference rules. The challenge is diagnostic: it does not claim natural adaptation, mechanism identification, prospective validation, or independent future generalization. For each of 48 streams and four family--context blocks, the first 24 adaptation tasks define a forced history. The current item is fixed to future local index 11. Its counterfactual outcome table is available from the sealed public-data artifact for analysis but is withheld from every prompt, so no future outcome enters the model input.

The reference rules are \texttt{mean}, \texttt{sum}, \texttt{count}, \texttt{latest}, and \texttt{best}. Ten fixed history groups correspond to the unordered rule pairs. A deterministic PCG64 search tests block-local action schedules for the required disagreement on six-decimal displayed utilities. Groups with fewer feasible blocks are allocated first, with fixed block-order tie-breaking, until all 192 blocks receive a group. The sum--count disagreement is feasible in 100 of the 192 blocks; its 19 assigned histories are drawn from those feasible blocks. The allocation yields 20 blocks for two groups and 19 for each of the other eight. Comparisons involving count, and the mean--sum comparison, use a permutation of action counts $(7,6,6,5)$; other groups use $(6,6,6,6)$. The group label is retained only as client-side metadata, not as prompt text or an additional request factor. Group-level summaries therefore use only the 5--10 unique streams represented in the frozen history metadata; the expanded request matrix is not an independent sample size. Ties use first appearance in the visible ledger followed by the canonical arm order. Each history is exposed in five forms: chronological full records, a utility-only permutation, complete pair transport using the same permutation, a key-only slot permutation, or keys without utilities. The value and pair forms share the exact utility permutation, including across repeats. Four cells cross identity and a fixed involution for the history labels with identity and the involution for the visible action menu. Semantic aliases and opaque one-character tokens provide two namespaces. Prompts show menu descriptions but no history relabeling, source-action dictionary, reference rule, or intended winner.

The matrix contains $48\times2\times5\times4\times2=3{,}840$ requests. Each request has four numbered block items; each cell is evaluated with two fresh requests, and repeats are separate ephemeral threads. Stream is the primary cluster; items and repeats are paired. The frozen transport uses GPT-6 Astra with max reasoning and the default service tier. The output schema requires one displayed menu token per item; the client maps each returned token to its semantic action for analysis. Equal request counts and prompt structure do not imply equal token usage; all actual usage and raw attempts are retained. A schema failure permits at most one format-only retry. The original queue returns 3,826 complete responses and 14 capacity errors. After the queue finishes, an explicit protocol amendment permits replacement calls for those failed requests using the unchanged model, prompts, schema, and reasoning setting. All 14 replacements succeed on their first attempt; successful original requests are not repeated. The resulting matrix contains 3,840 successful requests from 3,854 client attempts. Original failures, replacement calls, their available usage, and the selected source of every response are retained separately. The amendment precedes inspection of aggregate joint effects and is not described as part of the original no-retry protocol. Subsequent utility and disagreement analyses condition on the realized forced histories and use the stream-level metadata, including the small 5--10-stream rule-group support.

All four items in a request share the assigned condition. Responses therefore describe this batched interface; the protocol does not isolate the effect of changing one item while holding the other three histories fixed.

\subsection{Same-architecture shortcut-resistant learned control}
\label{app:method-learned-control}

We train a new synthetic recurrent writer with the same 16-dimensional tanh architecture used by the learned pair in Table~\ref{tab:endpoint-separation}:
\begin{equation}
h_t=\tanh(W h_{t-1}+p_{\ell_t}+(u_t-1/2)v_{\ell_t}),\quad
s_a=q_a^\mathsf{T}h_T .
\label{eq:learned-shortcut-recurrence}
\end{equation}
The initial state is zero and the model has no action or readout bias. Each episode contains 5--10 complete four-producer blocks, with every producer appearing once per block and utilities drawn independently from $\operatorname{Uniform}(0.43,0.57)$. The supervised target is the producer with the largest sum of centered selected utilities, encoded as $0.95$ for that producer and $0.05$ for the others. Fixed rejection sampling balances the target classes. The train, development, and held-out partitions contain 512, 128, and 96 episodes, respectively; held-out future utility rows are generated once and are never used for fitting. Five fixed parameter seeds are each run once under both conditions, giving ten retained fits. The ordinary condition preserves each sampled block order. The permutation-augmented condition applies a new deterministic arbitrary occupied-slot permutation of complete key--utility records to each training episode at every epoch, preserving the aggregate target. Both conditions use the same full-batch Adam schedule for 4,800 updates per fit. There is no restart selection, early stopping, development selection, seed replacement, outcome-conditioned rerun, model API call, or executor call. Development excess binary cross-entropy (BCE) over the fixed target entropy is a fit-quality audit with threshold $0.080$; it does not select fits. After freezing each fit, we evaluate all nine four-key derangements on held-out episodes. Value transport permutes occupied-slot utilities; \texttt{pair\_slot} moves complete mapped key--utility records with the same realized permutation, whereas \texttt{pair\_block} moves complete four-record blocks. The primary readout is the all-held-out \texttt{pair\_slot} action response, accompanied by aggregate-target accuracy; episode is the statistical unit and maps and future utility rows are paired interventions. Each fit receives a 4,096-draw episode bootstrap interval. The shared-correct subgroup is computed only after the complete held-out analysis; an empty subgroup is omitted and cannot replace the complete result.

\begin{table}[ht]
\vspace{-4mm}
\centering
\small
\caption{All completed learned-control fits. Fit indices 1--5 correspond respectively to model seeds 20260951, 20260952, 20260953, 20260954, and 20260955. Target accuracies use the mapped-aligned reference and the arbitrary complete-pair slot replay, respectively. Pair action is the mapped-aligned versus pair-slot action-change proportion; its Lower and Upper columns are 95\% episode-bootstrap limits (96 held-out episodes; nine maps paired within episode). Excess BCE is binary cross-entropy less the soft-target entropy.}
\label{tab:learned-control-all-fits}
\setlength{\tabcolsep}{4pt}\begin{tabular}{lrrrrrrr}
\toprule
Regimen & Fit & \shortstack{Dev.\\ excess BCE} & \shortstack{Mapped aligned\\ target acc.} & \shortstack{Pair replay\\ target acc.} & \shortstack{Pair\\ action} & Lower & Upper \\
\midrule
Ordinary & 1 & 0.471 & 0.367 & 0.343 & 0.670 & 0.637 & 0.704 \\ Ordinary & 2 & 0.670 & 0.262 & 0.237 & 0.736 & 0.708 & 0.764 \\ Ordinary & 3 & 0.404 & 0.536 & 0.500 & 0.557 & 0.515 & 0.597 \\ Ordinary & 4 & 0.583 & 0.422 & 0.394 & 0.613 & 0.579 & 0.647 \\ Ordinary & 5 & 0.605 & 0.280 & 0.256 & 0.740 & 0.712 & 0.766 \\
\midrule
Permutation aug. & 1 & 0.133 & 0.810 & 0.833 & 0.186 & 0.141 & 0.234 \\ Permutation aug. & 2 & 0.390 & 0.292 & 0.242 & 0.576 & 0.543 & 0.612 \\ Permutation aug. & 3 & 0.390 & 0.247 & 0.240 & 0.375 & 0.340 & 0.409 \\ Permutation aug. & 4 & 0.197 & 0.659 & 0.670 & 0.194 & 0.159 & 0.233 \\ Permutation aug. & 5 & 0.384 & 0.314 & 0.312 & 0.589 & 0.551 & 0.626 \\
\bottomrule
\end{tabular}
\vspace{-2mm}
\end{table}

\begin{table}[ht]
\vspace{-4mm}
\centering
\small
\caption{Permutation-augmented minus ordinary contrasts. The first three rows report the primary pair-slot readouts; the last two report secondary consequences. Intervals use a joint 4,096-draw bootstrap of the same 96 episode IDs across the five fixed seed pairs, with maps averaged within episode. Utility is in $10^{-3}$ units; all other metrics are proportions.}
\label{tab:learned-control-primary-contrasts}
\label{tab:learned-control-secondary-contrasts}
\begin{tabular}{llrrr}
\toprule
Metric & Unit & Estimate & Lower & Upper \\
\midrule
Pair action change & proportion & -0.279 & -0.300 & -0.259 \\
Mapped aligned target acc. & proportion & 0.091 & 0.068 & 0.114 \\
Pair replay target acc. & proportion & 0.114 & 0.093 & 0.135 \\
\midrule
Aligned--replay utility & $10^{-3}$ normalized utility & 0.555 & -0.361 & 1.531 \\
Replay--aligned oracle agreement & proportion & -0.002 & -0.006 & 0.002 \\
\bottomrule
\end{tabular}
\vspace{-2mm}
\end{table}

\noindent\textbf{Frozen fit-quality gate and scope.} The frozen shortcut-resistant learned-response interpretation required development excess BCE $\leq 0.080$ together with a pair-slot action lower limit above $0.050$. Every completed fit exceeds the excess-BCE threshold (ordinary: 0.404--0.670; permutation augmented: 0.133--0.390), so the quality gate is failed for all ten fits and supports 0/5 augmented fits. The negative paired action contrast therefore has descriptive scope for this fixed synthetic interface only; these runs do not establish a target-consistent learned shortcut response.

\section{Sequential experiment-planning agent}
\label{app:sequential-agent}

All trajectories use GPT-6 Astra with max reasoning effort and the ordinary service tier. The prompt and tool schemas are shared across conditions. Each trajectory starts a fresh conversation; only \texttt{run\_experiment} and \texttt{submit\_configuration} are permitted. Short choice rationales are retained but do not enter scoring. The agent selects classification pipelines on 48 new resampled tasks, with 12 tasks in each wine/breast-cancer and i.i.d./shift stratum. Each task has three disjoint pools of 96 observations. Candidate observations supply training data; validation observations provide feedback during exploration; final-test observations are used after all choices are sealed. In the shift condition, candidates are sampled from observations at or below the source median of the first feature, and validation and test observations are sampled without overlap from those above it. Preprocessing parameters are estimated from the candidate pool. The source datasets, binary labels, and four sampling policies follow Appendix~\ref{app:public-data-replication}.

The action menu combines four sampling policies, two preprocessing choices (candidate standardization, with or without clipping at three standard deviations), two classifiers (ridge logistic regression and shrinkage linear discriminant analysis), three regularization levels, and two training-label budgets (12 and 24). This yields 96 configurations. Eight distinct configurations initialize each history with measured validation utilities; their selection and order use fixed seeds independently of the outcomes. Every condition exposes all eight records and the full menu. Public task cards also describe the unlabeled candidate pool. Configuration IDs have equal length and do not encode numerical settings. The regularization values are 0.02, 0.08, and 0.32. Logistic regression starts from zero coefficients and takes 80 gradient steps of size 0.12, penalizing all coefficients except the intercept. Each gradient component is clipped at magnitude five and each coefficient at 25. Linear discriminant analysis shrinks the pooled within-class covariance toward its diagonal, then adds $10^{-6}I$ before pseudoinversion. Class priors use a half-count correction. With a single training class, linear discriminant analysis uses that smoothed prior as a constant predictor; this valid outcome is retained. Scoring clips probabilities to $[10^{-6},1-10^{-6}]$. Value reassignment permutes the eight displayed utilities while retaining the configuration sequence. Pair transport uses the same single-cycle permutation to move complete configuration--utility pairs, thereby producing the same utility sequence as value reassignment. Consistent renaming applies a fixed bijection to configuration IDs in the history, menu, tool arguments, and tool responses; the menu's executable order is preserved. All comparisons decode IDs to canonical configurations. Both repeats of each task--condition cell receive identical initial inputs in fresh threads. The acquisition schedule interleaves all four conditions and both repeats using a fixed shuffle.

\begin{table}[ht]
\vspace{-4mm}
\centering
\small
\setlength{\tabcolsep}{3pt}
\caption{Action responses relative to aligned histories, with descriptive 95\% intervals. Cross averages all four cross-repeat disagreements; Within and Aligned give each condition's repeat disagreement. Each estimate averages the same 48 paired tasks. Later choices follow the preceding experiment feedback.}
\begin{tabular}{llrrrrrr}
\toprule
Choice & Condition & $T_{\rm gr}$ & Lower & Upper & Cross & Within & Aligned \\
\midrule
First & Value reassigned & 0.708 & 0.625 & 0.792 & 0.990 & 0.250 & 0.312 \\
First & Pair transport & 0.031 & -0.031 & 0.104 & 0.292 & 0.208 & 0.312 \\
First & Consistent renaming & -0.042 & -0.099 & 0.026 & 0.323 & 0.417 & 0.312 \\
Second & Value reassigned & 0.542 & 0.438 & 0.646 & 1.000 & 0.438 & 0.479 \\
Second & Pair transport & 0.042 & -0.052 & 0.141 & 0.542 & 0.521 & 0.479 \\
Second & Consistent renaming & -0.052 & -0.120 & 0.010 & 0.490 & 0.604 & 0.479 \\
Third & Value reassigned & 0.516 & 0.417 & 0.615 & 0.984 & 0.354 & 0.583 \\
Third & Pair transport & 0.016 & -0.078 & 0.115 & 0.557 & 0.500 & 0.583 \\
Third & Consistent renaming & -0.057 & -0.146 & 0.036 & 0.578 & 0.688 & 0.583 \\
Submit & Value reassigned & 0.875 & 0.812 & 0.927 & 1.000 & 0.062 & 0.188 \\
Submit & Pair transport & 0.000 & -0.042 & 0.052 & 0.188 & 0.188 & 0.188 \\
Submit & Consistent renaming & -0.021 & -0.057 & 0.016 & 0.219 & 0.292 & 0.188 \\
\bottomrule
\end{tabular}
\label{tab:sequential-responses}
\vspace{-2mm}
\end{table}

Each trajectory has four decision stages. In each of the first three, the agent chooses an untried configuration through \texttt{run\_experiment}. The executor trains the chosen classifier and returns validation utility, $\exp(-\mathrm{log\ loss})$. The verified result enters the same conversation before the next decision. The fourth stage calls \texttt{submit\_configuration} with an initial configuration or a successfully tested new configuration. Test scores are unavailable to either tool. Each trajectory permits three experimental fits and at most 72 training-label accesses, counting repeated accesses. Token usage is recorded separately. A failed scientific fit consumes its experiment slot; format errors allow one correction per trajectory, with both attempts retained.

The initial-incumbent reference selects the highest displayed initial utility. The random-search reference evaluates three seed-selected untried configurations and selects the highest displayed utility among the resulting 11 records. Its canonical trial sequence is identical across conditions. Both references break ties in canonical order. Agent and reference choices are sealed together before the executor computes any final-test utility. The first choice measures the response to the transformed initial history. Steps two and three and the final submission also incorporate the consequences of new experiments and their feedback. We report cross-repeat action disagreement corrected for within-condition variation at each step. The predeclared first-choice diagnostic is the value-versus-aligned response minus the pair-versus-aligned response. It compares the strength of the two intervention responses. Utility contrasts first average the two repeats within each task. Ten thousand paired bootstrap draws resample 12 task seeds within each of the four fixed strata. Intervals for the two primary final-utility contrasts, value-minus-aligned and pair-minus-aligned, use individual 97.5\% bootstrap intervals; other intervals are descriptive 95\% intervals. These intervals are conditional on the two source datasets and four strata. The supplement provides task definitions, frozen prompts, tool schemas, complete public tool trajectories, and analysis code.

\begin{table}[ht]
\vspace{-4mm}
\centering
\small
\caption{Value-minus-pair response gaps, $T_{\rm gr}(\mathrm{value},\mathrm{aligned})-T_{\rm gr}(\mathrm{pair},\mathrm{aligned})$, with descriptive 95\% intervals. This compares two responses to the same aligned reference.}
\setlength{\tabcolsep}{3pt}
\begin{tabular}{lrrr}
\toprule
Choice & Estimate & Lower & Upper \\
\midrule
First & 0.677 & 0.568 & 0.776 \\
Second & 0.500 & 0.391 & 0.609 \\
Third & 0.500 & 0.391 & 0.604 \\
Submit & 0.875 & 0.812 & 0.932 \\
\bottomrule
\end{tabular}
\label{tab:sequential-gaps}
\vspace{-2mm}
\end{table}

\noindent\textbf{Acquisition amendments.} The scientific inputs, model settings, and experimental budget remain those frozen before acquisition. Collector repairs corrected runtime-directory conflicts, handling of the one permitted format correction, and rejection of completed calls because of unrelated background diagnostics. Five completed trajectories were recovered from their verified tool records without new calls. A subsequent host restart interrupted four trajectories before submission; these restarted from their original inputs in fresh conversations. All completed trajectories were retained without rerunning them. The supplement preserves the original failures, partial attempts, recovery maps, and amendments. Three additional fits occurred during baseline preparation before a variable error was corrected; their outputs were not persisted. The initial failed agent attempt used one validation fit, and the host-interrupted attempts used three. These seven fits are acquisition overhead in addition to the completed experimental matrix. Reported token usage from incomplete attempts is retained; usage unavailable for those attempts remains unavailable. The completed matrix has 1,536 accepted decisions and 383 format corrections, for 1,919 model turns; the token totals include all of them. The initial tool schemas specify a 300-character rationale limit. All corrections address this length limit and retain the rejected call's tool and configuration choice. The entire final choice matrix is sealed before test scoring. Per-condition call and token accounting remains in the run manifests; it is omitted here because it does not define an estimand or change a reported result.

\begin{table}[ht]
\vspace{-4mm}
\centering
\small
\caption{Final-test utility with descriptive 95\% intervals. Agent estimates first average the two repeats within each task. References use their frozen choices on the same test pools.}
\setlength{\tabcolsep}{3pt}
\begin{tabular}{llrrr}
\toprule
Policy & Condition & Estimate & Lower & Upper \\
\midrule
Agent & Aligned & 0.618 & 0.598 & 0.639 \\
Agent & Value reassigned & 0.359 & 0.303 & 0.415 \\
Agent & Pair transport & 0.617 & 0.597 & 0.639 \\
Agent & Consistent renaming & 0.619 & 0.599 & 0.640 \\
Random search & Aligned & 0.610 & 0.588 & 0.632 \\
Random search & Value reassigned & 0.443 & 0.379 & 0.506 \\
Random search & Pair transport & 0.610 & 0.588 & 0.632 \\
Random search & Consistent renaming & 0.610 & 0.588 & 0.632 \\
Initial incumbent & Aligned & 0.594 & 0.573 & 0.614 \\
Initial incumbent & Value reassigned & 0.342 & 0.287 & 0.399 \\
Initial incumbent & Pair transport & 0.594 & 0.573 & 0.614 \\
Initial incumbent & Consistent renaming & 0.594 & 0.573 & 0.614 \\
\bottomrule
\end{tabular}
\label{tab:sequential-utility}
\vspace{-2mm}
\end{table}

\begin{table}[ht]
\vspace{-4mm}
\centering
\small
\caption{Paired final-utility contrasts. The two primary contrasts use individual 97.5\% bootstrap intervals, a Bonferroni adjustment for two comparisons; the renaming interval is descriptive.}
\setlength{\tabcolsep}{3pt}
\begin{tabular}{lrrrr}
\toprule
Contrast & Estimate & Lower & Upper & Level \\
\midrule
Pair transport minus aligned & 0.000 & -0.002 & 0.001 & 0.975 \\
Value reassigned minus aligned & -0.259 & -0.330 & -0.190 & 0.975 \\
Consistent renaming minus aligned & 0.001 & -0.001 & 0.003 & 0.950 \\
\bottomrule
\end{tabular}
\label{tab:sequential-utility-contrasts}
\vspace{-2mm}
\end{table}

\begin{table}[ht]
\vspace{-4mm}
\centering
\small
\caption{Reference-minus-agent utility under the same history condition, with descriptive 95\% intervals. Random search has the same three-experiment allowance; the initial incumbent makes no new experimental calls. Both references use no model calls.}
\setlength{\tabcolsep}{3pt}
\begin{tabular}{llrrr}
\toprule
Reference & Condition & Estimate & Lower & Upper \\
\midrule
Random search & Aligned & -0.008 & -0.028 & 0.014 \\
Random search & Value reassigned & 0.084 & 0.036 & 0.137 \\
Random search & Pair transport & -0.008 & -0.028 & 0.015 \\
Random search & Consistent renaming & -0.009 & -0.029 & 0.013 \\
Initial incumbent & Aligned & -0.024 & -0.039 & -0.011 \\
Initial incumbent & Value reassigned & -0.017 & -0.035 & -0.003 \\
Initial incumbent & Pair transport & -0.024 & -0.039 & -0.011 \\
Initial incumbent & Consistent renaming & -0.025 & -0.040 & -0.012 \\
\bottomrule
\end{tabular}
\label{tab:sequential-references}
\vspace{-2mm}
\end{table}

\begin{table}[ht]
\vspace{-4mm}
\centering
\small
\caption{Reference-minus-agent utility under the same history condition, with descriptive 95\% intervals. Random search has the same three-experiment allowance; the initial incumbent makes no new experimental calls. Both references use no model calls.}
\setlength{\tabcolsep}{3pt}
\begin{tabular}{lrrrr}
\toprule
Condition & Distinct & Gap & Lower & Upper \\
\midrule
Aligned & 3.708 & 0.027 & 0.009 & 0.046 \\
Value reassigned & 3.417 & 0.264 & 0.208 & 0.321 \\
Pair transport & 3.583 & 0.027 & 0.009 & 0.046 \\
Consistent renaming & 3.896 & 0.027 & 0.008 & 0.045 \\
\bottomrule
\end{tabular}
\label{tab:sequential-exploration}
\vspace{-2mm}
\end{table}

\section{Code-debugging tool-agent experiment}
\label{app:code-debugging-agent}

The code-debugging study is a complete replay audit on a non-classification task substrate. It uses 24 deterministic bug-fixing tasks from eight function families, with three task instances per family: offset, clamping, ordered deduplication, top-$k$ selection, rotation, sign handling, suffix mean, and dictionary lookup. Every task exposes twelve candidate Python implementations and an initial ledger of four measured patch outcomes. Candidate identifiers are randomized per task and the prompt shows source code without a correctness label, so the model cannot use a name or a supplied rank as a correctness shortcut. The visible test inputs and expected outputs are fixed before acquisition; the hidden test inputs are sealed until the complete choice matrix is recorded.

The tool interface has two calls. \texttt{run\_tests} executes a selected candidate on the public tests and returns only the number of passing tests, and \texttt{submit\_patch} records the final candidate without revealing a hidden result. Each trajectory permits three sequential test calls followed by one submission. The logged key is the patch identifier, the observed outcome is the public pass fraction, the placement relation is the test-call position in the initial ledger, and the external evaluator is the hidden test suite. The same candidate menu, executor, prompt schema, model, and tool budget are used in every replay condition.

The frozen matrix has four conditions: aligned, outcome reassignment with utility permutation $(1,0,3,2)$, complete-pair transport with position permutation $(2,3,0,1)$, and consistent renaming, which maps each history and menu identifier $P_i$ to $R_{(i+1)\bmod 12}$. Each task--condition cell has two fresh trajectories, giving $24\times4\times2=192$ trajectories and three tool calls plus one submission per trajectory, or 768 tool events. GPT-6 Astra uses max reasoning and the default service tier. One capacity failure occurred before hidden scoring; the frozen capacity-replacement amendment allowed one replacement with the same inputs and settings, and the replacement completed. All 192 trajectories then passed completeness checks, with no missing trajectory, protocol violation, or format correction. Hidden scores were released only after the choice seal.

Table~\ref{tab:code-debugging-intervals} reports task-bootstrap intervals from 10,000 draws. The first and final readouts compare each replay condition with aligned on the same task and repeat; hidden utility is the held-out pass fraction. Outcome reassignment changes the first test choice on 60.4\% of task-level paired comparisons and the final submission on 16.7\%, with a mean hidden-utility loss of 0.125. Pair transport and consistent renaming have small first-choice responses and zero final or hidden-utility response on this support. These results are a domain transfer of the audit protocol, not evidence about open-ended repository repair or unrestricted code generation.

\begin{table}[ht]
\vspace{-4mm}
\centering
\small
\setlength{\tabcolsep}{3pt}
\caption{Code-debugging replay readouts relative to aligned histories. Estimates are task-level means over 24 tasks; intervals are descriptive 95\% bootstrap limits from 10,000 resamples of the 24 task values. First and Final are action-disagreement proportions; Hidden utility is the held-out test-score difference.}
\label{tab:code-debugging-intervals}
\begin{tabular}{llrrr}
\toprule
Readout & Condition & Estimate & Lower & Upper \\
\midrule
First action & Value reassigned & 0.604 & 0.438 & 0.771 \\
First action & Pair transport & 0.042 & 0.000 & 0.125 \\
First action & Consistent renaming & 0.083 & 0.021 & 0.167 \\
Final action & Value reassigned & 0.167 & 0.042 & 0.312 \\
Final action & Pair transport & 0.000 & 0.000 & 0.000 \\
Final action & Consistent renaming & 0.000 & 0.000 & 0.000 \\
Hidden utility & Value reassigned & -0.125 & -0.292 & 0.000 \\
Hidden utility & Pair transport & 0.000 & 0.000 & 0.000 \\
Hidden utility & Consistent renaming & 0.000 & 0.000 & 0.000 \\
\bottomrule
\end{tabular}
\vspace{-2mm}
\end{table}

The compact public export retains the frozen manifest, task definitions, trajectory plan, choice seal, hidden-score release record, replacement amendment, all accepted tool events, and the analysis script. It also records the acquisition transport hash and the source hash used for model calls. The raw conversation logs remain in the project evidence directory; the export is sufficient to recompute the table without making additional model calls.

\section{Joint black-box results and recovery provenance}
\label{app:joint-blackbox-results}

The joint comparison uses all 3,840 successful requests and 15,360 decoded item decisions. Input reconstruction is checked against server-accepted user text, with frozen output schemas, menu decoding, model, reasoning effort, and service tier. A saved prompt file alone is not treated as evidence that the server accepted that input. The original queue contains 14 capacity failures; an author-authorized amendment replaces only these requests after the original queue ends. All replacements succeed on their first call. The original failures and the selected replacement sources remain in the supplement, and no successful original response is replaced. The resulting 3,854 client attempts include 14 failed attempts without reported usage. The 3,840 successful attempts report 69,055,719 input tokens and 716,772 output tokens, including 624,752 reasoning tokens within the output count. These are observed usage totals; usage for failed calls is unavailable. Request counts, format retries, provider notifications, and usage availability are audited separately.

\begin{figure}[ht]
\centering
\includegraphics[width=\linewidth]{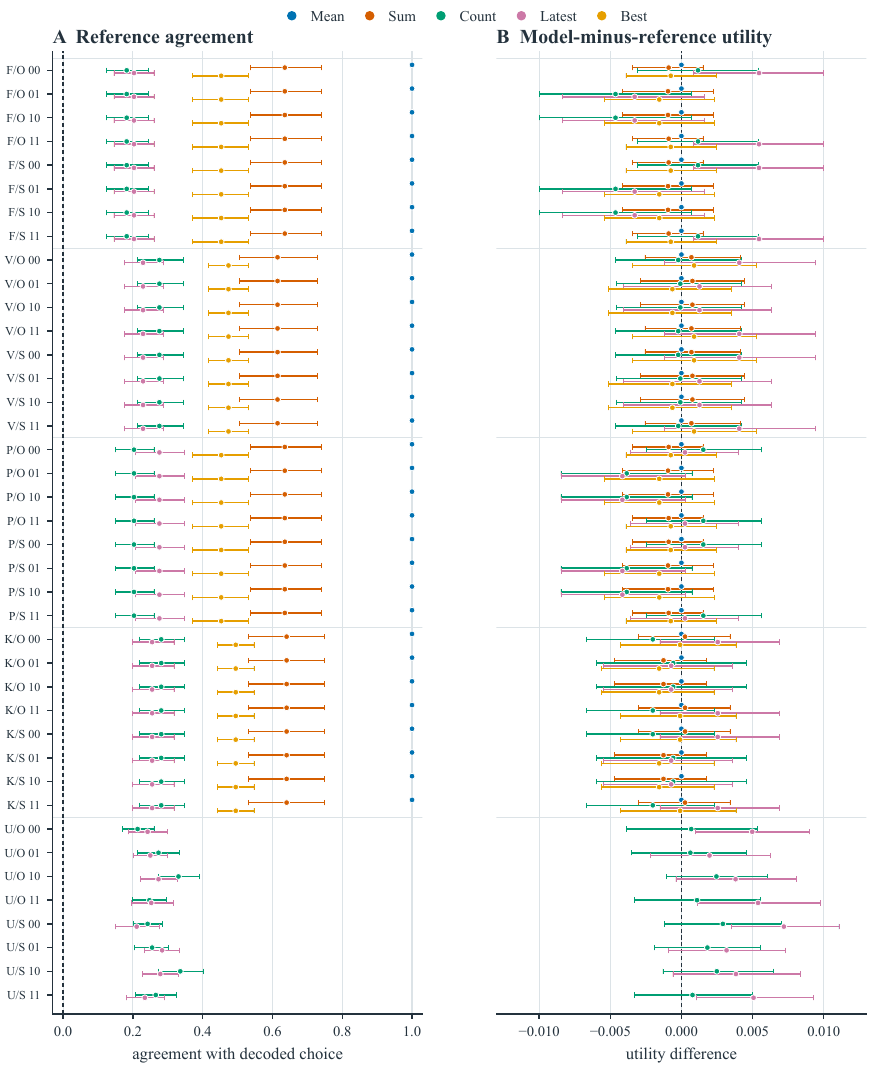}
\caption{\textbf{Reference-rule agreement and utility.} Each row is one form/namespace/cell combination: F, V, P, K, and U denote Full, Value, Pair, Key-slot, and Utility-masked; O and S denote OPAQUE and SEMANTIC. Colored points are the five prespecified reference rules and horizontal bars are 95\% stream-bootstrap intervals over 48 streams. Undefined rules for Utility-masked histories are left blank. The figure is a compact display of the complete cell-level matrix; exact numeric values and joins remain in the reproducibility archive.}
\label{fig:joint-blackbox-reference-rules}
\end{figure}

Figure~\ref{fig:joint-blackbox-reference-rules} summarizes every defined reference-rule estimate across the prespecified forms, naming cells, and namespaces; points are estimates and bars are 95\% stream-bootstrap intervals. Cross-form comparisons use within-condition repeats from both forms. Paired bootstrap draws resample 48 streams within the three fixed involution strata; intervals condition on the constructed histories, tasks, and model interface. The ten targeted rule-disagreement groups use their actual eligible blocks and streams, which differ across rules and transformed histories. Rows with no eligible blocks are omitted because the corresponding subgroup statistics are undefined. The complete per-cell numeric matrices and all reference actions and task-utility joins are retained in the reproducibility archive and item-level analysis records. Full, value, pair, and key-slot conditions have no observed within-condition disagreement. Utility-masked histories have variable responses, so their corrected estimates can be negative. A zero empirical distance or a zero-width bootstrap interval on these fixed histories does not establish population invariance. Reference agreement describes the decisions on this support; it does not identify the model's internal computation. Joint utility contrasts use archived future outcomes, while the separate memory test in Appendix~\ref{app:new-diagnostic-experiments} supplies newly executed future-task evidence.

\begin{figure}[ht]
\centering
\includegraphics[width=0.86\linewidth]{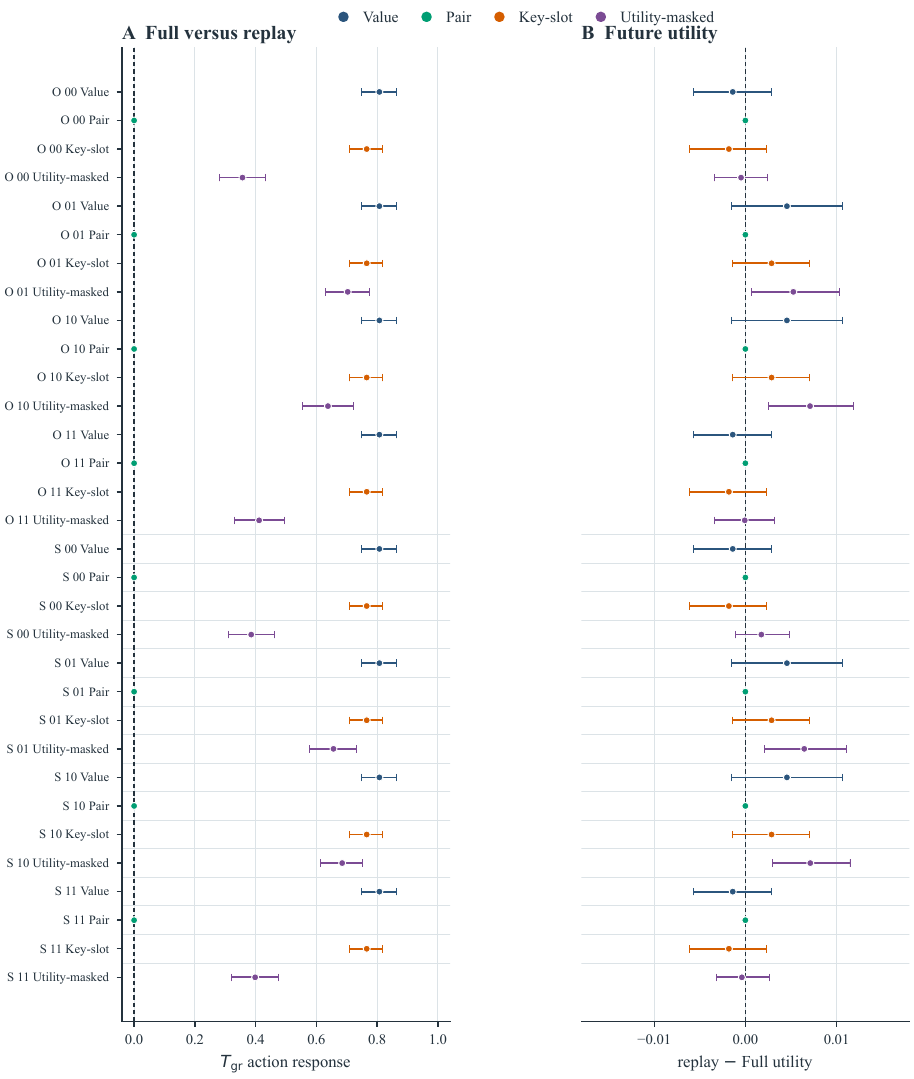}
\caption{\textbf{Cell-level joint black-box responses.} Points are estimates and horizontal bars are 95\% stream-bootstrap intervals over 48 streams. Rows use O/S for OPAQUE/SEMANTIC, the four history/menu cells, and the four replay forms. Panel A reports $T_{\mathrm{gr}}$ against Full; panel B reports replay-minus-Full future utility.}
\label{fig:joint-blackbox-replay-effects}
\vspace{2pt}
\includegraphics[width=0.86\linewidth]{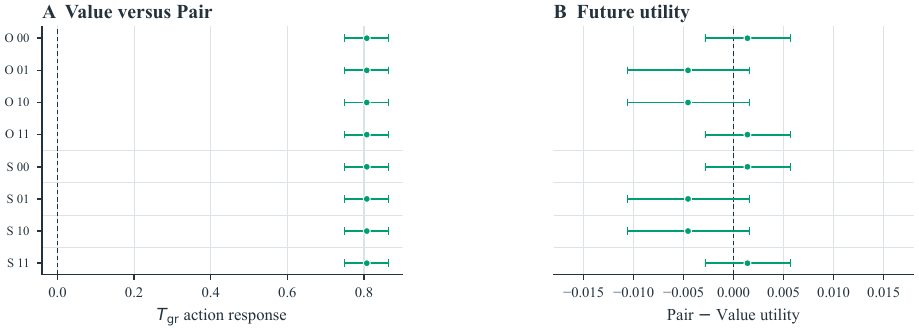}
\caption{\textbf{Value--Pair control.} Points and horizontal bars show the estimate and 95\% stream-bootstrap interval over the same 48 streams. Panel A is the direct action response; Panel B is the Pair-minus-Value future-utility difference, shown for both namespaces and all four cells.}
\label{fig:joint-blackbox-value-pair}
\end{figure}

Figures~\ref{fig:joint-blackbox-replay-effects} and \ref{fig:joint-blackbox-value-pair} retain the two scientific summaries from the full matrix: cell-level action and utility responses, followed by the value--pair comparison. The complete per-cell numeric matrices, reference actions, and task--utility joins remain in the reproducibility archive rather than being repeated as long appendix tables. The subgroup support is fixed by the eligible blocks rather than by a post-hoc choice of rows. Full and Pair contain all ten designed groups in both namespaces; Value and Key-slot contain every group except $g05$; Utility-masked contains only $g08$. The metadata table below records the corresponding block and stream counts.

\begin{table}[ht]
\vspace{-4mm}
\centering
\small
\setlength{\tabcolsep}{2pt}
\caption{Shared metadata for the ten designed rule-disagreement groups. B and S denote block and stream counts; the six map columns report each quantity separately.}
\label{tab:joint-app-reference-subgroup-metadata}
\begin{tabular}{lllrrrrrrrr}
\toprule
Group & Rule pair & Freq. & Nom. B & Nom. S & \multicolumn{2}{c}{$\rho_{01}$} & \multicolumn{2}{c}{$\rho_{02}$} & \multicolumn{2}{c}{$\rho_{03}$} \\
\midrule 
 &  &  &  &  & B & S & B & S & B & S \\
g01 & Mean--Sum & unbalanced & 20 & 10 & 6 & 4 & 9 & 4 & 5 & 2 \\
g02 & Mean--Count & unbalanced & 20 & 6 & 4 & 1 & 4 & 2 & 12 & 3 \\
g03 & Mean--Latest record & balanced & 19 & 6 & 4 & 1 & 5 & 2 & 10 & 3 \\
g04 & Mean--Best & balanced & 19 & 6 & 5 & 2 & 4 & 1 & 10 & 3 \\
g05 & Sum--Count & unbalanced & 19 & 9 & 10 & 4 & 6 & 3 & 3 & 2 \\
g06 & Sum--Latest record & balanced & 19 & 5 & 15 & 4 & 4 & 1 & 0 & 0 \\
g07 & Sum--Best & balanced & 19 & 5 & 4 & 1 & 7 & 2 & 8 & 2 \\
g08 & Count--Latest record & unbalanced & 19 & 6 & 0 & 0 & 11 & 4 & 8 & 2 \\
g09 & Count--Best & unbalanced & 19 & 6 & 1 & 1 & 10 & 3 & 8 & 2 \\
g10 & Latest record--Best & balanced & 19 & 5 & 15 & 4 & 4 & 1 & 0 & 0 \\
\bottomrule
\end{tabular}
\vspace{-2mm}
\end{table}

\end{document}